\documentclass[journal]{IEEEtran}

\usepackage{xcolor,soul,framed} 

\colorlet{shadecolor}{yellow}
\usepackage[pdftex]{graphicx}
\graphicspath{{../pdf/}{../jpeg/}}
\DeclareGraphicsExtensions{.pdf,.jpeg,.png}

\usepackage[cmex10]{amsmath}
\usepackage{array}
\usepackage{mdwmath}
\usepackage{mdwtab}
\usepackage{eqparbox}
\usepackage{url}
\usepackage{multirow}
\usepackage{amssymb}
\usepackage{stfloats}
\usepackage[ruled,linesnumbered, boxed, inoutnumbered]{algorithm2e}

\begin{document}
\bstctlcite{IEEEexample:BSTcontrol}
    \title{Graph Probability Aggregation Clustering}
  \author{Yuxuan Yan, Na Lu, Difei Mei, Ruofan Yan, and Youtian Du

\thanks{This work was funded by National Natural Science Foundation of China
Grant No. U22B2036, and Xi'an Jiaotong University-China Mobile Communications Group Co., Ltd. Digital Government Joint Institute. (Corresponding author: Na Lu.)}
  \thanks{Yuxuan Yan, Na Lu, Ruofan Yan and Youtian Du are with School of Automation Science and Engineering, Xi’an Jiaotong University, Xi’an 710049, China (e-mail: yan1611@stu.xjtu.edu.cn; lvna2009@xjtu.edu.cn; yanruofan@stu.xjtu.edu.cn; duyt@xjtu.edu.cn).}
  \thanks{Difei Mei is with China Mobile Xiong'an Information and Communication Technology Co., Ltd., China Mobile System Integration Co.,Ltd., China Mobile Information System Integration Co., Ltd., Beijing 100032, China (meidifei@cmict.chinamobile.com).}%

}

\maketitle

\begin{abstract}
Traditional clustering methods typically focus on either cluster-wise global clustering or point-wise local clustering to reveal the intrinsic structures in unlabeled data. Global clustering optimizes an objective function to explore the relationships between clusters, but this approach may inevitably lead to coarse partition. In contrast, local clustering heuristically groups data based on detailed point relationships, but it tends to be less coherence and efficient. To bridge the gap between these two concepts and utilize the strengths of both, we propose Graph Probability Aggregation Clustering (GPAC), a graph-based fuzzy clustering algorithm. GPAC unifies the global clustering objective function with a local clustering constraint. The entire GPAC framework is formulated as a multi-constrained optimization problem, which can be solved using the Lagrangian method. Through the optimization process, the probability of a sample belonging to a specific cluster is iteratively calculated by aggregating information from neighboring samples within the graph. We incorporate a hard assignment variable into the objective function to further improve the convergence and stability of optimization. Furthermore, to efficiently handle large-scale datasets, we introduce an acceleration program that reduces the computational complexity from quadratic to linear, ensuring scalability. Extensive experiments conducted on synthetic, real-world, and deep learning datasets demonstrate that GPAC not only exceeds existing state-of-the-art methods in clustering performance but also excels in computational efficiency, making it a powerful tool for complex clustering challenges.

\end{abstract}


%
\IEEEpeerreviewmaketitle


\section{Introduction}
\IEEEPARstart{C}{lustering} algorithms \cite{arabie1996complexity, xie2016unsupervised, zhong2005efficient}, as a fundamental technique in data mining, play a crucial role in uncovering hidden patterns within unlabeled data. It helps identify intrinsic structures of unlabeled data, which can significantly improve our understanding and analysis of the underlying relationships between data points \cite{von2012clustering}. Clustering algorithms maximize similarity within data clusters and minimize similarity between data clusters \cite{feng2023review}. Over the years, numerous clustering methods have been developed, with applications spanning a wide range of domains, such as image recognition \cite{li2022twin, chang2017deep}, representation learning \cite{huang2022learning}, and social networks \cite{handcock2007model}. These diverse applications highlight the versatility and importance of clustering in both theoretical and practical data science.

Conventional clustering methods are widely favored by researchers because of their strong interpretability and stable optimization properties. From a data processing perspective, clustering algorithms can be categorized into two main approaches: local clustering and global clustering. Local clustering methods focus on point-wise data relationships, performing clustering through message passing between nodes. Hierarchical clustering \cite{murtagh2012algorithms} is the most classical method, which groups data sequentially based on a hierarchy of points. For example, the First Integer Neighbor Clustering Hierarchy (FINCH) \cite{sarfraz2019efficient} algorithm merges nearest neighbors to perform agglomerative hierarchical clustering. Additionally, local clustering methods often utilize graph data for information transfer. Affinity Propagation (AP) \cite{frey2007clustering}, for instance, updates responsibility and availability messages between data points to progressively determine the clustering structure of unlabeled data. These methods typically rely on heuristic algorithms rather than explicit objective functions to frame the clustering problem, emphasizing the detailed relationships between instances to achieve finer data division. However, practical challenges persist, including theoretical simplicity, the need to tune additional hyperparameters \cite{schubert2017dbscan, bai2020new}, and higher computational costs \cite{murtagh2012algorithms, frey2007clustering}.
In contrast to local clustering, global clustering methods aim to find a cluster-wise partition that optimizes the specific objective function. K-means and its variants \cite{lin2015cann, van2003new, yang2017towards} are among the most commonly used clustering algorithms, partitioning datasets based on cluster centers. Their widespread use is due to their theoretical simplicity and robustness \cite{nie2021coordinate, nie2022effective}. However, global clustering methods have notable drawbacks, including sensitivity to initialization \cite{khan2004cluster}, reduced effectiveness in high-dimensional spaces \cite{aggarwal2004framework, gu2017fuzzy}, inconsistent performance \cite{arthur2006k, bradley2000constrained}, and limited precision in data division.

Fuzzy mathematics \cite{klir1995fuzzy} provides a new perspective for clustering. Fuzzy clustering \cite{ruspini2019fuzzy}, introduces the concept of fuzzy sets \cite{zimmermann2011fuzzy}, allowing data samples to belong to multiple clusters simultaneously. Fuzzy C-Means (FCM) \cite{bezdek1984fcm} is the most prominent fuzzy clustering algorithm, which iteratively updates the cluster centers and the membership matrix to achieve clustering. Subsequently, several fuzzy clustering algorithms \cite{krishnapuram1993possibilistic, pal1997mixed, pal2005possibilistic, ahmed2002modified, nie2021fast} have been derived from modifications of FCM. Knowledge-induced Multiple Kernel Fuzzy Clustering (KMKFC) \cite{tang2023knowledge} integrates the relative density-based knowledge extraction method and multiple kernel mechanism into FCM to better extract knowledge, improve adaptability, and guide the cluster centers. However, these methods remain rooted in incremental improvements of classical algorithms, which inherently limit their performance due to the constraints of the foundational algorithms. To this end, researchers have explored new frameworks \cite{bai2018ensemble, li2023fuzzy, gu2017fuzzy} for fuzzy clustering. One such approach is Fast Fuzzy Clustering based on Anchor Graph (FFCAG) \cite{gu2017fuzzy}, which combines anchor-based similarity graph construction with fuzzy membership matrix learning to mitigate the high computational cost and sensitivity to noise that is common in traditional fuzzy clustering methods. Similarly, Graph-based Soft-Balanced Fuzzy Clustering (GBFC) \cite{liu2022graph} uses the membership matrix in the Spectral Clustering (SC) framework and introduces a balancing constraint to regularize the clustering results. However, despite these advancements, these approaches still face limitations related to model theory, parameter selection, and overall algorithm performance.

\begin{figure}[t]
	\centering
	\includegraphics[width=1\linewidth]{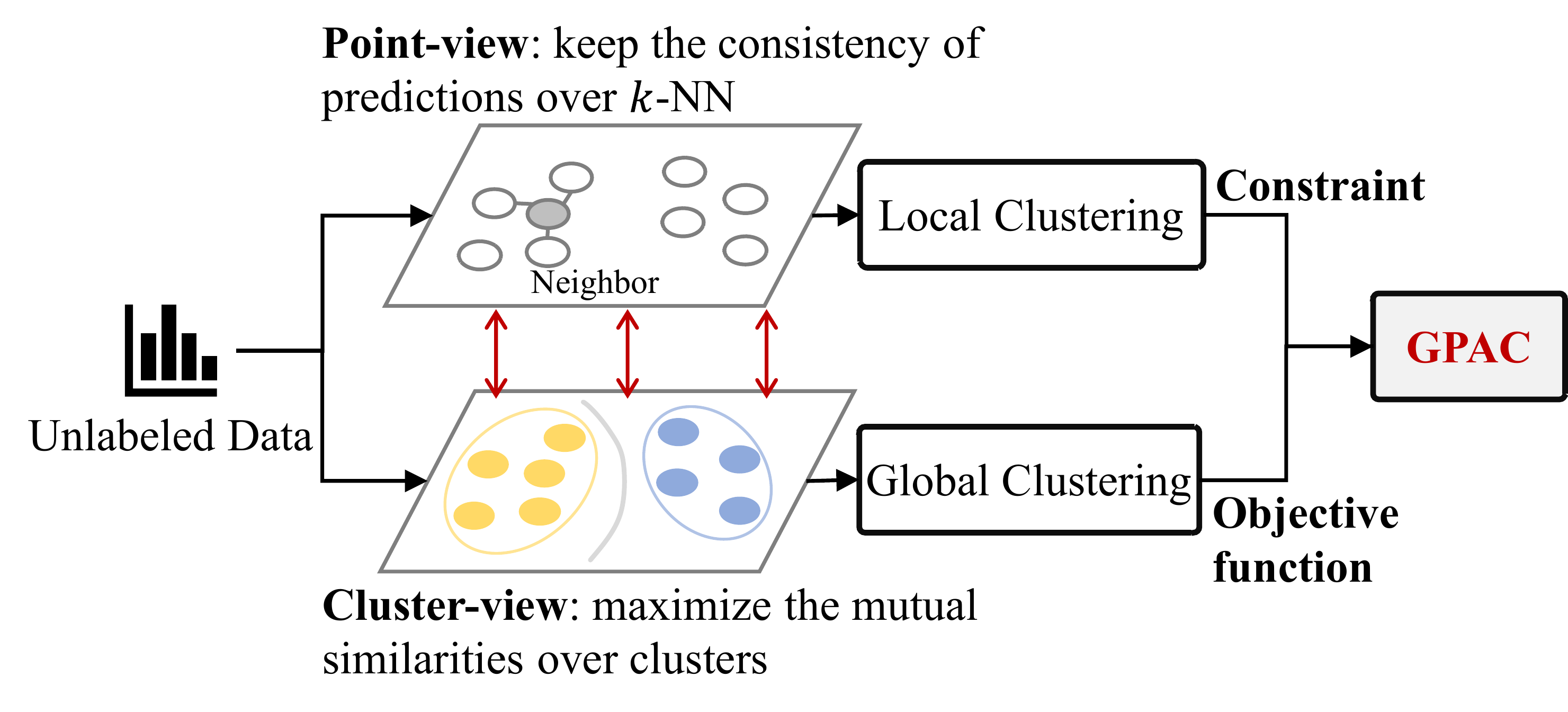}
	\caption{Overall research idea of the of GPAC. GPAC introduces an objective function for cluster-wise clustering and incorporates a consistency constraint for point-wise clustering. By combining global and local clustering approaches, GPAC significantly enhances clustering performance while maintaining low complexity.}
	\label{fig_1}
\end{figure}

In this paper, we introduce a novel graph-based fuzzy clustering algorithm, Graph Probability Aggregation Clustering (GPAC), which offers a new perspective of clustering by integrating graph theory with fuzzy techniques. As illustrated in Fig. \ref{fig_1}, GPAC utilizes an objective function for global clustering throughout the entire dataset, while incorporating a constraint to perform local clustering within neighborhoods. By combining the strengths of graph and fuzzy clustering, GPAC provides greater flexibility and robustness for the clustering process and hyperparameter tuning. The global clustering is driven by the objective function composed of two key components: a clustering term and a self-constrained term. The clustering term is designed to promote coherence of the clustering predictions across the global data distribution. It encourages the model to assign similar samples to the same cluster. The self-constrained term, acting as a regularization term, on the other hand, addresses the issue of cluster size imbalance. It penalizes solutions where a disproportionate number of samples are assigned to a small subset of clusters, thus preventing trivial solutions brought by the clustering term. We demonstrate that our objective function effectively leverages both similarities and dissimilarities among samples, showcasing its ability to capture the intrinsic structure of the data. Additionally, we prove that a special case of our objective function is mathematically equivalent to minimizing the sum of cosine distances within clusters. For local clustering, a local consistency constraint is employed to learn the local structural relationships within the data.  This constraint preserves the intricate relationships among the neighboring nodes in the graph, facilitating more accurate and meaningful clustering results. By incorporating both objective function and constraint, graph-based knowledge is aggregated by a fuzzy way to effectively capture both local and global structural relationships within the dataset.

As a result, the entire clustering problem is formulated as a multi-constrained quadratic optimization problem. To solve this problem efficiently, we integrate a fuzzy weighting exponent and a hard assignment matrix into the objective function. The fuzzy weighting exponent acts as a hyperparameter to control the degree of fuzziness in the probability output, while the hard assignment matrix functions as an auxiliary variable to facilitate convergence during optimization. Then we develop a concise optimization framework based on the projected Lagrange multiplier method to solve this problem. In this framework, the probability of a sample belonging to a cluster is iteratively refined by aggregating information from neighboring samples within the graph structure. Therefore, we refer to our method as Graph Probability Aggregation Clustering (GPAC). However, each iteration of GPAC requires the use of the entire probability matrix, resulting in quadratic time complexity. To improve scalability, we incorporate the mini-batch aggregation acceleration algorithm. GPAC strikes an effective balance between efficiency and performance, making it particularly well-suited for partitioning problems with a known number of clusters.

Finally, we evaluate GPAC on a diverse set of datasets that span various domains. Extensive experiments on multiple clustering benchmarks demonstrate that GPAC significantly outperforms state-of-the-art methods. It is worthwhile to highlight the major contributions of this work as follows:
\begin{itemize}
\item A well-defined objective function based on the sum of probabilities, incorporating the local clustering constraint, is proposed for unlabeled graph data. Our model effectively combines local and global clustering, enabling the grouping of data at both cluster and point views of the data distribution.

\item A Lagrangian optimization method, incorporating a fuzzy weighting exponent and an auxiliary variable, is proposed to effectively solve the proposed model. Additionally, an acceleration scheme is developed to reduce complexity during the optimization process. The optimization algorithm ensures faster convergence and improves the scalability of the clustering model, making it more suitable for large-scale datasets.

\item Extensive experiments have been conducted on nine widely used datasets to validate the effectiveness of our algorithm. The results demonstrate that our algorithm outperforms state-of-the-art methods, yielding satisfactory performance across all datasets.
\end{itemize}

The structure of this study is organized as follows: Section II reviews related work; Section III introduces the concept of the proposed GPAC algorithm; Section IV presents the experiments and analysis of the GPAC algorithm; and Section V concludes the paper with a summary of our findings.

\section{Related Work}
Our research is built upon graph-based and fuzzy-based methods. Therefore, we briefly review the development of both fuzzy and graph clustering approaches. We use $\mathcal{X}=\{\boldsymbol{x}_i \}_{i=1}^n$ to denote an $n$-point set, where $\boldsymbol{x}_{i}\in\mathbb{R}^{d}$ is the i-th $d$-dimensional sample. $\boldsymbol{W}$ is the $n \times n$ similarity graph based on $\mathcal{X}$. Our work aims to study the problem that the number of classes is known to be $c \in \mathbb{N}$, therefore the clustering goal aims to divide $\mathcal{X}$ into $c$ mutually disjoint clusters. In our previous work, Probability Aggregation Clustering (PAC) \cite{yan2024pac}, we introduced a preliminary centerless fuzzy clustering model based on the probability aggregation operation. Building on this approach, the current paper presents an effective improvement leveraging graph knowledge and provides a thorough theoretical analysis.

\subsection{Fuzzy Clustering}
In hard (or crisp) clustering, each instance is assigned to a single specific cluster. However, such approaches struggle to capture data where instances are intertwined or exhibit overlapping characteristics. To address this limitation, fuzzy clustering methods have been proposed, drawing on fuzzy mathematics to provide a more flexible representation of the data. 

Set $\boldsymbol{P}=[p_{i,l}]_{n\times c}$ denote the fuzzy membership matrix, where $p{i,l}$ represents the membership degree of instance $i$ to cluster $l$, which can also be interpreted as the assignment probability.
$\boldsymbol{P}$ satisfies $\boldsymbol{P}\in \mathcal{P}=\{[\phi_{i,l}]_{n\times c}|\phi_{i,l} \in [0,1]; \sum_{l=1}^{c}\phi_{i,l}=1; 0<\sum_{i=1}^{n}\phi_{i,l}<n,i=1,\cdots,n,l=1,\cdots,c\}$.  
Fuzzy C-Means (FCM) \cite{bezdek1984fcm} formulates the clustering problem based on Euclidean distances between instances and its corresponding cluster centers. The formulation of FCM is:
\begin{equation}\label{eqn-0-1} 
\min\limits_{\boldsymbol{P} \in \mathcal{P}} \sum_{i=1}^n \sum_{l=1}^c p_{i,l}^m \Vert \boldsymbol{x}_i- \boldsymbol{v}_l\Vert_2^2,
\end{equation}  
where $m>1$ is the fuzzy weighting exponent, and $\boldsymbol{v}_l$ represents the cluster center which is defined as ($l=1,\cdots,c$):
\begin{equation}\label{eqn-0-2} 
\boldsymbol{v}_l = \frac{\sum_{i=1}^n p_{i,l}^m\boldsymbol{x}_i}{\sum_{i=1}^n p_{i,l}^m}.
\end{equation}  
The optimal solution of FCM is ($i=1,\cdots,n; l=1,\cdots,c$):
\begin{equation}\label{eqn-3} 
    p_{i,l}= \frac{\Vert \boldsymbol{x}_i- \boldsymbol{v}_l\Vert_2^{2/(m-1)}}{\sum_{r=1}^c\Vert \boldsymbol{x}_i- \boldsymbol{v}_r\Vert_2^{2/(m-1)}}.
\end{equation} 
In FCM, the final cluster prediction of $\boldsymbol{x}_{i}$ is the highest probability, i.e., $\displaystyle \hat{p_{i}}=\arg\max\limits_{l}(\boldsymbol{p}_i)_l$.

\subsection{Graph-based Clustering}
In graph-based clustering, data points are represented as nodes in a graph. Among various techniques, the Spectral Clustering (SC) family is one of the most widely adopted graph-based methods. Self-Constrained Spectral Clustering (Self-CSC) \cite{bai2022self} enhances the objective function of SC by incorporating pairwise and label self-constrained terms, allowing high-quality clustering without relying on prior information. However, the time complexity of Self-CSC is cubic, limiting its scalability. Beyond modifications to the basic SC algorithm, researchers have also combined the SC framework with other clustering approaches to leverage the advantages of multiple techniques. For example, Centerless Clustering (K-sums) \cite{pei2022centerless} integrates SC with K-means to achieve efficient clustering. The objective function of K-sums minimizes the sum of distances between points within the same cluster, but it lacks the modeling of local detail structure. Despite these efforts, graph-based approaches often struggle to find a good balance between complexity and performance.

\section{Methodology}
This section presents the theoretical derivation of the GPAC algorithm. We begin by describing the motivation behind our method. Next, we construct the optimization model, providing a detailed formulation that captures the essential components and relationships within the algorithm. Finally, we develop a fast optimization program to efficiently solve the model.

\subsection{Main Idea of Global Clustering}
Given a probability vector $\boldsymbol{p}_{i}$ of dimension $c$ denoting the probability of the $i^{th}$ sample belonging to each cluster, the inner product of two probability vectors can be viewed as the probability that the two samples belong to the same cluster, which takes the following form:
\begin{equation}\label{eqn-1-1} 
		\rho_{i,j} = \boldsymbol{p}_i^{T}\boldsymbol{p}_j,
\end{equation}  
where $\rho_{i,j} \in [0,1]$. The larger $\rho_{i,j}$, the more likely the samples $i$ and $j$ are in the same cluster. An intuitive clustering approach is to maximize $\rho$ between adjacent samples under the guidance of the similarity matrix. And the clustering optimization problem can formulated as
\begin{equation}\label{eqn-1-3} 
		\max\limits_{\boldsymbol{P}\in\mathcal{P}} \sum_{i=1}^n \sum_{j\in \mathcal{A}_i} \rho_{i,j} w_{i,j},
\end{equation}  
where $\mathcal{A}_i$ is the set of adjacent samples of sample $i$, $w_{i,j}$ is the similarity of $i$ and $j$. 

However, there is a trivial solution to Eq. (\ref{eqn-1-3}), where all the samples are grouped into a single cluster. To address this, we propose a self-constrained term that sums all $\rho_{i,j}$
values to enforce the uniformity of the clustering prediction. By introducing this self-constrained term and transforming the maximization problem into a minimization problem, we formulate the new optimization problem as follows:
\begin{equation}\label{eqn-1-4} 
\min\limits_{\boldsymbol{P}\in\mathcal{P}} \sum_{i=1}^n \sum_{j\neq i} \rho_{i,j}- \alpha \sum_{i=1}^n \sum_{j\in \mathcal{A}_i} \rho_{i,j}w_{i,j},
\end{equation}  
where $\alpha>0$ is the penalty hyperparameter that controls the strength of the self-constrained term. We set $\alpha$ to 1 in this paper.

The structure of Eq. (\ref{eqn-1-4}) is similar to that of the SC problem and can be expressed in matrix form as:
\begin{equation}\label{eqn-e-1} 
\min\limits_{\boldsymbol{P}\in\mathcal{P}} Tr(\boldsymbol{P}^T(\boldsymbol{H}-\alpha\boldsymbol{W})\boldsymbol{P}),
\end{equation} 
where $\boldsymbol{H}=\boldsymbol{1}_n^{T}\boldsymbol{1}_n - \boldsymbol{I}$, $\boldsymbol{1}_{n}$ is the 1 vector of length $n$. While the graph partition problem of SC is formulated as:
\begin{equation}\label{eqn-e-2} 
  \begin{aligned}
	&\min\limits_{\boldsymbol{F}\in\mathbb{R}^{n\times c}} Tr(\boldsymbol{F}^T(\boldsymbol{I}-\boldsymbol{W})\boldsymbol{F}),\\
        & \mathrm{s.t.} \quad \boldsymbol{F}^{T}\boldsymbol{F}=\boldsymbol{I},
    \end{aligned}
\end{equation} 
where $\boldsymbol{F}$ is the indicator matrix, containing arbitrary real values subject to an orthogonality constraint.
We observe that the formulations of Eq. (\ref{eqn-e-1}) and Eq. (\ref{eqn-e-2}) are quadratic, which can be solved directly to obtain the minimum. Eq. (\ref{eqn-e-1}) transforms $\boldsymbol{F}$ to $\boldsymbol{P}$ and removes the orthogonal constraint in Eq. (\ref{eqn-e-2}) by replacing $\boldsymbol{I}$ with $\boldsymbol{H}$.

Depending on the values of $\boldsymbol{W}$ and $\alpha$, Eq. (\ref{eqn-1-4}) can be converted into two special cases:
\begin{itemize}
    \item If $w_{i,j}=1$, $j \in \mathcal{A}_i$, otherwise $w_{i,j}=0$, $\forall i$ and $\alpha=1$, Eq. (\ref{eqn-1-4}) equals to:
        \begin{equation}\label{eqn-1-5} 
        \min\limits_{\boldsymbol{P}\in\mathcal{P}} \sum_{j \in \Bar{\mathcal{A}_i}} \rho_{i,j},
        \end{equation}
        where $\Bar{\mathcal{A}_i}$ is the complement of $\mathcal{A}_i$. Eq. (\ref{eqn-1-6}) forces the non-adjacent samples into different clusters, that is, applying dissimilarity in clustering. Therefore, Eq. (\ref{eqn-1-5}) is the clustering model based on dissimilarity criteria. 
        
    \item If $w_{i,j}$ is the cosine similarity, $w_{i,j}=\frac{\boldsymbol{x}_i^T\boldsymbol{x}_j}{\Vert \boldsymbol{x}_i\Vert_2 \Vert \boldsymbol{x}_j\Vert_2}$ and $\alpha=0.5$, Eq. (\ref{eqn-1-5}) can be rewritten as:
    \begin{equation}\label{eqn-1-6} 
        \begin{aligned}
	&\min\limits_{\boldsymbol{P}\in\mathcal{P}} \sum_{i=1}^n \sum_{j \neq i} \rho_{i,j} ( 1-\frac{0.5\boldsymbol{x}_i^T\boldsymbol{x}_j}{\Vert \boldsymbol{x}_i\Vert_2 \Vert \boldsymbol{x}_j\Vert_2}),\\
        & =\min\limits_{\boldsymbol{P}\in\mathcal{P}}  \sum_{i=1}^n \sum_{j \neq i} \rho_{i,j} d_{i,j}^2,
        \end{aligned}
     \end{equation} 
    where $d_{i,j}^2 = 1-\frac{0.5\boldsymbol{x}_i^T\boldsymbol{x}_j}{\Vert \boldsymbol{x}_i\Vert_2 \Vert \boldsymbol{x}_j\Vert_2}$ is the square of standard cosine dissimilarity distance. Therefore, $\rho_{i,j} d_{i,j}^2$ can be understood as the sum of the distances between the samples within different clusters. Compared with FCM in Eq. (\ref{eqn-0-1}), Eq. (\ref{eqn-1-6}) circumvents the cluster center in formulation and model the clustering by sum of mutual distances.    
\end{itemize}

In summary, the core idea of GPAC is to implement fuzzy clustering by modifying the Spectral Clustering framework. The objective function of GPAC is designed to capture the underlying relationships between data points by considering both the similarities among neighboring samples and the dissimilarities between non-adjacent ones. Regarding optimization, the optimal solution to the problem in Eq. (\ref{eqn-1-6}) typically lies at the endpoints, causing the fuzzy partition to degenerate into a hard partition, which makes the problem difficult to solve. To address this issue and improve the optimization process, we incorporate a fuzzy weighting exponent and a hard assignment matrix. The details of final optimization problem are formulated in the following section.

\subsection{Optimization Problem Formulation}
In this section, we present the detailed mathematical model for Graph Aggregation Probability Clustering (GPAC). We begin by introducing the clustering model with local consistency constraint and constructing the $k$-nearest neighbor graph ($k$-NN). Next, we provide the optimization derivation for the proposed problem. Finally, we propose a fast optimization algorithm to efficiently solve the GPAC problem, addressing scalability concerns and enabling practical implementation.

\subsubsection{Clustering Model}
The entire optimization problem of GPAC in the matrix form is defined as:
\begin{equation}\label{eqn-2-1}
    \begin{aligned}
	&\min\limits_{\boldsymbol{P}\in\mathcal{P}, \boldsymbol{V}\in\mathcal{V}} Tr(\boldsymbol{V}^{T}\boldsymbol{H} \boldsymbol{V}
        +\boldsymbol{P}^{T}\boldsymbol{H} \boldsymbol{P}^m- \alpha \boldsymbol{V}^{T}\hat{\boldsymbol{W}}\boldsymbol{P}^m),\\
        &\quad\quad \mathrm{s.t.} \quad \boldsymbol{L}\boldsymbol{P}=0,
    \end{aligned}
\end{equation}
where $\boldsymbol{V}$ is the hard assignment matrix that satisfies $\boldsymbol{V}\in \mathcal{V}=\{[\phi_{i,l}]_{n\times c}|\phi_{i,l} \in \{0,1\}; \sum_{l=1}^{c}\phi_{i,l}=1; 0<\sum_{i=1}^{n}\phi_{i,l}<n,i=1,\cdots,n,l=1,\cdots,c\}$, $m\in(1,+\infty)$, $\boldsymbol{P}^m=[p_{i,k}^m]\in\mathbb{R}^{+n \times c}$ denotes that every element of the matrix is raised to a power. We set $\alpha$ to 1 for all experiments to balance the effort of the components. In constraint, $\boldsymbol{L}$ is the Laplacian matrix for graph $\boldsymbol{W}$, $\boldsymbol{L}=\boldsymbol{\Delta}-\boldsymbol{W}$, $\boldsymbol{\Delta}$ is the degree matrix of $\boldsymbol{W}$, $\Delta_{i,i}=\sum_{j=1}^n w_{i,j}$. 

In Eq. (\ref{eqn-2-1}), the first two terms are the regularization terms, while the third term corresponds to the clustering term. To prevent extreme points from appearing at the endpoints, we incorporate the fuzzy weighting exponent $m$ into the objective function. 
Additionally, we introduce an auxiliary hard assignment matrix $\boldsymbol{V}$ to facilitate optimization. $\boldsymbol{V}$, consisting of 0 and 1, accelerates convergence in the early stages of optimization, making $\boldsymbol{P}$ more confident. In later stages, $\boldsymbol{V}$ is updated less frequently than $\boldsymbol{P}$, which improves the stability of the model. 

\subsubsection{Local Clustering Constraint}
The objective function in Eq. (\ref{eqn-2-1}) aims to identify a reasonable grouping pattern from a global perspective. From the viewpoint of an individual instance, we seek to ensure that the probability of sample $i$ is consistent with its $k$-nearest neighbors, thereby preserving the consistency of local information \cite{zhou2003learning}. To achieve this, we incorporate a local consistency constraint (LCC) into the model to ensure that samples remain consistent with their neighbors. By introducing the Laplace matrix definition into the constraint, the constraint in Eq. (\ref{eqn-2-1}) can be rewritten as:
\begin{equation}\label{eqn-2-5}
    \boldsymbol{L}\boldsymbol{P}=(\boldsymbol{\Delta}-\boldsymbol{W})\boldsymbol{P}=0.
\end{equation}
Right multiply both sides of the equation by the matrix $\boldsymbol{\Delta}^{-1}$, we can get:
\begin{equation}\label{eqn-2-5}
   \boldsymbol{P}=\boldsymbol{\Delta}^{-1}\boldsymbol{W}\boldsymbol{P},
\end{equation}
which establishes the relationship of predictions between instance and its $k$NN. The probability vector equals to $\boldsymbol{p}_i=\frac{\sum_{j = 1}^nw_{i,j}\boldsymbol{p}_j}{\sum_{j=1}^n w_{i,j}}, \forall i$, that is, $\boldsymbol{p}_i$ is the weighted average of its $k$ nearest neighbors predictions. 

\subsubsection{Similarity Graph Construction}
The clustering model involves two graphs: the $k$-NN graph $\boldsymbol{W}$ and the adjacency indicator graph $\hat{\boldsymbol{W}}$. The weighted adjacency matrix of $k$-NN graph $\boldsymbol{W}= [w_{i,j}]_{n\times n}$ is constructed by ($i=1,\cdots,n; j=1,\cdots,n$):   
\begin{equation}\label{eqn-2-2}
    w_{i,j}=\begin{cases}
		\exp{(-\frac{\Vert\boldsymbol{x}_i-\boldsymbol{x}_j\Vert_2^2}{2\sigma})}, & j \in \mathcal{N}^k(i) \text{ or } i \in \mathcal{N}^k(j)\\
			0, & \text{otherwise}
		\end{cases}
\end{equation}
where $\sigma$ is the kernel parameter of exponent, $\mathcal{N}^k(i)$ denotes the set of k-nearest neighbors for sample $i$. 

For the adjacent indicator graph $\hat{\boldsymbol{W}}$, it is constructed based on $\boldsymbol{W}$. The adjacent indicator graph $\hat{\boldsymbol{W}}$ is introduced to find sufficient neighborhood information to support clustering when $k$ is too small. Specifically, we use the random walk operation \cite{lovasz1993random} to extend the neighborhood of $\boldsymbol{W}$. $\hat{\boldsymbol{W}}$ is calculated as:
\begin{equation}\label{eqn-2-3}
    \hat{\boldsymbol{W}} = \mathbb{I}((\boldsymbol{W}+\boldsymbol{I})^\theta>0)-\boldsymbol{I},
\end{equation}
where $\theta$ is the positive integer, $\mathbb{I}(\cdot)$ is the element-wise indictor function, $\mathbb{I}(\boldsymbol{W}>0)=[\mathbb{I}(w_{i,j}>0)]_{n\times n}$, which equals one if $w_{i,j}>0$. $\hat{\boldsymbol{W}}$ is the sparse matrix of 0 and 1, which can greatly reduce the computation consumption in the iterative program. The hyperparameter $\theta$ controls the size of adjacent samples. Empirically, we set the $\theta$ so that $\max\{rank(\boldsymbol{w}_i)\}_i$ approaches $n/c$:
\begin{equation}\label{eqn-2-4}
    \theta = [ \log_{k}{\frac{n}{c}} ],
\end{equation}
where $[ \cdot ]$ is the integer up function.

\subsection{Iterative Optimization of GPAC}
The graph probability aggregation clustering problem in Eq. (\ref{eqn-2-5}) is a multi-constrained optimization problem. It is noted that $\boldsymbol{P}$ is independent of $\boldsymbol{V}$, so GPAC optimizes the objective function by iteratively updating $V$ and $P$.

\subsubsection{Fix $\boldsymbol{V}$ and Update $\boldsymbol{P}$} Since the modeling of Eq.(\ref{eqn-2-5}) is not conducive to direct solution, we treat the local consistency constraint as a slack constraint and use the projected Lagrangian algorithm \cite{calamai1987projected} to solve $\boldsymbol{P}$. 

Firstly, the objective function without local consistency constraint is simplified to:
\begin{equation}\label{eqn-2-6}
 \begin{aligned}
&\min\limits_{\boldsymbol{P}\in \mathcal{P}}Tr(\boldsymbol{V}^{T}\boldsymbol{H} \boldsymbol{V}
        +\boldsymbol{P}^{T}\boldsymbol{H} \boldsymbol{P}^m- \alpha \boldsymbol{V}^{T}\hat{\boldsymbol{W}}\boldsymbol{P}^m)\\
&\Leftrightarrow\min\limits_{\boldsymbol{P}\in \mathcal{P}}
Tr((\boldsymbol{P}^{T}(\boldsymbol{1}_{n}\boldsymbol{1}_{n}^{T}-\boldsymbol{I})-\alpha\boldsymbol{V}^{T}\hat{\boldsymbol{W}})\boldsymbol{P}^m).
 \end{aligned}
\end{equation} 
Treat $\boldsymbol{p}_i$ as the variable and the others as fixed constants and decompose the above problem into $n$ sub-problems to solve them in turn. The above equation can be transformed into an unconstrained sub-problem by the Lagrangian multiplier method ($i=1,\cdots,n$):
\begin{equation}\label{eqn-2-7}
     \begin{aligned}
        L_1(\boldsymbol{p}_i,a_i,\boldsymbol{b}_i)=&\sum_{l=1}^{c} \sum_{j\neq i} p_{i,l}^m (p_{j,l}-\alpha v_{j,l} \hat{w}_{i,j} )\\
        &  +a_i (1-\sum_{l=1}^c p_{i,l})- \sum_{l=1}^c b_{i,l}p_{i,l},
     \end{aligned}
\end{equation}  
where $a_i$ and $\boldsymbol{b}_i=[b_{i,l}]_{c \times 1}$ are the Lagrange multipliers respectively for the sum constraint and the non-negativity constraint on $\boldsymbol{p}_i$.
The partial derivative of ${L}_1$ with respect to $p_{i,l}$ should be equal to zero at the minimum point, so we can obtain ($i=1\cdots,n; l=1\cdots,c$):
 \begin{equation}\label{eqn-2-8}
     \frac{\partial L_1}{\partial p_{i,l}} = mp_{i,l}^{m-1}\sum_{j\neq i}(p_{j,l} - \alpha v_{j,l}\hat{w}_{i,j})  - a_i - b_{i,l}=0.
\end{equation}
Then, we can get following Karush-Kuhn-Tucker (K.K.T.) conditions ($i=1,\cdots,n; l=1,\cdots,c$):
\begin{equation}\label{eqn-2-9}
    \begin{cases}
    &\frac{\partial L^{'}}{\partial a_i} = 1-\sum_{l=1}^{c}p_{i,l}=0,\\
    &b_{i,l}p_{i,l}=0,\\
     &b_{i,l} \geq 0. 
    \end{cases}
\end{equation}
We define the fuzzy clustering score $s_{i,l}^p\in \mathbb{R}^+$ as ($i=1\cdots,n; l=1\cdots,c$):
\begin{equation}\label{eqn-2-10}
    s_{i,l}^p=\sum_{j\neq i}(p_{j,l} - \alpha v_{j,l}\hat{w}_{i,j}),
\end{equation}
 and it has vector form:
\begin{equation}\label{eqn-2-11}
    \boldsymbol{s}^{p}_{i}=[s_{i,1}^{p}, \cdots, s_{i,c}^{p}]^{T}=(\boldsymbol{1}_{n}^{T}\boldsymbol{P}-\boldsymbol{p}_{i})-\alpha \hat{\boldsymbol{w}}_{i}^{T}\boldsymbol{V},
\end{equation}
where $\hat{\boldsymbol{w}}_i$ is the $i$ row of matrix $\hat{\boldsymbol{W}}$. The fuzzy clustering score in Eq. (\ref{eqn-2-10}) indicates the score that $\boldsymbol{x}_i$ belongs to cluster $k$, which is non-negative. Eq. (\ref{eqn-2-8}) can be rewritten as
$\frac{\partial L_1}{\partial p_{i,l}} = mp_{i,l}^{m-1}\boldsymbol{s}^{p}_{i}  - a_i - b_{i,l}=0$.
According to the conditional relation of Eq. (\ref{eqn-2-9}), if $b_{i,l}>0$, we have $p_{i,l}=0$ and $a_i<0$. Meanwhile, there exists a $l^+$, $p_{i,l^+}>0$, which means that $b_{i,l^+}=0$. Based on $a_i<0$ and $b_{i,l^+}=0$, we can get $\frac{\partial L_1}{\partial p_{i,l^+}}>0$. It is not difficult to find that the above extrapolations are contradictory, therefore $b_{i,l}=0, \forall l$. And Eq.(\ref{eqn-2-8}) can be reformulated as ($i=1,\cdots,n; l=1\cdots,c$):
\begin{equation}\label{eqn-2-12}
        p_{i,l} =\kappa\frac{a_i^{-1/(m-1)}}{(\sum_{j\neq i} 2p_{j,l}^m - \alpha v_{j,l}\hat{w}_{i,j})^{-1/(m-1)}},
\end{equation}
where $\kappa$ is a constant, $\kappa=m^{-1/(m-1)}$. Considering the probability sum constraint $\sum_{l=1}^{c}p_{i,l}=1$, we have ($i=1,\cdots,n$):
\begin{equation}\label{eqn-2-a1}
       a_i^{-1/(m-1)} = \frac{1}{\kappa}\sum_{k=1}^{c}{{s_{i,k}^p}^{-1/(m-1)}}.
\end{equation}
Substituting $a_i^{1/(m-1)}$ into Eq. (\ref{eqn-2-12}),  we can finally obtain following equivalence relation ($i=1,\cdots,n; k=1\cdots,c$):
\begin{equation}\label{eqn-2-13}
        p_{i,l}^*=\frac{{s_{i,l}^p}^{-1/(m-1)}}{\sum_{r=1}^{c}{s_{i,r}^p}^{-1/(m-1)}}.
\end{equation}
Eq. (\ref{eqn-2-13}) sharpens $\boldsymbol{s}_{i}$ by power operation and normalizes the clustering score to get the probabilistic output.

Secondly, we project the obtained $\boldsymbol{P}^*$ to the feasible region according to the local consistency constraint. Let $\boldsymbol{P}^*$ denote the extreme point of Eq. (\ref{eqn-2-6}), the projected optimization problem is formulated as:
\begin{equation}\label{eqn-2-14}
    \begin{aligned}
	&\min\limits_{\boldsymbol{P}\in\mathcal{P}} \Vert\boldsymbol{P}-\boldsymbol{P}^*  \Vert_\mathcal{F}^2,\\
        & \mathrm{s.t.} 
        (\boldsymbol{I}-\boldsymbol{W}\boldsymbol{\Delta}^{-1})\boldsymbol{P}=0.
    \end{aligned}
\end{equation}
We introduce the slack constraint into objective function and transform Eq. (\ref{eqn-2-14}) into following problem:
\begin{equation}\label{eqn-2-15}
\min\limits_{\boldsymbol{P}\in\mathcal{P}} \Vert\boldsymbol{P}-\boldsymbol{P}^*  \Vert_\mathcal{F}^2 + \beta \Vert  \boldsymbol{P}-\boldsymbol{W}\boldsymbol{\Delta}^{-1}\boldsymbol{P}\Vert_\mathcal{F}^2,
\end{equation}
where $\beta>0$ the penalty hyperparameter. Eq. (\ref{eqn-2-15}) can be regared as a label propagation problem that limits the feasible domain to the probability space. We can obtain the following Lagrange function ($i=1,\cdots,n$):
\begin{equation}\label{eqn-2-16}
 \begin{aligned}
    L_2(\boldsymbol{p}_i,\boldsymbol{p}_i^*, c_i,\boldsymbol{d}_i)=&
    \Vert\boldsymbol{p}_i-\boldsymbol{p}_i^* \Vert_2^2 
    +\beta \Vert \boldsymbol{p}_i-\frac{\sum_{j = 1}^nw_{i,j}\boldsymbol{p}_j}{\sum_{j=1}^n w_{i,j}} \Vert_2^2
    \\& +c_i (1-\sum_{l=1}^c p_{i,l})- \sum_{l=1}^c d_{i,l}p_{i,l}.
 \end{aligned}
\end{equation} 
Using $\Bar{p_i}$ to denote $\Bar{p}_{i,l} = \frac{\sum_{j = 1}^nw_{i,j}{p}_{i,l}}{\sum_{j=1}^n w_{i,j}}$, the partial derivative is equal to ($i=1,\cdots,n$):
\begin{equation}\label{eqn-2-17}
     \frac{\partial L_2}{\partial p_{i,l}} = 
     2*(p_{i,l} -p_{i,l}^*)+2\beta(p_{i,l}-\Bar{p}_{i,l}) - c_i - d_{i,l}=0
\end{equation}
Afterwards, same as above solving process, taking K.K.T. condition into consideration, the solution is ($i=1,\cdots,n; l=1,\cdots,c$):
\begin{equation}\label{eqn-2-18}
    p_{i,l} = \frac{1}{1+\beta}p_{i,l}^*+\frac{\beta}{1+\beta} \bar{p}_{i,l}.
\end{equation}
Therefore, the final optimal solution of Eq. (\ref{eqn-2-1}) is:
\begin{equation}\label{eqn-2-21}
      p_{i,l} = \frac{1}{1+\beta}\frac{{s_{i,l}^p}^{-1/(m-1)}}{\sum_{r=1}^{c}{s_{i,r}^p}^{-1/(m-1)}}
                +\frac{\beta}{1+\beta} \frac{\sum_{j=1}^n w_{i, j}p_{j,l}}{\sum_{j=1}^n w_{i, j}}.
\end{equation}
\subsubsection{Fix $\boldsymbol{V}$ and Upadte $\boldsymbol{V}$} 
When fuzzy partition matrix $\boldsymbol{P}$ is fixed, the problem in Eq. (\ref{eqn-2-5}) degenerates to:
\begin{equation}\label{eqn-2-22} 
\min\limits_{\boldsymbol{V}}
Tr(\boldsymbol{V}^{T}(\boldsymbol{1}_{n}\boldsymbol{1}_{n}^{T}-\boldsymbol{I})-\alpha{\boldsymbol{P}^{m}}^{T}\hat{\boldsymbol{W}})\boldsymbol{V}.
\end{equation} 
Treating $\boldsymbol{v}_i$ as a variable and the others as constants, Eq. (\ref{eqn-2-22}) can be reformulated as ($i=1,\cdots,n$):
\begin{equation}\label{eqn-2-23} 
    \min\limits_{\boldsymbol{v}_i}
    {(\boldsymbol{V}^{T}\boldsymbol{1}_{n}-\boldsymbol{v}_i\alpha{\boldsymbol{P}^{m}}^{T}\hat{\boldsymbol{w}}_i)}^T \boldsymbol{v}_i,
\end{equation} 
Same as $\boldsymbol{P}$, there also is a hard clustering score $s_{i,l}^v\in \mathbb{R}^+$  ($i=1\cdots,n; l=1\cdots,c$):
\begin{equation}\label{eqn-2-24} 
 \boldsymbol{s}^{v}_{i}=[s_{i,1}^v, \cdots, s_{i,c}^v]^{T}=\boldsymbol{1}_{n}^{T}\boldsymbol{V} -\boldsymbol{v}_i-\alpha\hat{\boldsymbol{w}}_i^{T}\boldsymbol{P}^{m}.
\end{equation} 
Based on the properties of the hard assignment matrix, the minimum point of the Eq. (\ref{eqn-2-23}) is ($i=1\cdots,n$):
\begin{equation}\label{eqn-2-25} 
 \begin{aligned}
  &\begin{cases}
     &v_{i, l}=1, \quad l = l^*, \\
     &v_{i, l}=0,  \quad l \neq l^*, \\
 \end{cases}\\
    &\text{with} \quad l^* = \arg\min\limits_{l}(\boldsymbol{s}^{v}_{i})_l.
 \end{aligned}
\end{equation} 
Eq. (\ref{eqn-2-25}) indicates that the hard label $\boldsymbol{v}_i$ is the minimum point of the hard clustering score. 

Both $\boldsymbol{P}$ and $\boldsymbol{V}$ can be solved by iteratively optimizing Eq. (\ref{eqn-2-21}) and Eq. (\ref{eqn-2-25}) for each $\boldsymbol{p}_i$ and $\boldsymbol{v}_i$ until converging.

\subsection{Theoretical Analysis for Self-Constraint Terms}
In this section we analyze the roles of the self-constraint terms in GPAC, which make the cluster output uniform. For hard label term $Tr(\boldsymbol{V}^{T}\boldsymbol{H} \boldsymbol{V})$, it can be reformulated as:
\begin{equation}\label{eqn-2-26}
\begin{aligned}
    Tr(\boldsymbol{V}^{T}\boldsymbol{H} \boldsymbol{V})&= Tr(\boldsymbol{V}^{T}\boldsymbol{1}_n^{T} \boldsymbol{1}_n \boldsymbol{V}) -  Tr(\boldsymbol{V}^{T} \boldsymbol{V})\\
    &= \Vert \boldsymbol{V}^{T}\boldsymbol{1}_n  \Vert_2^2 - \Vert \boldsymbol{V} \Vert_{\mathcal{F}}^2 \\
    &= \sum_{l=1}^c(\sum_{i=1}^n v_{i,l})^2 - n.
 \end{aligned}
\end{equation}
According to the Jensen inequality, we can obtain:
\begin{equation}\label{eqn-2-27}
\begin{aligned}
   \sum_{l=1}^c(\sum_{i=1}^n v_{i,l})^2 - n \geq \frac{1}{c} (\sum_{l=1}^c\sum_{i=1}^n v_{i,l})^2 - n=\frac{n^2-cn}{c}.
 \end{aligned}
\end{equation}
To make the inequality hold with equality, we have $\sum_{i=1}^n v_{i,1}=\sum_{i=1}^n v_{i,2}=\sum_{i=1}^n v_{i,l}=\cdots=\sum_{i=1}^n v_{i,l}=n/c$.
For fuzzy label term $Tr(\boldsymbol{P}^{T}\boldsymbol{H} \boldsymbol{P}^m)$,
from Eq. (\ref{eqn-2-13}) we can know that the minimum point follows:
\begin{equation}\label{eqn-2-a2}
   p_{i,l}=\frac{({\sum_{j \neq i}p_{j,l}})^{-1/(m-1)}}{\sum_{r=1}^{c}({\sum_{j \neq i}p_{j,r}})^{-1/(m-1)}}.
\end{equation}
It is not difficult to find $p_{i,l}=1/c, \forall i,k$ satisfies Eq. (\ref{eqn-2-a2}) and is the global minimum point, because the extreme point is unique in the feasible domain. Hence, when we minimize the objective function, the self-constrained regular terms tend to distribute the predictions evenly across each cluster.

\subsection{Fast Optimization Algorithm}
Based on the theoretical solution of \(\boldsymbol{P}\) and \(\boldsymbol{V}\), we can iteratively solve the proposed optimization problem. However, in practical implementation, the floating-point calculations of the clustering score in Eq. (\ref{eqn-2-11}) and Eq. (\ref{eqn-2-24}) consume significant computational resources. For each clustering score, calculating the vector product requires about $2nc$ times additions and multiplications, resulting in a quadratic time complexity for the entire iteration process. Therefore, it is essential to reduce computational complexity to linear in order to better handle large-scale problems.
\subsubsection{Aggregation Acceleration} 
$\hat{\boldsymbol{W}}$ is the matrix of 0 and 1, which leads us to seek more efficient ways of aggregating calculations. Let ${\boldsymbol{P}(t)}$ denote the iterative sequence of $\boldsymbol{P}$ in optimization program, where $t$ is the iteration number. Eq.(\ref{eqn-2-11}) and Eq.(\ref{eqn-2-24}) can be rewritten as ($i=1\cdots,n$):
\begin{equation}\label{eqn-2-28}
    \boldsymbol{s}^{p}_{i}(t+1)=\tilde{\boldsymbol{P}}(t)-\boldsymbol{p}_{i}(t)-\alpha\sum_{j \in  \mathcal{A}_i}\boldsymbol{v}_{j}(t),
\end{equation}
\begin{equation}\label{eqn-2-29}
    \boldsymbol{s}^{v}_{i}(t+1)=\tilde{\boldsymbol{V}}(t) -\boldsymbol{v}_{i}(t)-\alpha\sum_{j \in \mathcal{A}_i}\boldsymbol{p}_{j}(t)^m,
\end{equation}
where $\tilde{\boldsymbol{P}}(t)=\sum_{i=1}^n\boldsymbol{p}_i(t)$ and $\tilde{\boldsymbol{V}}(t)=\sum_{i=1}^n\boldsymbol{v}_i(t)$, $\mathcal{A}=\{\mathcal{A}_1,\cdots,\mathcal{A}_n\}$ is the adjacent sample set $\mathcal{A}_i=\{j |w_{i,j}=1\}$. By Eq. (\ref{eqn-2-29}-\ref{eqn-2-30}), the multiplication operation is removed.
 Furthermore, $\tilde{\boldsymbol{P}}(t+1)$ and $\tilde{\boldsymbol{V}}(t+1)$ can be updated by:
\begin{equation}\label{eqn-2-30}
\tilde{\boldsymbol{P}}(t+1)= \tilde{\boldsymbol{P}}(t) - \boldsymbol{p}_{i}(t)+\boldsymbol{p}_{i}(t+1),
\end{equation}
\begin{equation}\label{eqn-2-31}
\tilde{\boldsymbol{V}}(t+1)= \tilde{\boldsymbol{V}}(t) - \boldsymbol{v}_{i}(t)+\boldsymbol{v}_{i}(t+1),
\end{equation}
which denotes that $\tilde{\boldsymbol{P}}$ and $\tilde{\boldsymbol{V}}$ can be updated and reused. 
Therefore, GPAC only needs to add the vector in $\mathcal{A}_i$ in each iteration. The specific algorithm is described in Algorithm 1.
\begin{algorithm}[h]
	\caption{Aggregation Acceleration Function: $F$}\label{alg:alg1}
	\KwIn{$\tilde{\boldsymbol{P}}$, $\tilde{\boldsymbol{V}}$, $\mathcal{A}_i$}
        $\boldsymbol{p}^{'}=\boldsymbol{v}^{'}=0$\\
        \For{$j$ in $\mathcal{A}_i$}
            {$\boldsymbol{p}^{'}+=\boldsymbol{p}_j^m$; $\boldsymbol{v}^{'}+=\boldsymbol{v}_j$
            }
	$\boldsymbol{s}^{p}_{i}=\tilde{\boldsymbol{P}}-\alpha\boldsymbol{v}^{'}$; $\boldsymbol{s}^{v}_{i}=\tilde{\boldsymbol{V}}-\alpha\boldsymbol{p}^{'}$;\\
        \KwOut {$\boldsymbol{s}^{p}_{i}$, $\boldsymbol{s}^{v}_{i}$}
\end{algorithm}

\subsubsection{Mini-batch Clustering} 
Most calculations of clustering scores are redundant, as the numerical relationship within $\boldsymbol{s}^{v}_i$ and $\boldsymbol{s}^{p}_i$ does not change significantly when calculating parts of the data. Therefore, it is natural to consider using a mini-batch clustering algorithm to speed up the optimization process. We randomly divide the dataset into $n_b$ mutually disjoint subsets, $\mathcal{X} = \{\mathcal{X}_1, \dots, \mathcal{X}_{n/n_b}\}$, where $n_b$ is the batch size. Vectors $\boldsymbol{s}^{v}_i$ and $\boldsymbol{s}^{p}_i$ are computed in subsets $\mathcal{X}_j$, with $\boldsymbol{x}_i \in \mathcal{X}_j$. It is worth noting that the proposed local consistency constraint mitigates the degeneration of mini-batch operations, as the samples are also constrained to remain consistent with their neighborhoods. In Section IV, we conduct experiments to demonstrate that our mini-batch strategy does not affect the final performance. By using mini-batch clustering, the computational complexity is further reduced.

\subsection{Algorithm Details} 
\subsubsection{Label Matrix Initialization}
GPAC initializes the hard assignment matrix $\boldsymbol{V}$ by K-means++ algorithm, and initializes the matrix $\boldsymbol{P}$ according to a uniform distribution ($i = 1, \dots, n; k = 1, \dots, c$):
 \begin{equation}\label{eqn-2-32}
    p_{i,k} = \frac{1}{c}.
\end{equation}

\subsubsection{Avoid Negative Score}
The clustering score $\boldsymbol{s}^{p}_{i}$ is greater than zero in the derivation. However, to prevent algorithm failure due to $\boldsymbol{s}^{p}_{i}$ becoming less than zero as a result of mini-batch clustering, we introduce an additional regularization formula for the calculation of $\boldsymbol{s}^{p}_{i}$. The adjusted $\boldsymbol{s}^{p}_{i}$ is defined as follows  ($i = 1, \dots, n; k = 1, \dots, c$):
 \begin{equation}\label{eqn-2-33}
   \boldsymbol{s}^{p}_{i} = \boldsymbol{s}^{p}_{i} - \min{\boldsymbol{s}^{p}_{i}} + 1.
\end{equation}
Eq. (\ref{eqn-2-33}) ensures that the minimum value of $\boldsymbol{s}^{p}_{i}$is constrained, which facilitates subsequent power operation.

\subsubsection{Algorithm Pseudocode}
Based on the above introduction, the final Graph Probability Aggregation Clustering algorithm is summarized in Algorithm 2. 

\subsubsection{Hyperparameters Selection}
The main hyperparameters in GPAC include the fuzzy weighting exponent $m$, the regularization hyperparameter $\alpha$, and penalty hyperparameter $\beta$. The fuzzy weighting exponent $m$ controls the degree of sharpening of the probability. We recommend choosing $m$ closer to 1 to make the probability outputs more confident. As for the penalty hyperparameter $\alpha$, it controls the strength of the regularization term. Through several experiments, we find that it is not highly sensitive to parameter selection. Therefore, we set it to $1$ for all experiments. As for penalty hyperparameter $\beta$, it controls the strength of the local consistency constraint. The selection principle for $\beta$ is to choose the largest possible value.

\begin{algorithm}[h]
	\caption{The Pseudocode of GPAC Algorithm}\label{alg:alg2}
	\KwIn{Dataset $\mathcal{X}$, the number of cluster $c$, parameters $m$, $\alpha$, $\beta$, $n_b$}
        Calculate $\boldsymbol{W}$ and $\hat{\boldsymbol{W}}$ by Eq. (\ref{eqn-2-2}) and Eq. (\ref{eqn-2-3});\\
        Construct adjacent sample set $\mathcal{A}=\{\mathcal{A}_1, \cdots, \mathcal{A}_n\}$ based on $\hat{\boldsymbol{W}}$;\\
        Initialize $\boldsymbol{V}$ randomly and set $p_{i,l}=1/c, \forall i,l$;\\
        \While{not converge}{
        \text{Randomly Partition dataset $\mathcal{X}=\{\mathcal{X}_1,\cdots,\mathcal{X}_{[n/n_b]}\}$;}\\
         \For{$r = 1$ \KwTo $[n/n_b]$}
         {$\tilde{\boldsymbol{P}}=\sum_{i=1}^n\boldsymbol{p}_i;\tilde{\boldsymbol{V}}=\sum_{i=1}^n\boldsymbol{v}_i;$\\
         \For{$\boldsymbol{x}_i$ in $\mathcal{X}_r$}
            {$\mathcal{A}_i^{'}=\mathcal{A}_i\cap\mathcal{X}_r$;\\
            $\tilde{\boldsymbol{V}}-=\boldsymbol{v}_i$; $\tilde{\boldsymbol{P}}-=\boldsymbol{p}_i$;\\
            $\boldsymbol{s}^{p}_{i}$, $\boldsymbol{s}^{v}_{i}$=$F(\tilde{\boldsymbol{P}}, \tilde{\boldsymbol{V}}, \mathcal{A}_i^{'})$;\\
            $\boldsymbol{s}^{p}_{i} =  \boldsymbol{s}^{p}_{i} - \min{\boldsymbol{s}^{p}_{i}} + 1$ \\
            Update $\boldsymbol{p}_i$ by Eq. (\ref{eqn-2-21});\\
            Update $\boldsymbol{v}_i$ by Eq. (\ref{eqn-2-25});\\
             $\tilde{\boldsymbol{V}}+=\boldsymbol{v}_i$; $\tilde{\boldsymbol{P}}+=\boldsymbol{p}_i$;
            }  
         }
	}
        \KwOut {Fuzzy partition matrix $\boldsymbol{P}$}
\end{algorithm}

\subsubsection{Time Complexity}
It takes $\mathcal{O}(nc)$ time to calculate the variables $\boldsymbol{s}^{p}_{i}$ and $\boldsymbol{s}^{v}_{i}$ without any acceleration algorithm. Therefore, the time complexity of GPAC is $\mathcal{O}(cn^2)$. By introducing mini-batch clustering and aggregation acceleration strategies, GPAC requires approximately $n_b$ addition operations to compute clustering scores. Hence, the time complexity of the core GPAC optimization program is $\mathcal{O}(n_bn)$, where $n_b \ll n$. Besides, the complexity of similarity graph construction is also linear, the overall time complexity of the GPAC algorithm is linear.

\begin{table}[h]
	\centering
	\caption{Dataset settings for our experiments.}
	\label{tab1}
	\begin{tabular}{c|cccc}	
		 \hline
		Dataset   &Sample &Cluster  &Dimension & Data Type\\
		 \hline
            PENDIGITS \cite{alimoglu1997combining} & 10,992 & 10 & 16 &atifical feature\\
            ISOLET \cite{cole1990isolet} & 7,796 & 26 & 617  & raw data\\
            MNIST \cite{deng2012mnist} &70,000 & 10 & 784  & raw data\\
            COIL-100 \cite{nene1996columbia} &7,200 & 100 & 4,952  & raw data\\
            USPS \cite{hull1994database} &9,298 & 10 & 256  & raw data\\
            EMNIST  \cite{deng2012mnist} &131,600 & 47 & 784  & raw data\\
		CIFAR-10 \cite{krizhevsky2009learning}  & 60,000 &10  & 512  & deep feature\\
		CIFAR-20 \cite{krizhevsky2009learning}  & 60,000 &20  & 512  & deep feature\\
		STL-10 \cite{coates2011analysis}  & 13,000 &10  & 512  & deep feature\\
        IMAGENET \cite{deng2009imagenet}  & 1,281,167 &1,000  & 512  & deep feature\\
		 \hline
	\end{tabular}
\end{table}

\begin{table*}[t]
	\caption{Performance of Different Algorithms on Various Kinds of Raw Data.}
	\label{tab2}
	\centering
 \resizebox{\linewidth}{!}{
		\begin{tabular}{ll |ccc| ccc| ccc}
			\hline
			\multicolumn{2}{c}{Dataset} \vline
			&\multicolumn{3}{c}{{PENDIGITS}} \vline
			&\multicolumn{3}{c}{{ISOLET}}\vline
			&\multicolumn{3}{c}{{MNIST-FULL}}\\
            \hline
		  \multicolumn{2}{c}{Metric} \vline &NMI&ACC&ARI &NMI&ACC&ARI &NMI&ACC&ARI\\
    
		\hline	
                \multirow{2}{*}{KM} &Avg &{0.680}&{0.699}&{0.553} 
			&{0.726}&{0.553}&{0.486} 
			&{0.497}&{0.546}&{0.376}\\
   
                &($\pm$Std) &{($\pm$0.011)}&{($\pm$0.051)}&{($\pm$0.034)} 
			&{($\pm$0.010)}&{($\pm$0.027)}&{($\pm$0.025)} 
			&{($\pm$0.014)}&{($\pm$0.029)}&{($\pm$0.020)}\\
   
            \hline	
                \multirow{2}{*}{FCM} 
                &Avg &0.684 &0.683 &0.561 
			&0.725 &0.535 &0.479
			&0.489 &0.546 &0.370 \\
   
                &($\pm$Std) &{($\pm$0.013)}&{($\pm$0.013)}&{($\pm$0.038)} 
			&{($\pm$0.007)}&{($\pm$0.024)}&{($\pm$0.019)} 
			&{($\pm$0.011)}&{($\pm$0.017)}&{($\pm$0.013)}\\
            \hline	
                \multirow{2}{*}{SPKM} 
                 &Avg &0.686&0.712&0.562 
                 &0.727&0.550&0.486 
                 &0.530&0.548 &0.399\\
                &($\pm$Std) &($\pm$1.1e-2)&($\pm$4.3e-2)
                &($\pm$0.031) &{($\pm$0.008)}&{($\pm$0.026)}
                &{($\pm$0.016)} &($\pm$0.013) &($\pm$0.024)&($\pm$0.019)\\
             \hline	
                \multirow{2}{*}{SC} 
                &Avg &0.729 &0.714 &  0.559
                    &0.697 &0.527 &0.445
                    &0.470 &0.555 &0.365\\
                &($\pm$Std) 
                    &($\pm$0.010)&($\pm$0.014)&($\pm$0.013)
                    &($\pm$0.007)&($\pm$0.018)&($\pm$0.015)
                    &($\pm$0.006)&($\pm$0.016)&($\pm$0.014)\\
             \hline	
                \multirow{2}{*}{RCC}
                &Avg &\textbf{0.850} &- &0.765
                &0.691 &- &0.408
                &0.844 &- &0.733
                \\ 
                &($\pm$Std) 
                    &($\pm$0.000)&($\pm$0.000)&($\pm$0.000)
                    &($\pm$0.000)&($\pm$0.000)&($\pm$0.000)
                    &($\pm$0.000)&($\pm$0.000)&($\pm$0.000)\\
             \hline	
                \multirow{2}{*}{FINCH}
                &Avg &0.725 &- &0.501
                &0.723 &- &0.418
                &0.726 &- &0.548
                \\ 
                &($\pm$Std) 
                    &($\pm$0.000)&($\pm$0.000)&($\pm$0.000)
                    &($\pm$0.000)&($\pm$0.000)&($\pm$0.000)
                    &($\pm$0.000)&($\pm$0.000)&($\pm$0.000)\\        
            \hline	
                \multirow{2}{*}{Self-CSC} 
                &Avg &0.823 &0.837 &0.723  &0.773 &0.625 &0.535 &0.807 &0.851 &0.755\\
                &($\pm$Std) 
                    &($\pm$0.026)&($\pm$0.037)& ($\pm$0.046)
                    &($\pm$0.007)&($\pm$0.011)& ($\pm$0.012)
                    &($\pm$0.016)&($\pm$0.036)& ($\pm$0.040)
                    \\
            \hline	
                \multirow{2}{*}{K-sums} 
                &Avg &0.657 &0.726 & 0.571
                &0.728 &0.615 &0.524
                &0.509 &0.585 &0.420\\
                &($\pm$Std) 
                    &($\pm$0.000)&($\pm$0.000)& ($\pm$0.000)
                    &($\pm$0.008)&($\pm$0.016)& ($\pm$0.014)
                    &($\pm$0.010)&($\pm$0.013)& ($\pm$0.014)
                    \\
            \hline	
                \multirow{2}{*}{PAC}
            &Avg &0.702 &0.773 &0.616
			 &{0.734}&{0.617}&{0.530} 
			 &0.495 &0.607 &0.409\\
            &($\pm$Std) 
            &{($\pm$0.004)}&{($\pm$0.012)}&{($\pm$0.004)} 
			&{($\pm$0.004)}&{($\pm$0.001)}&{($\pm$0.002)} 
			&{($\pm$0.005)}&{($\pm$0.004)}&{($\pm$0.005)}\\
            \hline	
		\multirow{2}{*}{GPAC} &Avg &\textbf{0.850}&\textbf{0.881}&\textbf{0.789} 
			&\textbf{0.776}&\textbf{0.630}&\textbf{0.581} 
			&\textbf{0.852}&\textbf{0.875}&\textbf{0.821}\\
                &($\pm$Std) &($\pm$0.005) &($\pm$0.004) &($\pm$0.006)
                &($\pm$0.001) &($\pm$0.003) &($\pm$0.002)
                &($\pm$0.001) &($\pm$0.000) &($\pm$0.001)\\
		\hline
		\end{tabular}
  }
\end{table*}
\begin{table*}[t]
	\caption{Performance of Different Algorithms on Various Kinds of Raw Data. (O.M.: out of memory)}
	\label{tab3}
	\centering
	\resizebox{\linewidth}{!}{
		\begin{tabular}{ll |ccc| ccc| ccc}
			\hline
			\multicolumn{2}{c}{Dataset} \vline
			&\multicolumn{3}{c}{{COIL-100}} \vline
			&\multicolumn{3}{c}{{USPS}}\vline
			&\multicolumn{3}{c}{{EMNIST}}\\
            \hline
		  \multicolumn{2}{c}{Metric} \vline &NMI&ACC&ARI &NMI&ACC&ARI &NMI&ACC&ARI\\
    
		\hline	
                \multirow{2}{*}{KM} &Avg 
                &0.516 &0.516 &0.359
                &0.607&0.643&0.524
                &{0.410}&{0.318}&{0.173}\\
                &($\pm$Std)
                &($\pm$0.013)&($\pm$0.032)&($\pm$0.018)
                 &($\pm$0.004)&($\pm$0.021)&($\pm$0.009)
                &{($\pm$0.003)}&{($\pm$0.006)}&{($\pm$0.003)} \\
            \hline	
                \multirow{2}{*}{FCM} 
                &Avg  &0.510 &0.513 &0.356
                &0.623&0.639&0.532
                &0.410 &0.319 & 0.174
                \\
                 
                &($\pm$Std)  &{($\pm$0.011)}&{($\pm$0.035)}&{($\pm$0.016)}
                &{($\pm$0.013)}&{($\pm$0.022)}&{($\pm$0.022)}
                &{($\pm$0.002)}&{($\pm$0.004)}&{($\pm$0.002)}
                \\
            \hline	
                \multirow{2}{*}{SPKM} 
                &Avg &0.595 &0.526 &0.407
                    &0.659 &0.694 & 0.583 
                    &0.417&0.319&0.177\\
                 &($\pm$Std) &{($\pm$0.016)}&{($\pm$0.028)}&{($\pm$0.020)}
                 &{($\pm$0.013)}&{($\pm$0.043)}&{($\pm$0.023)}
                 &{($\pm$0.004)}&{($\pm$0.010)}&{($\pm$0.003)}
                \\
             \hline	
                \multirow{2}{*}{SC} 
                &Avg &0.791 &0.517 & 0.373  
                &0.489 &0.561 &0.295
                &0.274 &0.269 &0.050 \\
                &($\pm$Std) &{($\pm$0.000)}&{($\pm$0.000)}&{($\pm$0.000)}
                &{($\pm$0.002)}&{($\pm$0.002)}&{($\pm$0.004)}
                  &{($\pm$0.021)}&{($\pm$0.005)}&{($\pm$0.006)}
                    \\
             \hline	
                \multirow{2}{*}{RCC} 
                &Avg &\textbf{0.975} &- &\textbf{0.830}
                    &0.655  &- &0.398
                    &O.M.&O.M.&O.M.\\
                &($\pm$Std) &{($\pm$0.000)}&{($\pm$0.000)}&{($\pm$0.000)} 
                  &{($\pm$0.000)}&{($\pm$0.000)}&{($\pm$0.000)} 
                   &O.M.&O.M.&O.M.
                    \\
            \hline	
                \multirow{2}{*}{FINCH}
                &Avg &0.849 &- &0.523
                &0.777 &- &0.617
                &0.574 &- &0.268
                \\ 
                &($\pm$Std) 
                    &($\pm$0.000)&($\pm$0.000)&($\pm$0.000)
                    &($\pm$0.000)&($\pm$0.000)&($\pm$0.000)
                    &($\pm$0.000)&($\pm$0.000)&($\pm$0.000)\\        
            \hline	
                \multirow{2}{*}{Self-CSC} 
                &Avg &0.882 &0.715  &0.602 
                &0.726  &0.754 &0.614  
                &0.586 &0.497&0.322\\
                &($\pm$Std) &($\pm$0.005) &($\pm$0.018) &($\pm$0.014)
                    &($\pm$0.021)&($\pm$0.034)& ($\pm$0.043)
                     &($\pm$0.003)&($\pm$0.008)& ($\pm$0.006)
                    \\
            \hline	
                \multirow{2}{*}{K-sums} 
                &Avg &0.817 &0.642 & 0.577
                &0.638 &0.718 &0.577
                &0.441 &0.343 &0.198\\
                &($\pm$Std) 
                    &($\pm$0.003)&($\pm$0.007)& ($\pm$0.007)
                    &($\pm$0.005)&($\pm$0.008)& ($\pm$0.006)
                    &($\pm$0.003)&($\pm$0.007)& ($\pm$0.004)
                    \\
            \hline	
                \multirow{2}{*}{PAC}
            &Avg 
             &0.838 &0.660 &0.616
			 &0.589 &0.657 &0.519
			 &{0.423}&{0.507}&{0.336}\\
   
            &($\pm$Std) &{($\pm$0.003)}&{($\pm$0.008)}&{($\pm$0.008)} 
			&{($\pm$0.000)}&{($\pm$0.000)}&{($\pm$0.000)} 
			&{($\pm$0.010)}&{($\pm$0.012)}&{($\pm$0.015)}\\
            \hline	
		\multirow{2}{*}{GPAC}
                &Avg&{0.868}&\textbf{0.717}&{0.656} 
			&\textbf{0.836}&\textbf{0.808}&\textbf{0.770} 
			&\textbf{0.648}&\textbf{0.547}&\textbf{0.417}\\
   
                &($\pm$Std) &($\pm$0.002) &($\pm$0.008) &($\pm$0.006)
                &($\pm$0.001) &($\pm$0.001) &($\pm$0.001)
                &($\pm$0.003) &($\pm$0.009) &($\pm$0.004)\\
                
		\hline
		\end{tabular}
 }
\end{table*}

\section{Experiment}
We evaluate GPAC on multiple clustering tasks to validate the superiority of our method. Our evaluation covers various aspects, including clustering performance, parameter sensitivity, and ablation studies. We employ three evaluation metrics: Clustering Accuracy (ACC), Normalized Mutual Information (NMI), and Adjusted Rand Index (ARI) to assess the effectiveness of GPAC.

\subsubsection{Dataset}
Nine real-world datasets are used to evaluate the clustering ability of GPAC. These datasets cover various types, including toy image data (MNIST \cite{deng2012mnist}, USPS \cite{hull1994database}, COIL-100 \cite{nene1996columbia}, USPS, EMNIST \cite{deng2012mnist}), voice data (Isolet \cite{cole1990isolet}), artificial feature data (Pendigits \cite{alimoglu1997combining}), and deep feature data (CIFAR-10 \cite{krizhevsky2009learning}, CIFAR-20 \cite{krizhevsky2009learning}, STL-10 \cite{coates2011analysis}). The details of these datasets, including the number of samples, features, types, and categories, are summarized in Table \ref{tab1}.

\begin{table*}[t]
	\caption{Performance of Different Algorithms on Contrastive Pre-trained Deep Features.}
	\label{tab4}
	\centering
	\resizebox{\linewidth}{!}{
		\begin{tabular}{ll |ccc| ccc| ccc}
			\hline
			\multicolumn{2}{c}{Dataset}\vline
			&\multicolumn{3}{c}{{CIFAR-10}} \vline
			&\multicolumn{3}{c}{{CIFAR-20}}\vline
			&\multicolumn{3}{c}{{STL-10}}\\
            \hline
		  \multicolumn{2}{c}{Metric}\vline&NMI&ACC&ARI &NMI&ACC&ARI  &NMI&ACC&ARI\\
            \hline	
                \multirow{2}{*}{KM} 
                &Avg &{0.750}&{0.770}&{0.639} 
			&{0.471}&{0.427}&{0.261} 
			&{0.651}&{0.668}&{0.503}\\
   
                &($\pm$Std)&{($\pm$0.022)}&{($\pm$0.044)}&{($\pm$0.037)} 
			&{($\pm$0.020)}&{($\pm$0.027)}&{($\pm$0.021)} 
			&{($\pm$0.028)}&{($\pm$0.045)}&{($\pm$0.041)}\\
            \hline	
                \multirow{2}{*}{FCM} 
                &Avg &{0.754}&{0.784}&{0.651} 
			&{0.464}&{0.427}&{0.261} 
			&{0.682}&{0.683}&{0.511}\\
   
                &($\pm$Std) &{($\pm$0.014)}&{($\pm$0.027)}&{($\pm$0.026)} 
			&{($\pm$0.007)}&{($\pm$0.013)}&{($\pm$0.010)} 
			&{($\pm$0.029)}&{($\pm$0.022)}&{($\pm$0.032)}\\
   
            \hline	
                \multirow{2}{*}{SPKM} 
                &Avg &0.753 &0.781 & 0.646
                    &0.471 &0.425 &0.259
                    &0.659 &0.680 &0.516\\
                &($\pm$Std) 
                    &($\pm$0.018)&($\pm$0.043)&($\pm$0.037)
                    &($\pm$0.013)&($\pm$0.018)&($\pm$0.014)
                    &($\pm$0.022)&($\pm$0.043)&($\pm$0.036)\\
             \hline	
                 \multirow{2}{*}{SC} 
                &Avg &0.714 &0.780 &0.659 
                    & & &
                    &0.609 &0.652 &0.462\\
                &($\pm$Std) 
                    &($\pm$0.018)&($\pm$0.043)&($\pm$0.037)
                    &($\pm$0.013)&($\pm$0.018)&($\pm$0.014)
                    &($\pm$0.008)&($\pm$0.022)&($\pm$0.024)\\
             \hline	
                \multirow{2}{*}{FINCH}
                &Avg &0.728 &- &0.545
                &0.478 &- &0.204
                &0.592 &- &0.384
                \\ 
                &($\pm$Std) 
                    &($\pm$0.000)&($\pm$0.000)&($\pm$0.000)
                    &($\pm$0.000)&($\pm$0.000)&($\pm$0.000)
                    &($\pm$0.000)&($\pm$0.000)&($\pm$0.000)\\        
             \hline	
                \multirow{2}{*}{Self-CSC} 
                &Avg &0.790 &0.815 & 0.698
                    &0.509 &0.468 & 0.247
                    &0.718 &0.754 &0.570\\
                &($\pm$Std) 
                    &($\pm$0.019)&($\pm$0.022)&($\pm$0.027)
                    &($\pm$0.012)&($\pm$0.016)&($\pm$0.022)
                    &($\pm$0.016)&($\pm$0.034)&($\pm$0.042)\\
            \hline	
                \multirow{2}{*}{K-sums} 
                &Avg &0.714 &0.780 & 0.659
                &0.491 &0.471 &0.323
                &0.629 &0.697 &0.547\\
                &($\pm$Std) 
                    &($\pm$0.016)&($\pm$0.020)& ($\pm$0.029)
                    &($\pm$0.007)&($\pm$0.018)& ($\pm$0.011)
                    &($\pm$0.028)&($\pm$0.059)& ($\pm$0.047)
                    \\
             \hline
                 \multirow{2}{*}{SCAN} 
                &Avg &0.787 &0.876 & 0.758
                    &0.468 &0.459 &0.301
                    &0.680 &0.767 &0.616\\
                &($\pm$Std) 
                    &($\pm$0.005)&($\pm$0.004)&($\pm$0.007)
                    &($\pm$0.013)&($\pm$0.027)&($\pm$0.021)
                    &($\pm$0.012)&($\pm$0.019)&($\pm$0.018)\\    
		      \hline	
                \multirow{2}{*}{PAC} 
                &Avg &{0.759}&{0.871}&{0.763} 
			     &{0.464}&{0.439}&{0.295} 
			     &{0.662}&{0.753}&{0.599}\\
   
                &($\pm$Std)&{($\pm$0.002)}&{($\pm$0.004)}&{($\pm$0.004)} 
			&{($\pm$0.005)}&{($\pm$0.006)}&{($\pm$0.003} 
			&{($\pm$1.4e-2)}&{($\pm$0.029)}&{($\pm$0.024)}\\
   
            \hline	
		      \multirow{2}{*}{GPAC} 
                &Avg &\textbf{0.832}&\textbf{0.904}&\textbf{0.814} 
			&\textbf{0.514}&\textbf{0.486}&\textbf{0.335} 
			&\textbf{0.779}&\textbf{0.872}&\textbf{0.749}\\
   
            &($\pm$Std)
            &{($\pm$0.000)}&{($\pm$0.000)}&{($\pm$0.000)} 
			&{($\pm$0.008)}&{($\pm$0.013)}&{($\pm$0.010)} 
			&{($\pm$0.000)}&{($\pm$0.000)}&{($\pm$0.001)}\\
		\hline
		\end{tabular}
}
\end{table*}

\subsubsection{Compared Methods}
To demonstrate the superiority of the proposed method, we evaluate GPAC against nine state-of-the-art clustering methods. These algorithms represent the latest advancements in clustering, each with unique characteristics and strengths. They are briefly described as follows:
\begin{itemize}
    \item K-means (KM) \cite{wang2014optimized}: A refined version of the traditional K-means by using an improved initialization strategy for cluster centroids. 
    
    \item Fuzzy C-means (FCM) \cite{bezdek1984fcm}: The most widely used fuzzy clustering methods.
    
    \item Spherical K-means (SPKM) \cite{hornik2012spherical}: A variation of the K-means algorithm that utilizes cosine dissimilarities instead of Euclidean distances.

    \item Spectral Clustering (SC) \cite{von2007tutorial}: A powerful graph-based clustering method that uses the eigenvalues of the similarity matrix to partition the data. 

    \item Robust Continuous Clustering (RCC) \cite{shah2017robust}: 
    A graph clustering method based on robust statistics. This method provides more stable clustering results, especially in the presence of corrupted or noisy data.

    \item First Integer Neighbor Clustering Hierarchy Clustering (FINCH) \cite{sarfraz2019efficient}: A hierarchical agglomerative clustering algorithm focuses on first-neighbor relations. 

    \item Centerless Clustering (K-sums) \cite{pei2022centerless}: 
    An efficient centerless clustering algorithm, which unifies the K-means and SC algorithms.

    \item Self-Constrained Spectral Clustering (Self-CSC) \cite{bai2022self}: A extension of spectral clustering algorithm based on self-constrained terms. 
\end{itemize}

\subsubsection{Parameter Settings}
The parameters involved in these algorithms are described below. In our experiments, the number of clusters, denoted by $c$, is set as prior knowledge, and it is assumed to be equal to the true number of data classes in each dataset.

For our GPAC, as discussed earlier, the following parameters are used in all experiments: the fuzzy exponent $m$ is set to 1.05, the number of neighbors $k$ is set to 10, the regularization parameters $\alpha$ are set to $1$.  The penalty hyperparamete $\beta$ increases in steps from 0 to 1. 

For FCM, the fuzzy weighting exponent $m$ is fixed at 1.05. In the case of graph-based clustering methods, such as Spectral Clustering, RCC, and Self-CSC, the $k$-nearest neighbor graph is used to represent the similarity between data points. The construction of the $k$-NN graph plays a crucial role in defining the local structure of the data and ensures that the graph captures the inherent relations.

All algorithms are initialized randomly and run 50 times to ensure robustness. For each run, the mean and variance of the clustering performance are reported to provide a comprehensive evaluation of each method.

\subsubsection{Deep Feature Acquisition}
The field of deep learning has experienced rapid growth in recent years. To further assess the performance of GPAC, we also evaluate its ability to process features learned through deep learning. We select contrastive learning \cite{chen2020simple} as the unsupervised feature extraction model and use SimCLR \cite{chen2020simple} to train a ResNet \cite{he2016deep}. The output of the final layer of ResNet is then used as the feature data to perform clustering.

\subsubsection{Main Results}
We first conduct experiments to test our GPAC on six real-world datasets. The evaluation results are shown in Table \ref{tab2} and Table \ref{tab3}. Obviously, our GPAC has better performance and stability on most datasets. Compared with center-based clustering (KM++, FCM, SPKM), GPAC has better performance on PENDIGITS, ISOLTE, COIL-100, USPS, and EMNIST. And GPAC significantly outperforms these algorithms in robustness. Compared with graph-based methods, although GPAC lacks stability, its performance is still far superior to these algorithms. 

Then we evaluate the performance of GPAC in deep feature datasets, which is demonstrated in Table \ref{tab4}. Deep feature is more abstract, and the data are more entangled. Therefore, the stability of previous algorithms on deep features is poor, which leads to performance degradation.  Especially in STL-10, contrastive learning can not distinguish cat and dog classes well, which leads to feature entanglement, and finally makes most clustering algorithms unable to perform effective clustering. Compared to them, GPAC performs best on deep datasets. We also introduced a deep clustering model SCAN in the comparison, which is the most famous three stage deep contrastive clustering method. We find that GPAC is even better than the deep model using various visual tricks on deep features. Compared with deep clustering methods, GPAC has fewer hyperparameters, stronger stability, faster computation, and better performance.

\begin{figure*}[t]
    \centering
    \scriptsize
    \begin{minipage}[b]{0.19\linewidth}
        \centering
        \includegraphics[width=\linewidth]{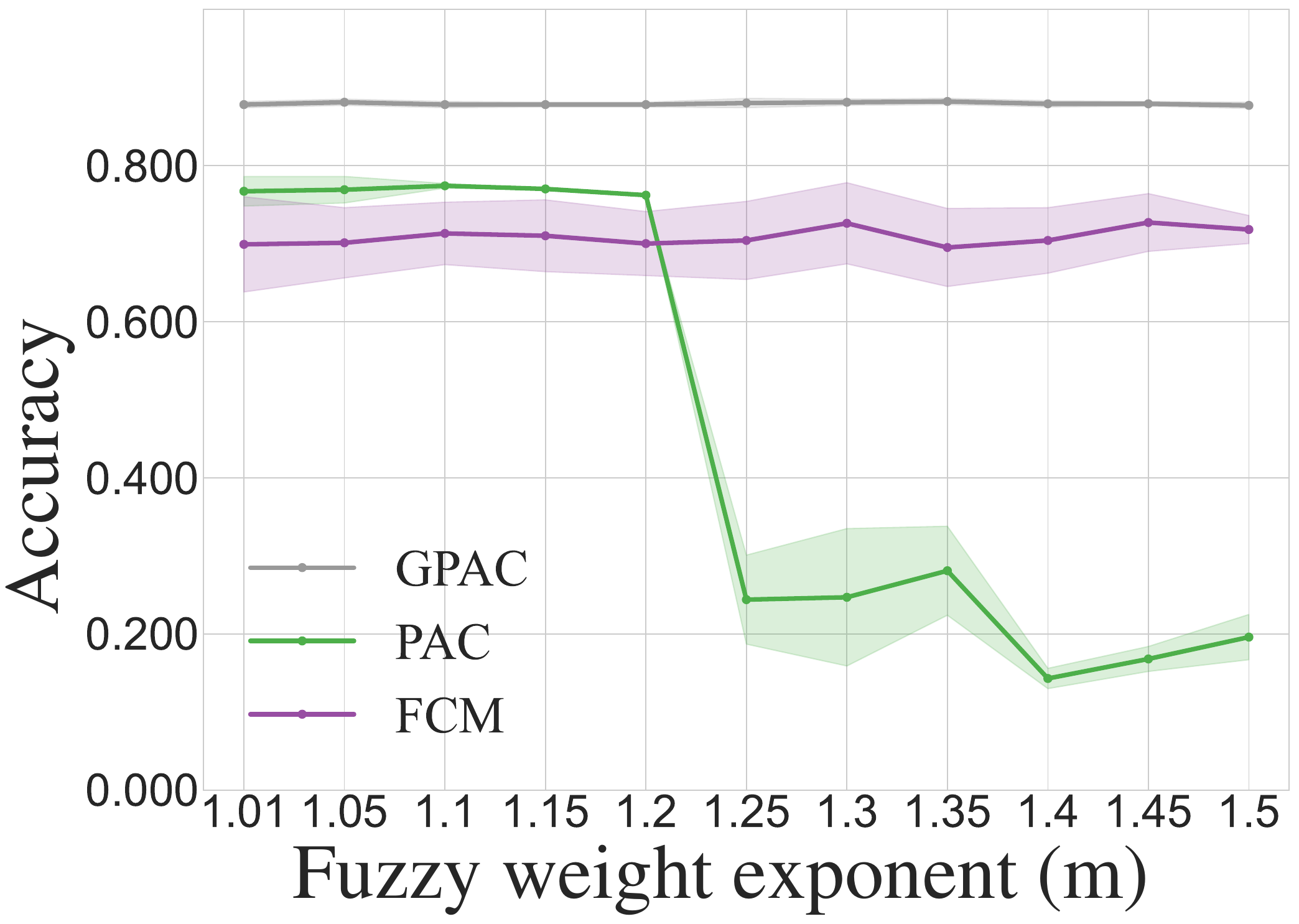}
        {(a) PENDIGITS} 
    \end{minipage} 
    \begin{minipage}[b]{0.19\linewidth}
        \centering
        \includegraphics[width=\linewidth]{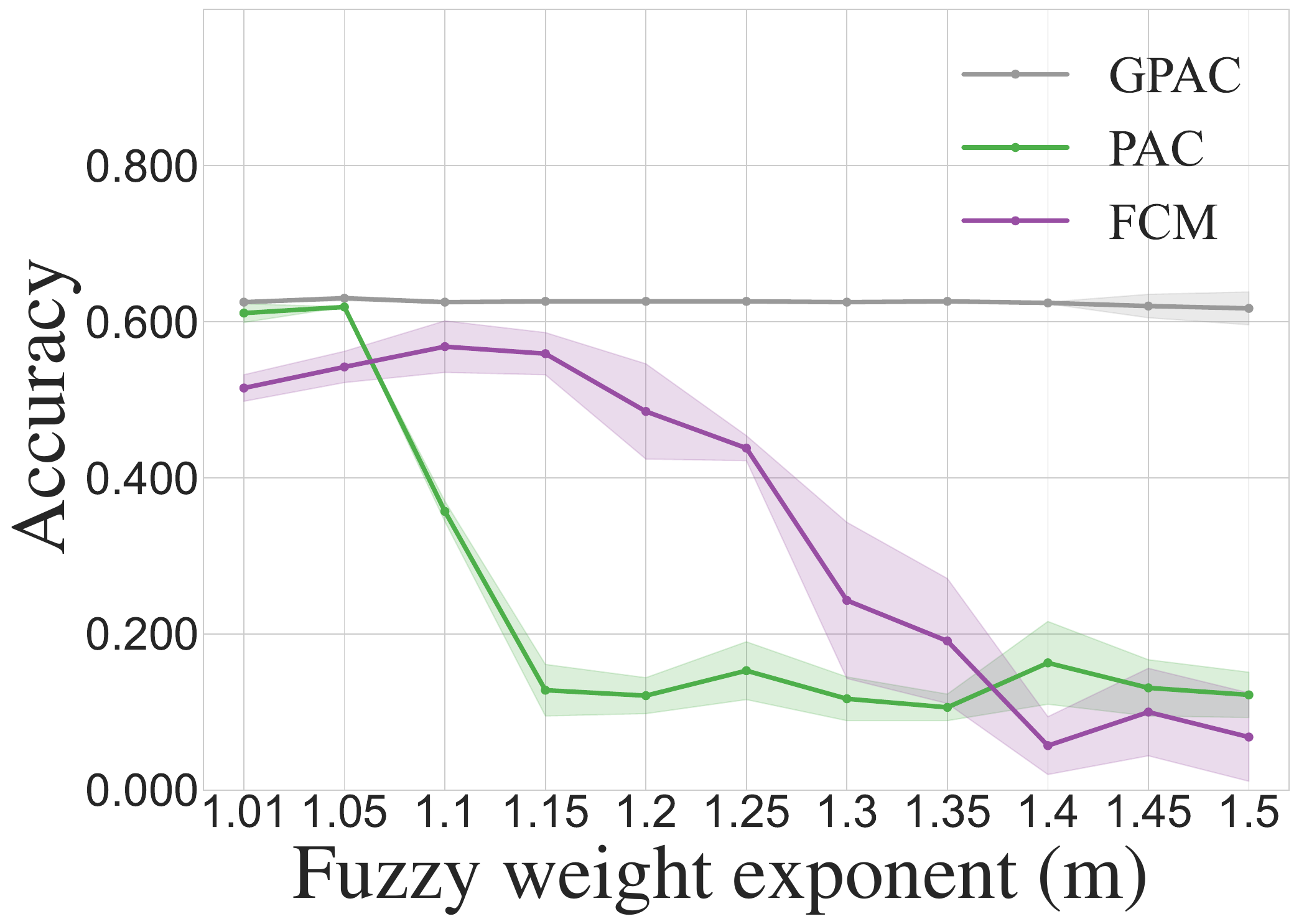}
        {(b) ISOLET} 
    \end{minipage} 
    \begin{minipage}[b]{0.19\linewidth}
        \centering
        \includegraphics[width=\linewidth]{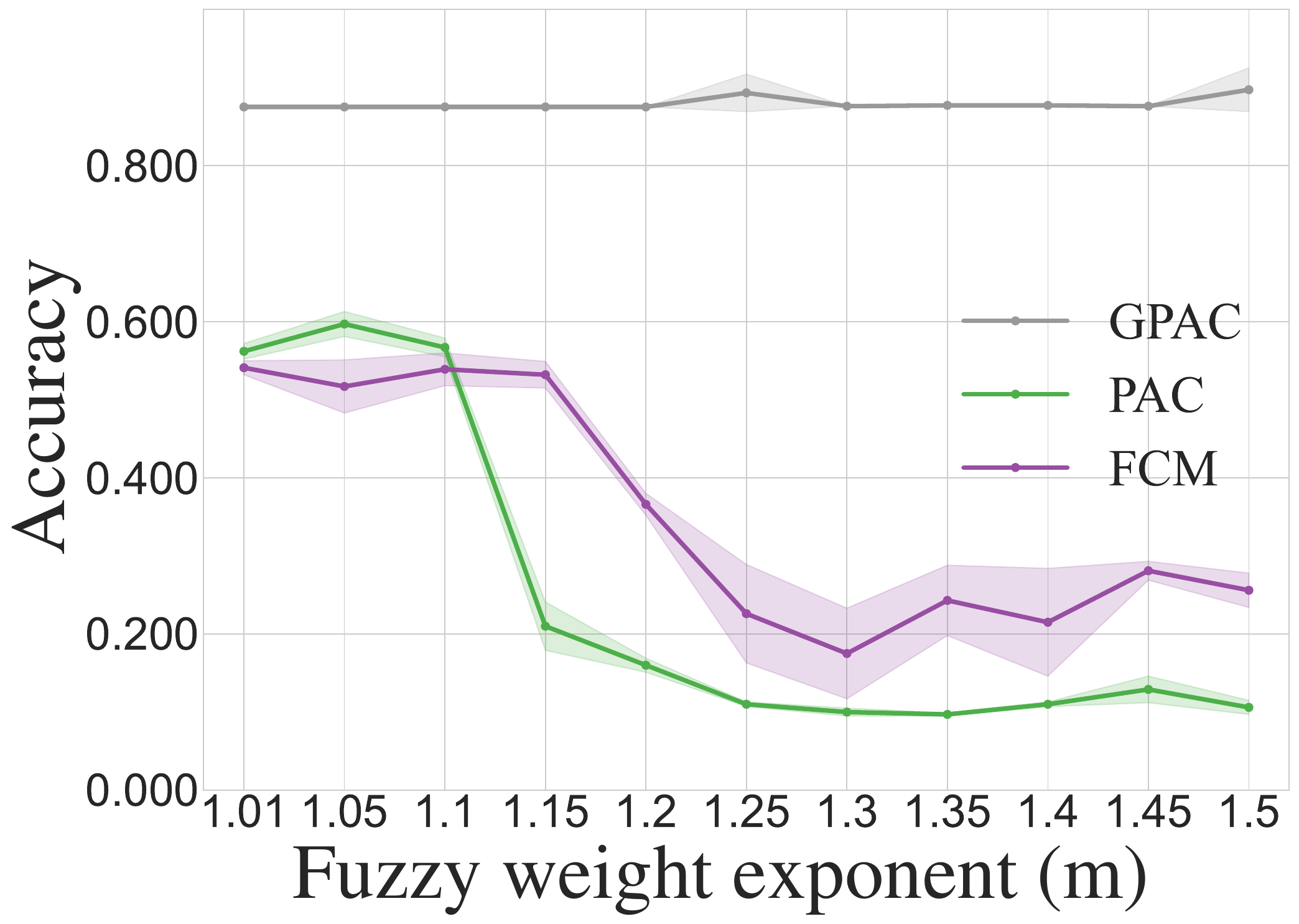}
        {(c) MNIST} 
    \end{minipage} 
    \begin{minipage}[b]{0.19\linewidth}
        \centering
        \includegraphics[width=\linewidth]{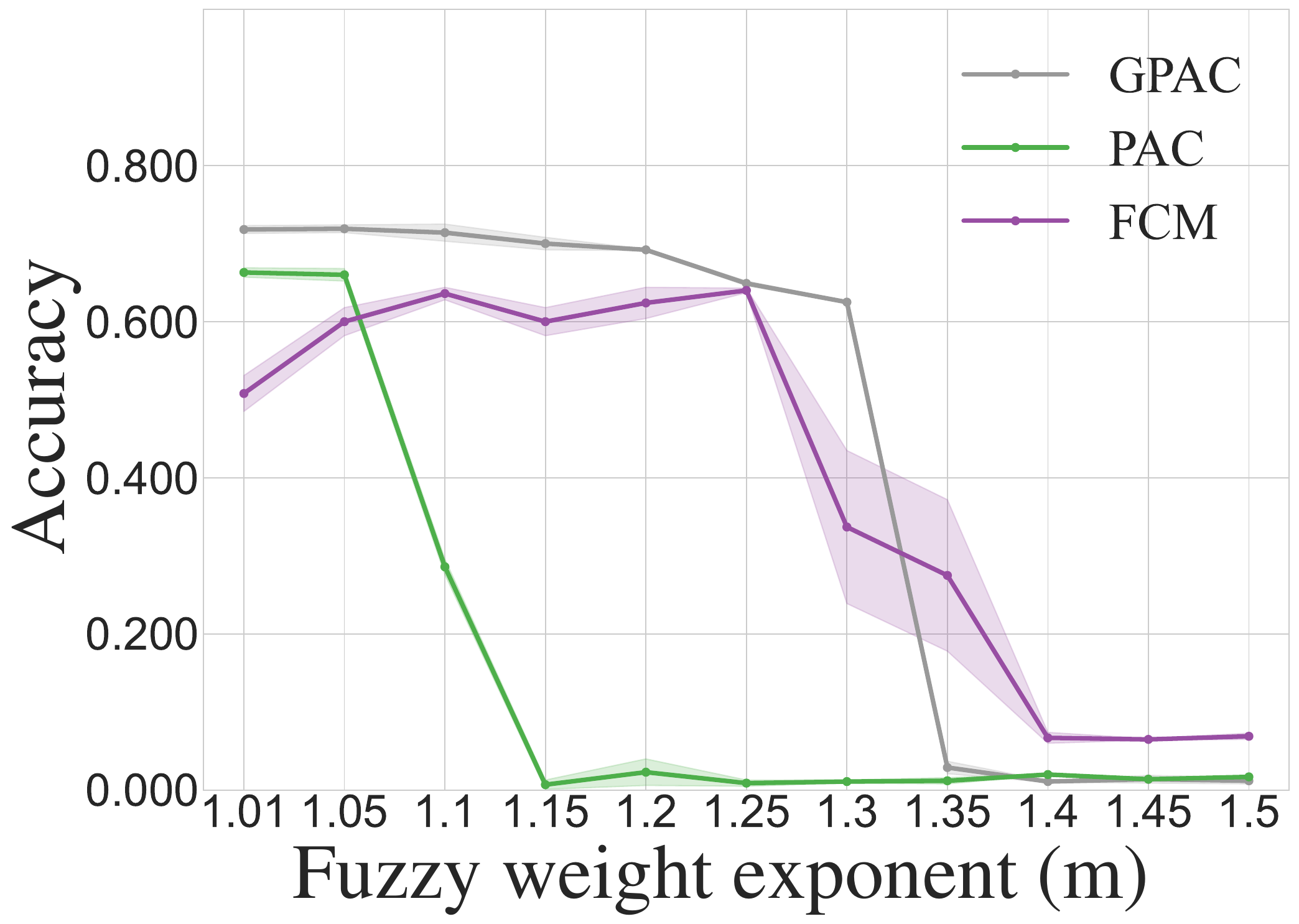}
        {(d) COIL-100} 
    \end{minipage} 
    \begin{minipage}[b]{0.19\linewidth}
        \centering
        \includegraphics[width=\linewidth]{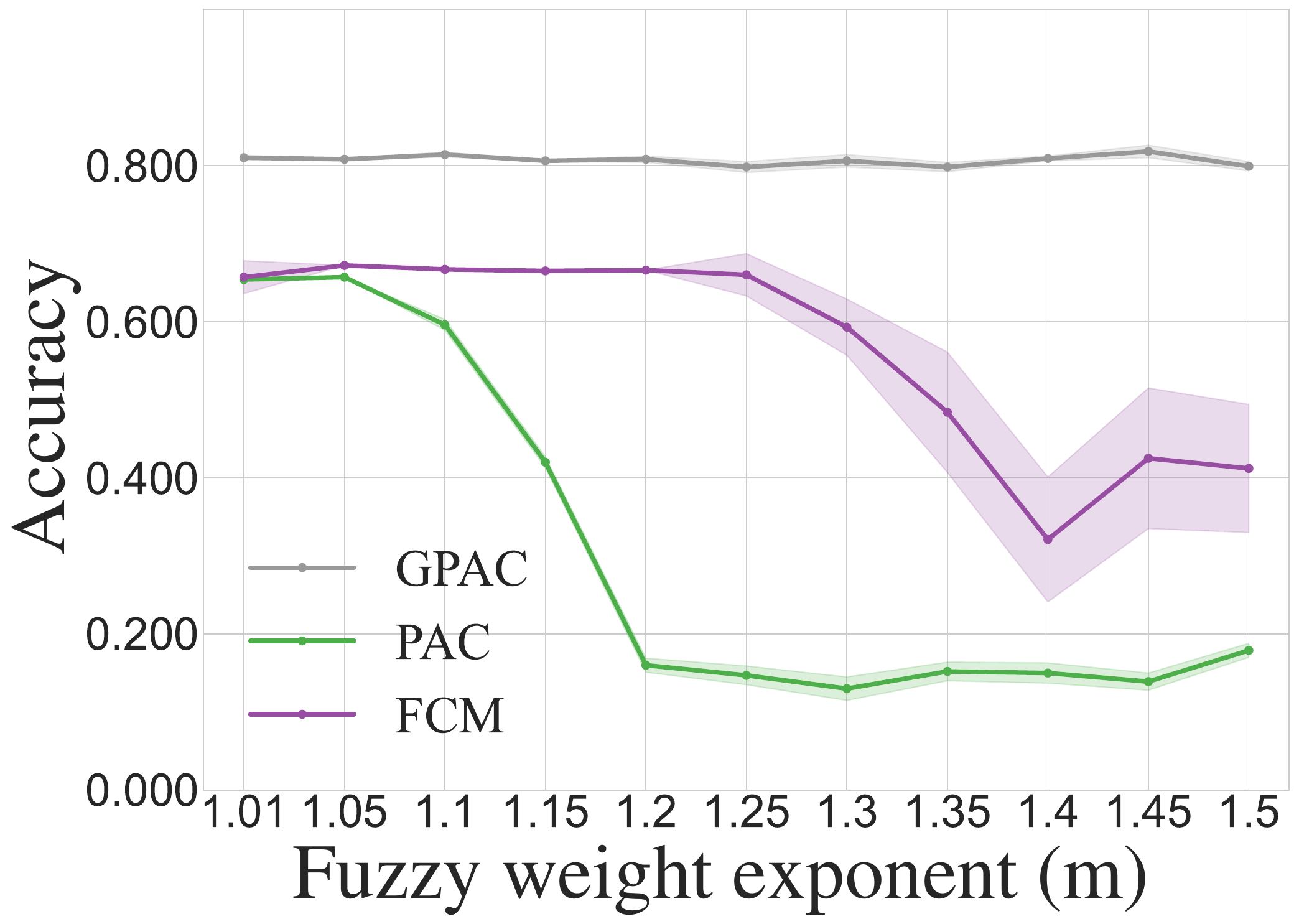}
        {(e) USPS} 
    \end{minipage} 
    
    \vspace{0.3cm} 
    \begin{minipage}[b]{0.19\linewidth}
        \centering
        \includegraphics[width=\linewidth]{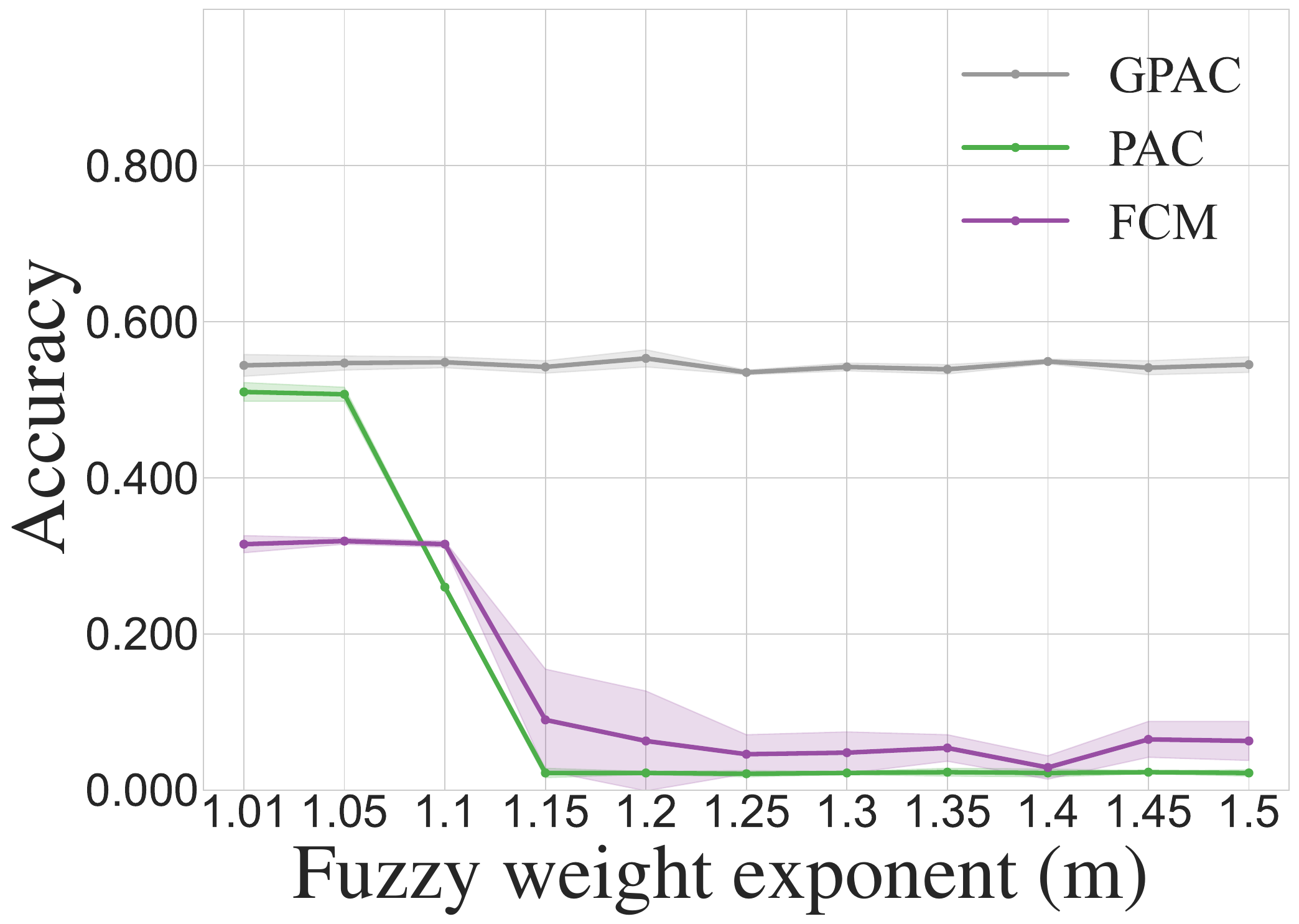}
        {(f) EMNIST} 
    \end{minipage} 
    \begin{minipage}[b]{0.19\linewidth}
        \centering
        \includegraphics[width=\linewidth]{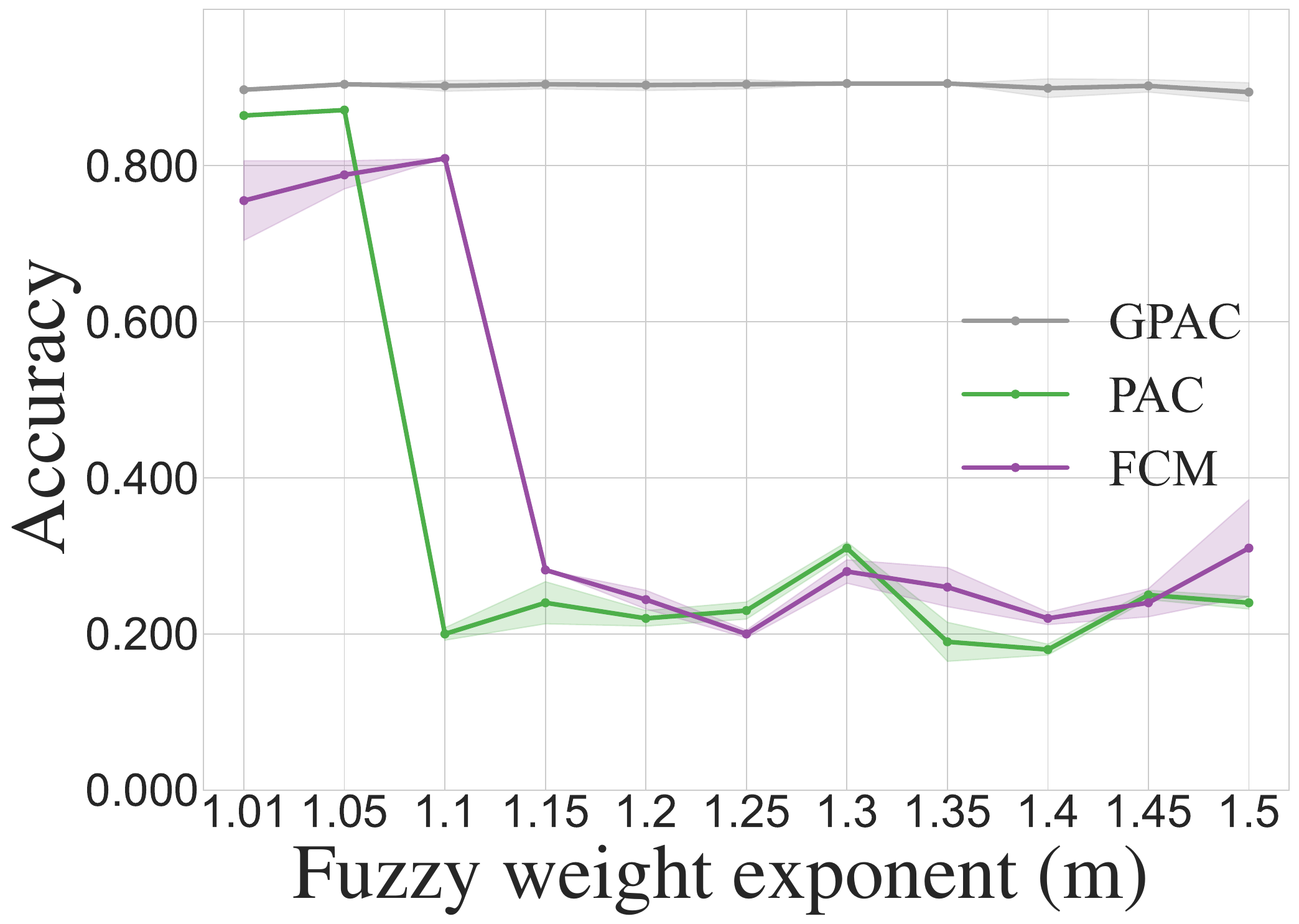}
        {(g) CIFAR-10} 
    \end{minipage} 
    \begin{minipage}[b]{0.19\linewidth}
        \centering
        \includegraphics[width=\linewidth]{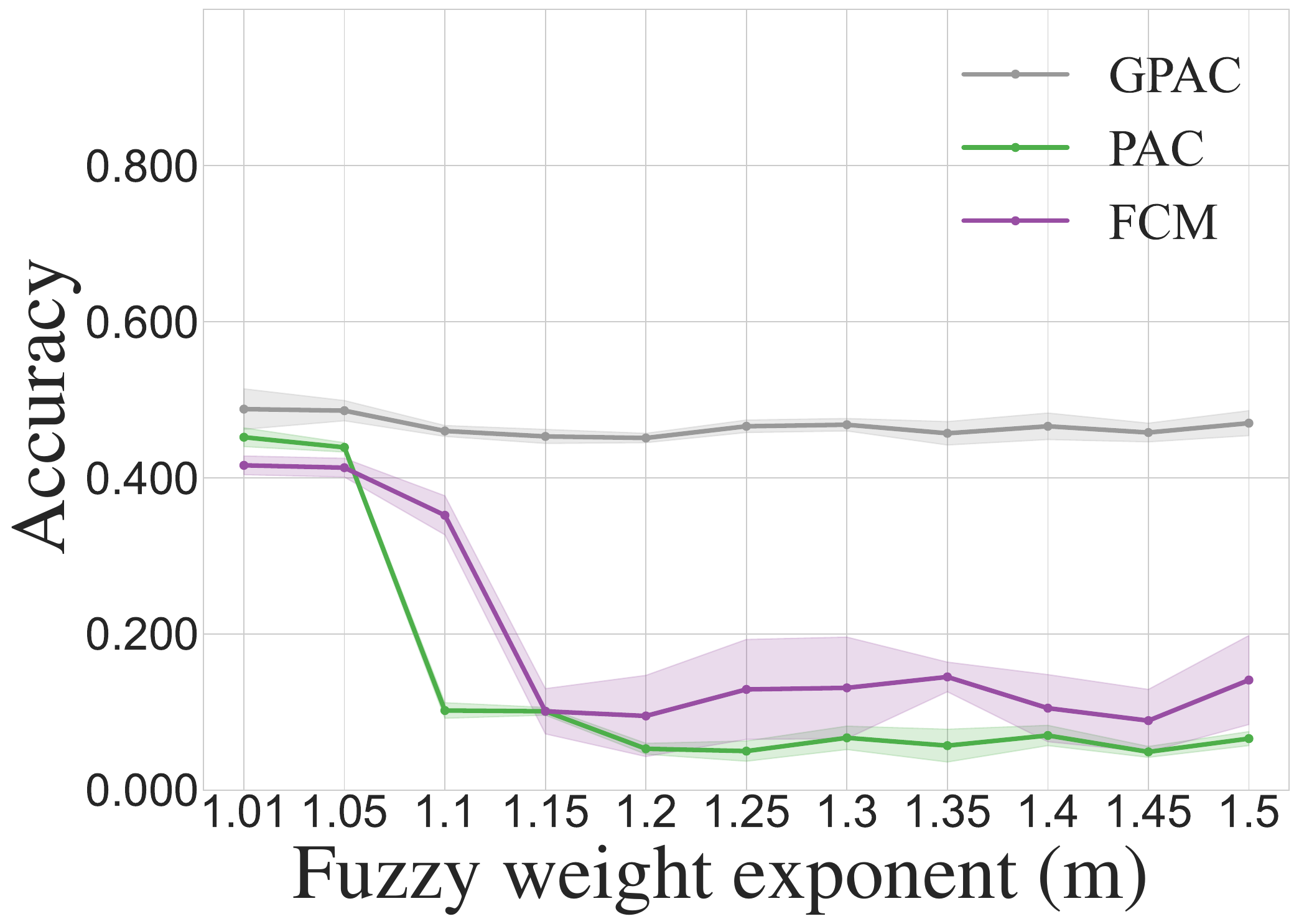}
        {(h) CIFAR-20} 
    \end{minipage} 
    \begin{minipage}[b]{0.19\linewidth}
        \centering
        \includegraphics[width=\linewidth]{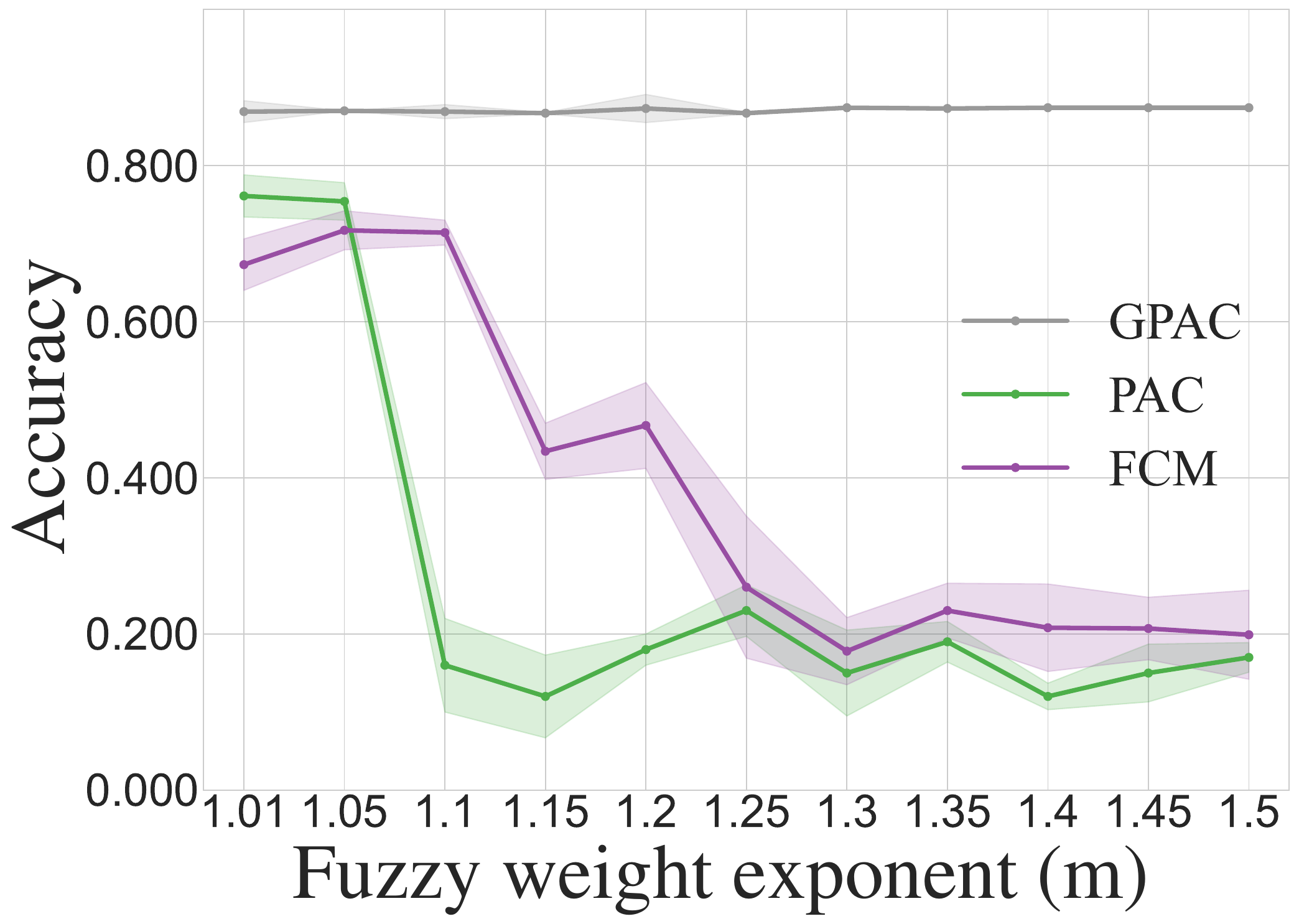}
        {(i) STL-10} 
    \end{minipage} 
    \begin{minipage}[b]{0.19\linewidth}
        \centering
        \includegraphics[width=\linewidth]{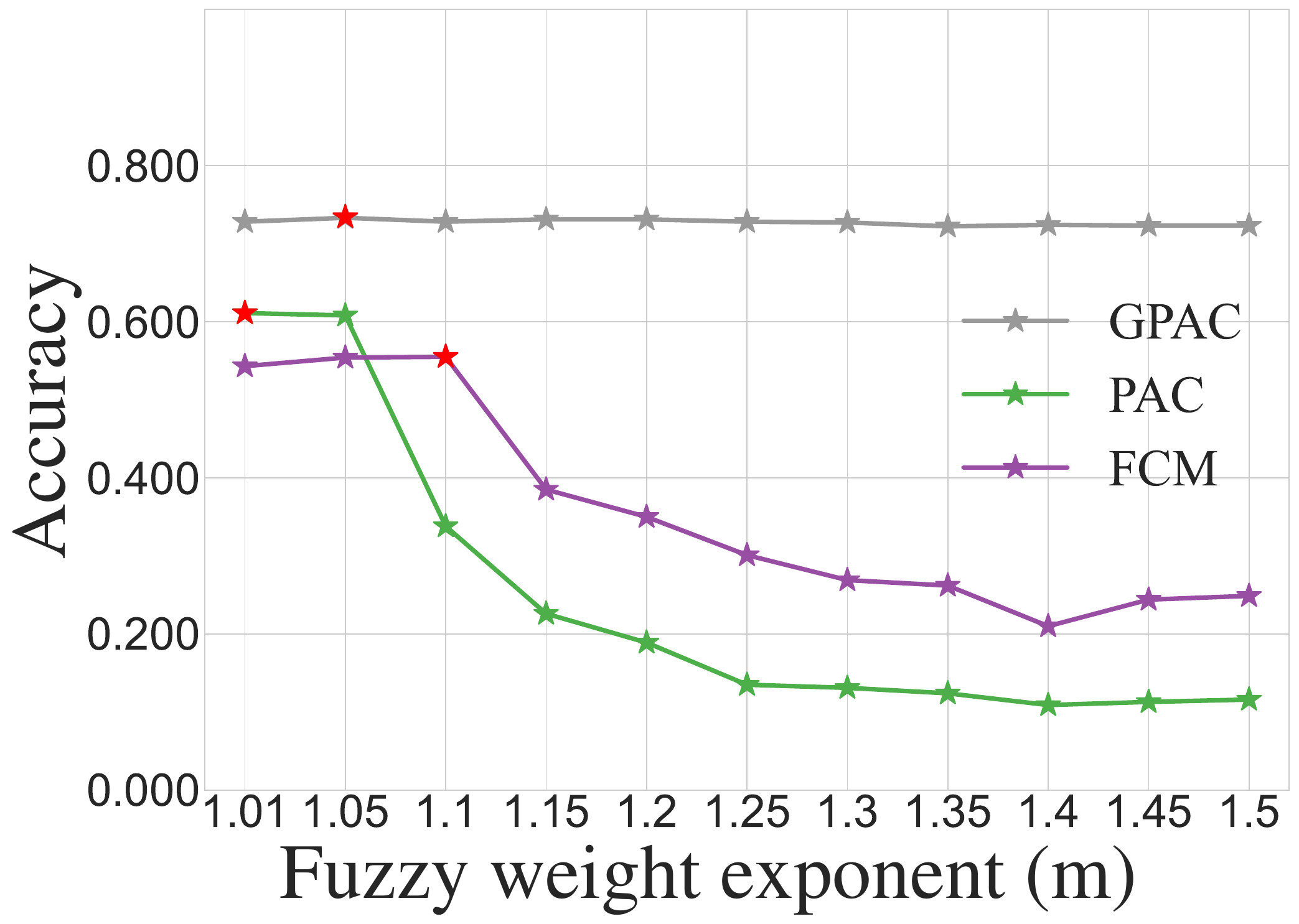}
        {(j) Mean of all results} 
    \end{minipage} 
    
    \caption{Accuracy comparison of GPAC, PAC, and FCM with different weighting exponent $m$.}
    \label{fig_2}
\end{figure*}

\begin{figure*}[t]
    \centering
    \scriptsize
    \begin{minipage}[b]{0.19\linewidth}
        \centering
        \includegraphics[width=\linewidth]{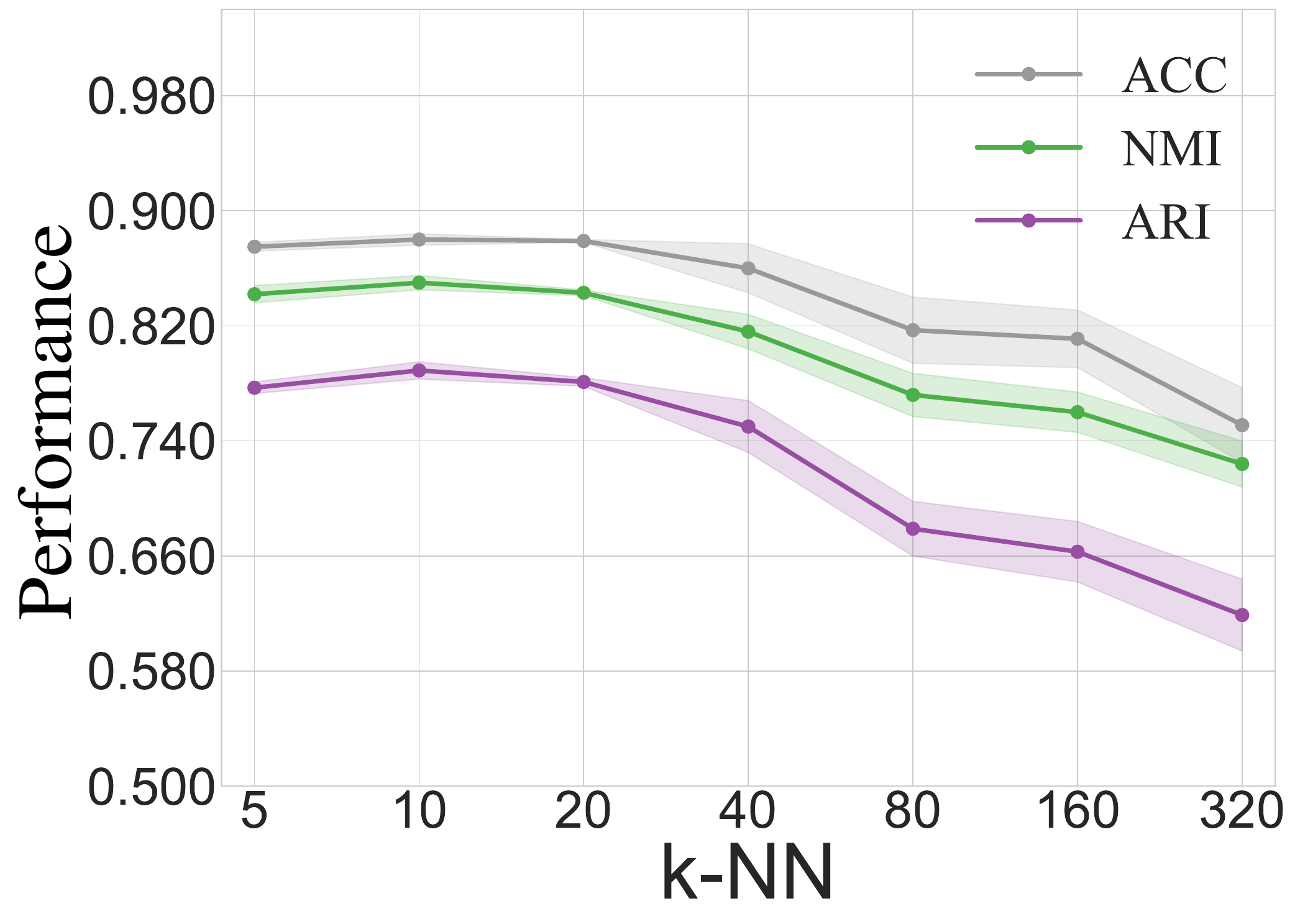}
        {(a) PENDIGITS} 
    \end{minipage} 
    \begin{minipage}[b]{0.19\linewidth}
        \centering
        \includegraphics[width=\linewidth]{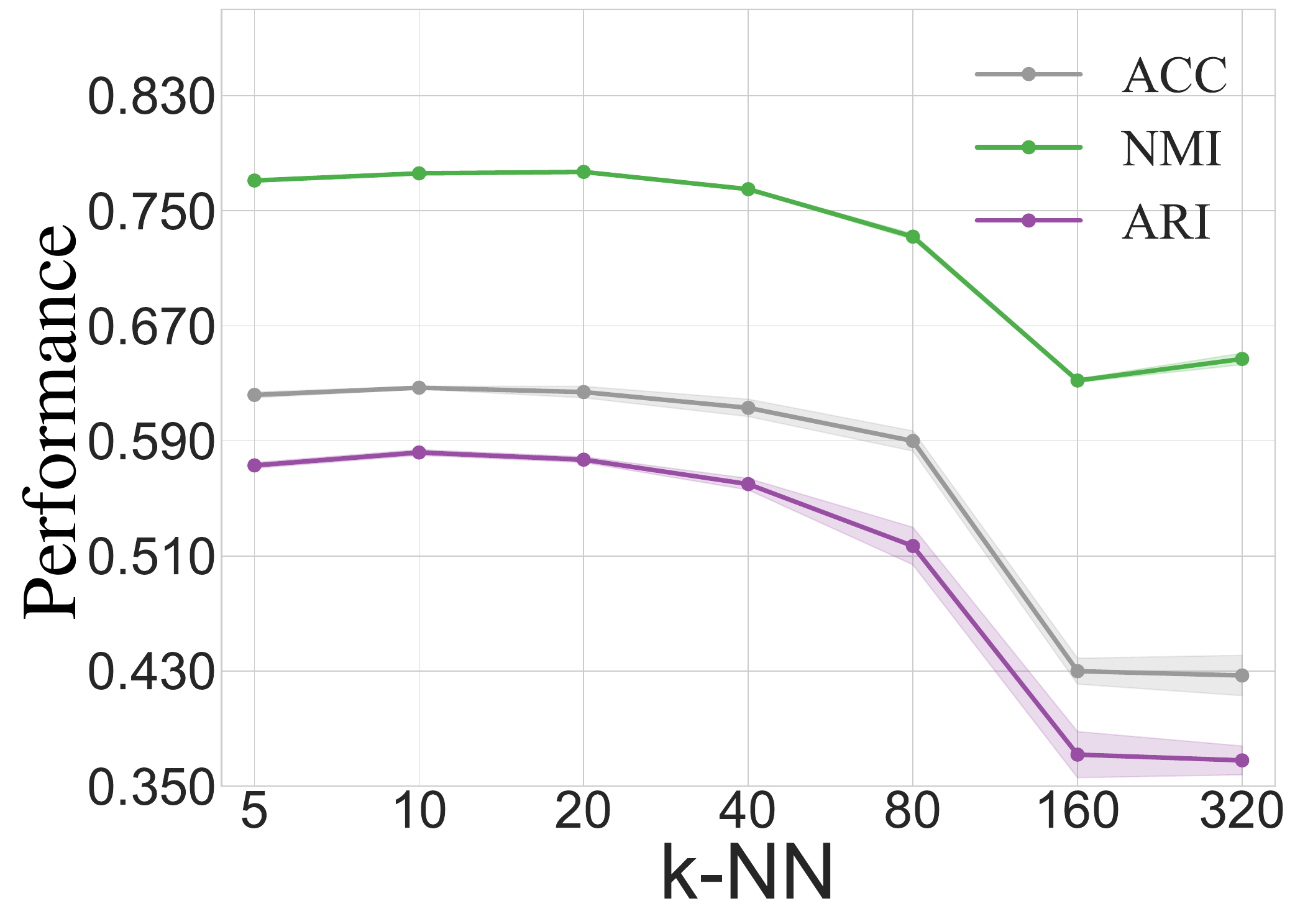}
        {(b) ISOLET} 
    \end{minipage} 
    \begin{minipage}[b]{0.19\linewidth}
        \centering
        \includegraphics[width=\linewidth]{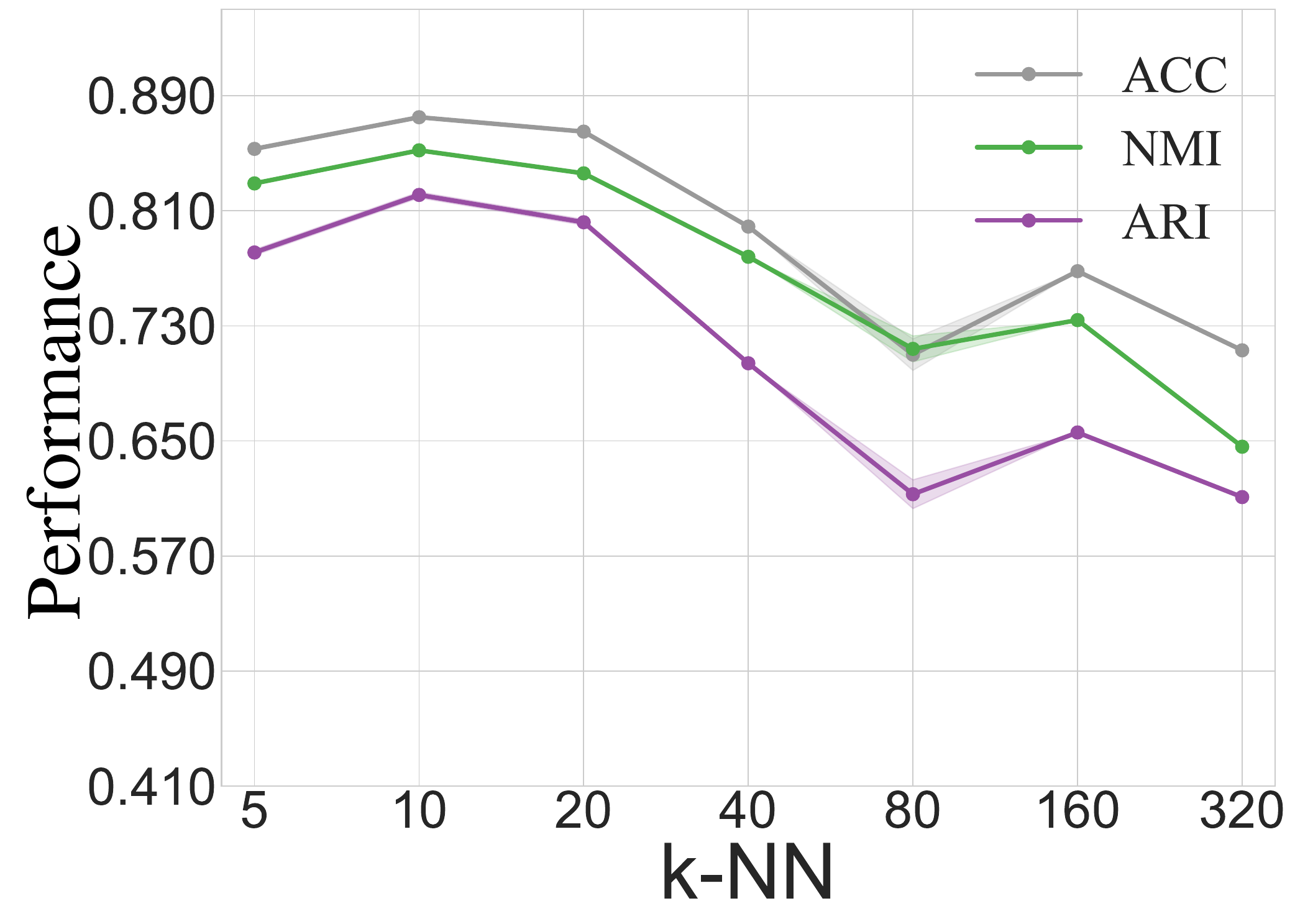}
        {(c) MNIST} 
    \end{minipage} 
    \begin{minipage}[b]{0.19\linewidth}
        \centering
        \includegraphics[width=\linewidth]{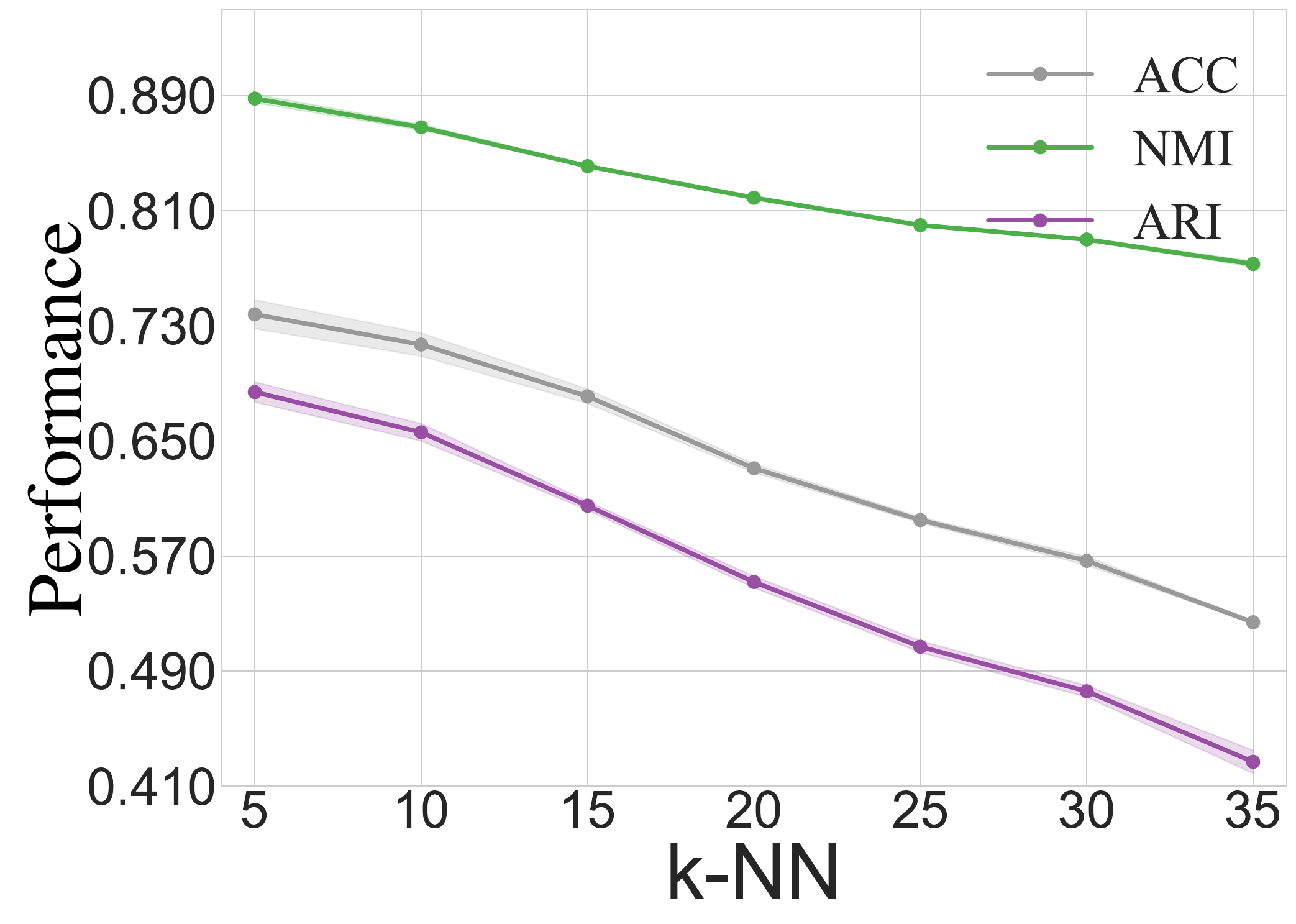}
        {(d) COIL-100} 
    \end{minipage} 
    \begin{minipage}[b]{0.19\linewidth}
        \centering
        \includegraphics[width=\linewidth]{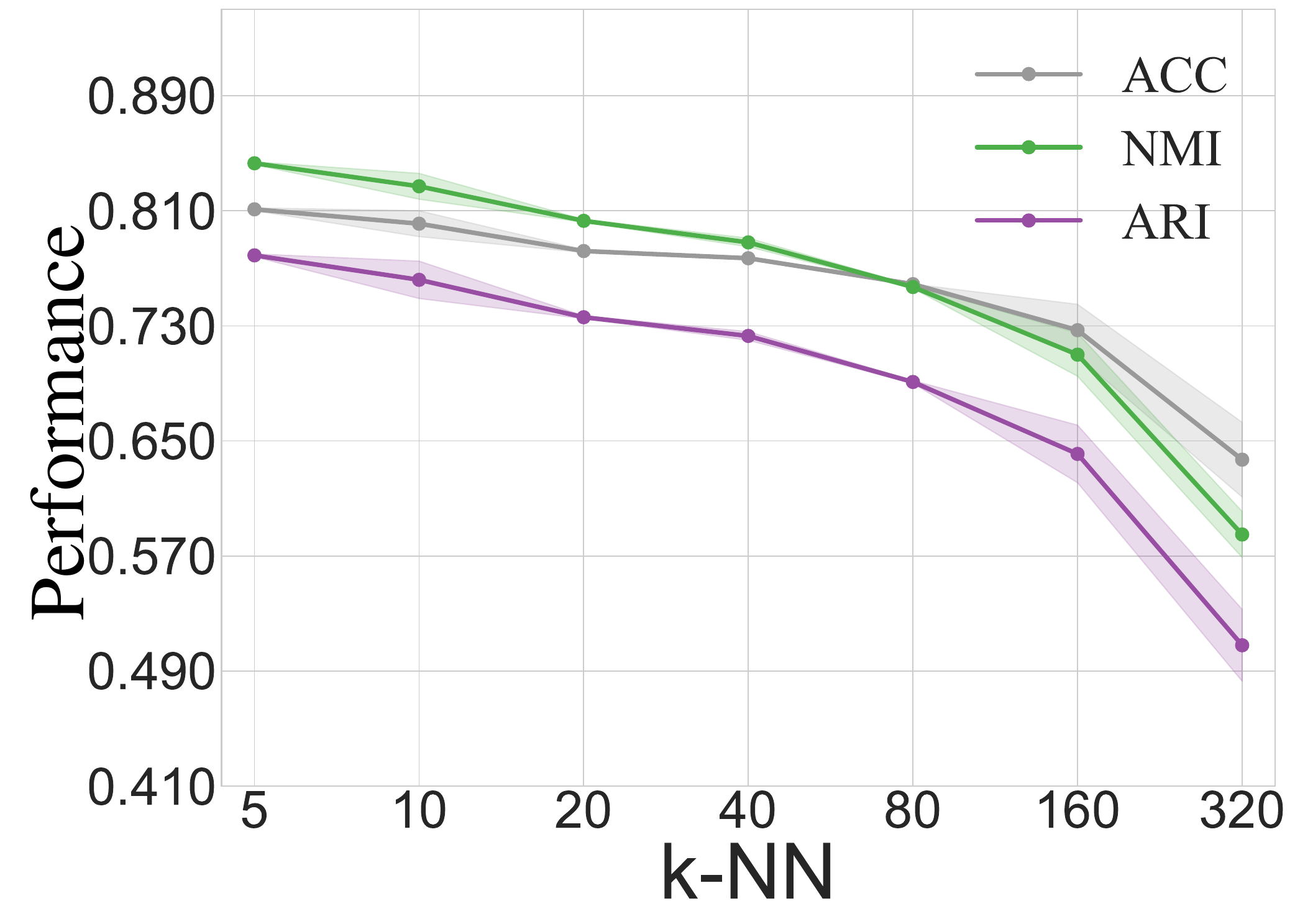}
        {(e) USPS} 
    \end{minipage} 
    
    \vspace{0.3cm} 
    \begin{minipage}[b]{0.19\linewidth}
        \centering
        \includegraphics[width=\linewidth]{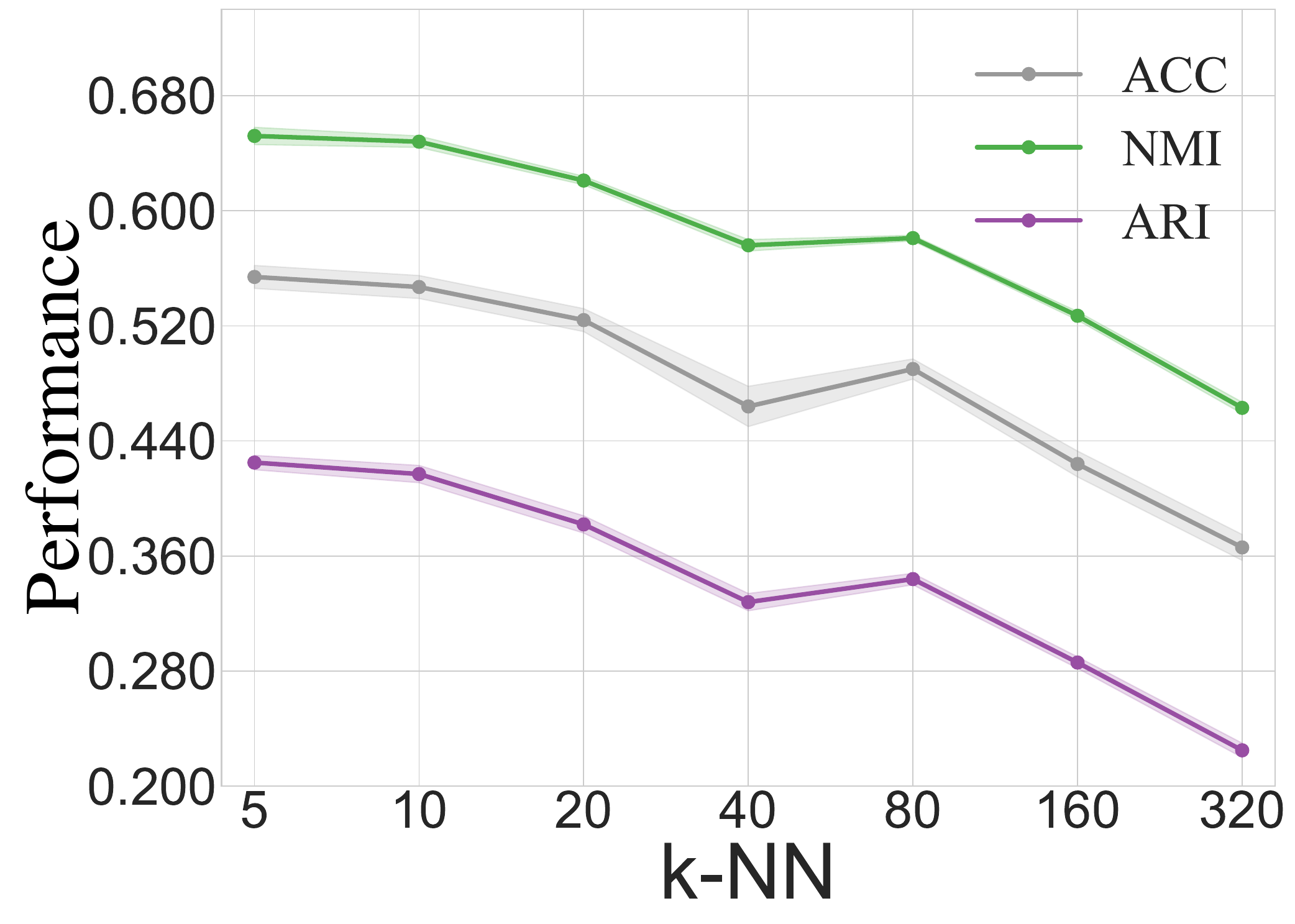}
        {(f) EMNIST} 
    \end{minipage} 
    \begin{minipage}[b]{0.19\linewidth}
        \centering
        \includegraphics[width=\linewidth]{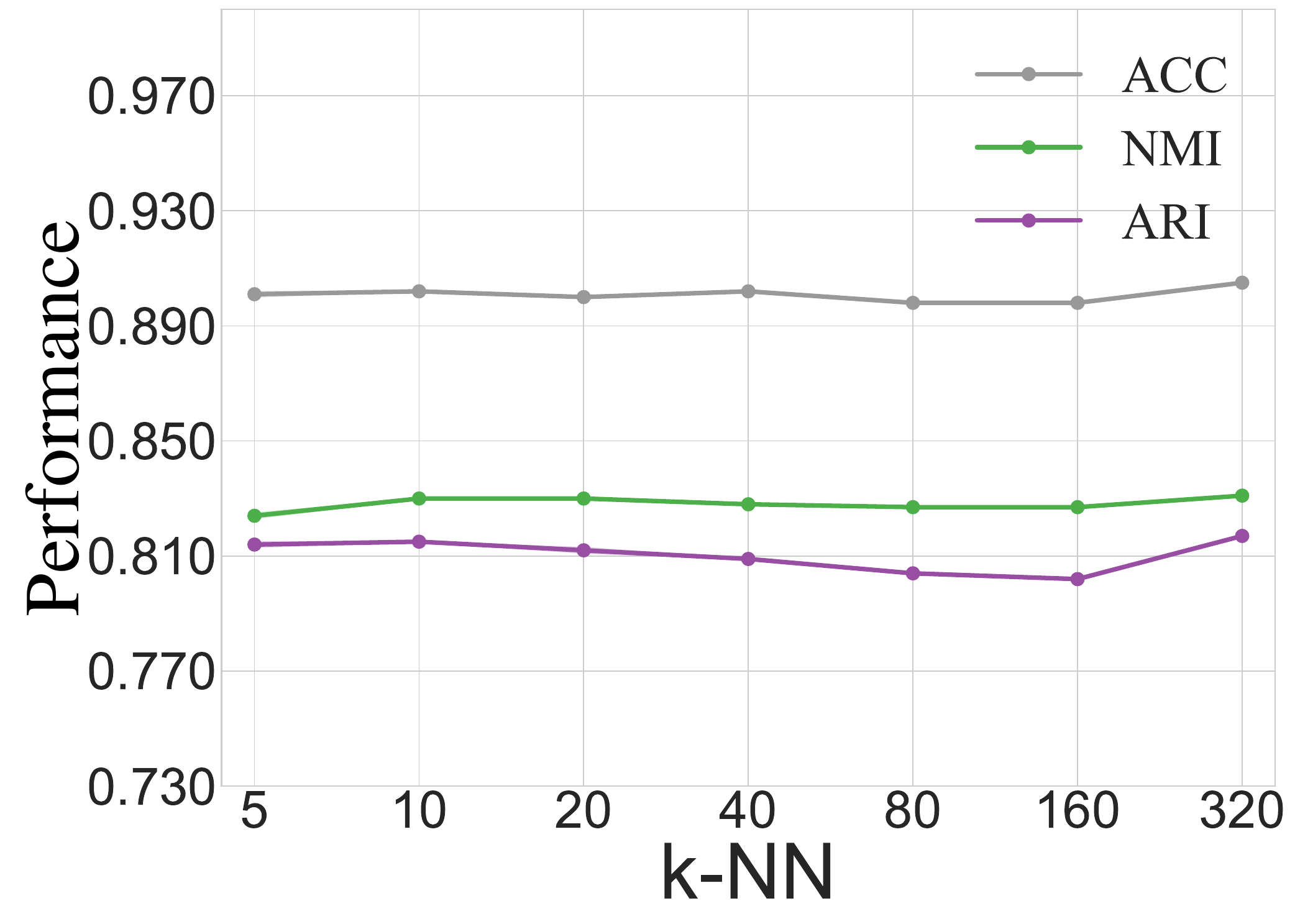}
        {(g) CIFAR-10} 
    \end{minipage} 
    \begin{minipage}[b]{0.19\linewidth}
        \centering
        \includegraphics[width=\linewidth]{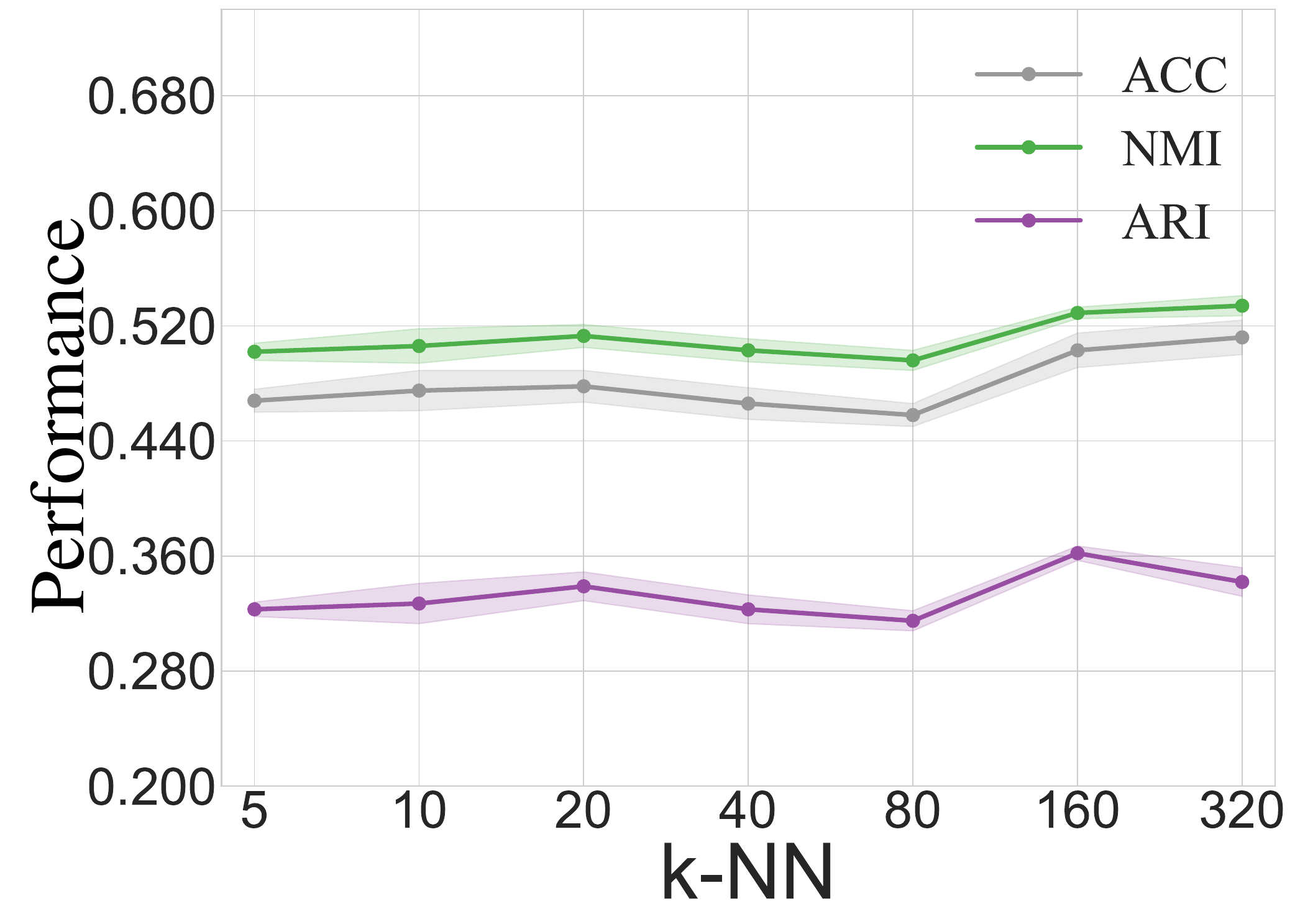}
        {(h) CIFAR-20} 
    \end{minipage} 
    \begin{minipage}[b]{0.19\linewidth}
        \centering
        \includegraphics[width=\linewidth]{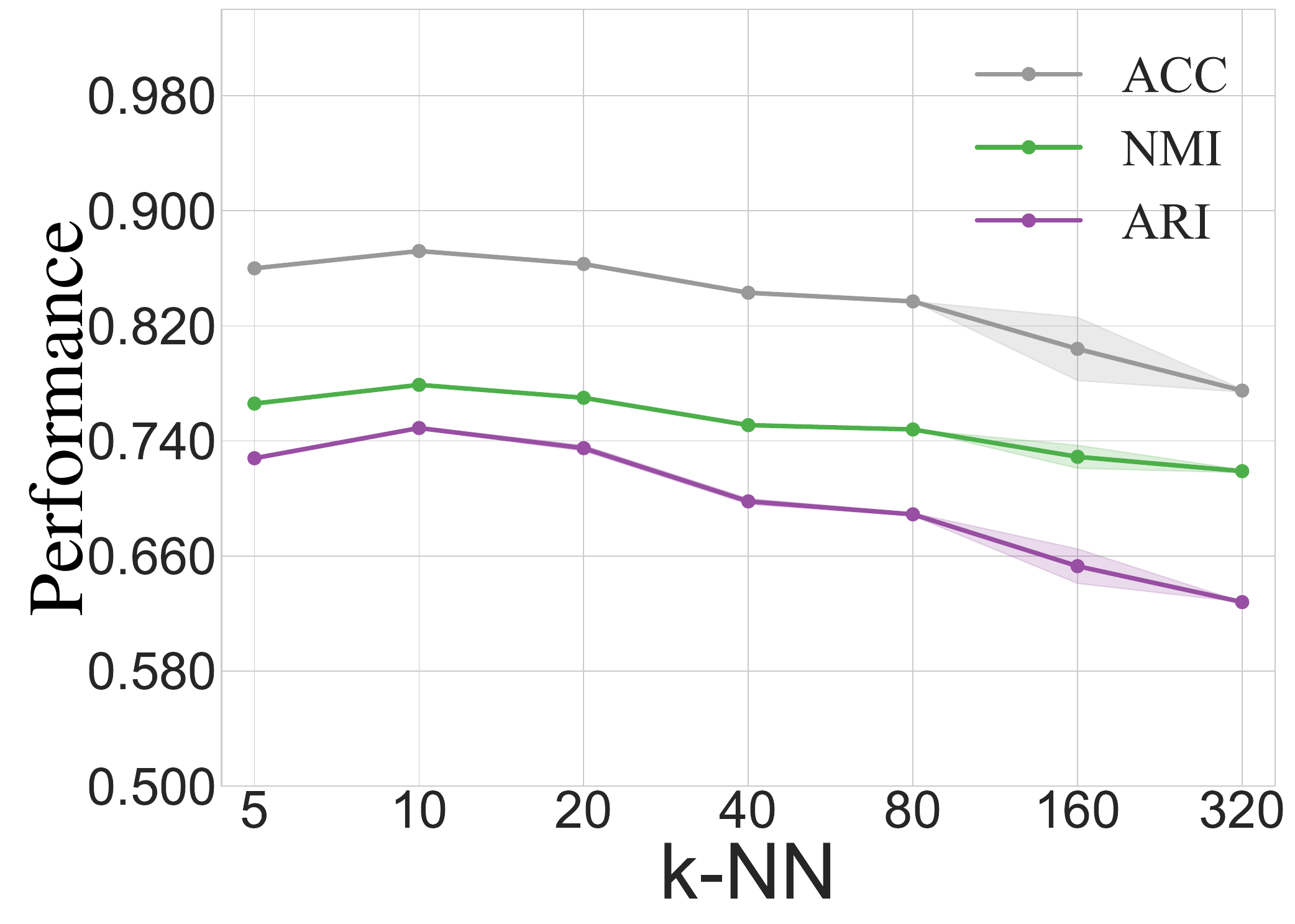}
        {(i) STL-10} 
    \end{minipage} 
    \begin{minipage}[b]{0.19\linewidth}
        \centering
        \includegraphics[width=\linewidth]{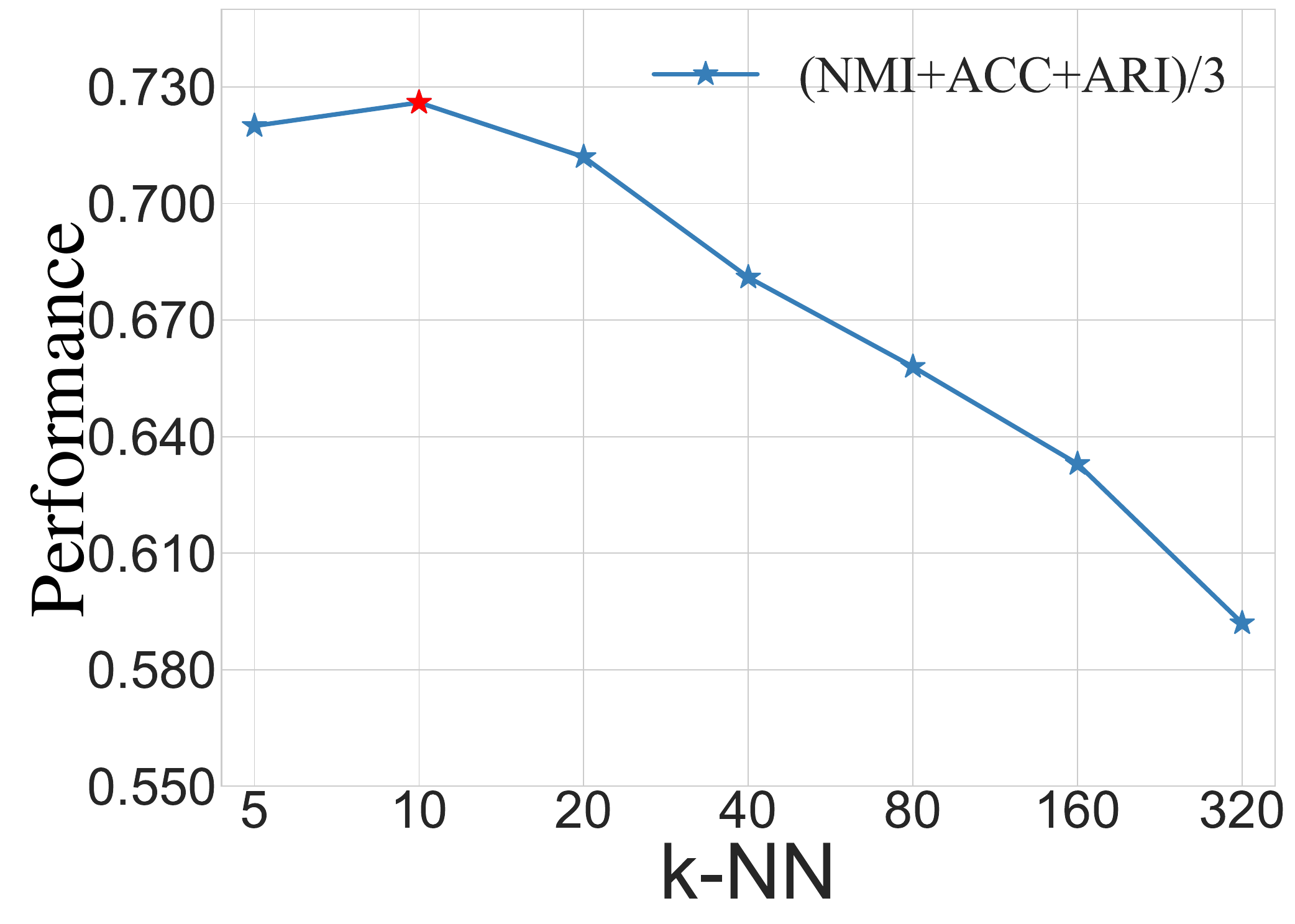}
        {(j) Mean of all results} 
    \end{minipage} 
    
    \caption{Performance of GPAC with different $k$ in $k$-NN.}
    \label{fig_3}
\end{figure*}
\begin{figure}[h]
    \centering
    \scriptsize
    \begin{minipage}[b]{0.45\linewidth}
        \centering
        \includegraphics[width=\linewidth]{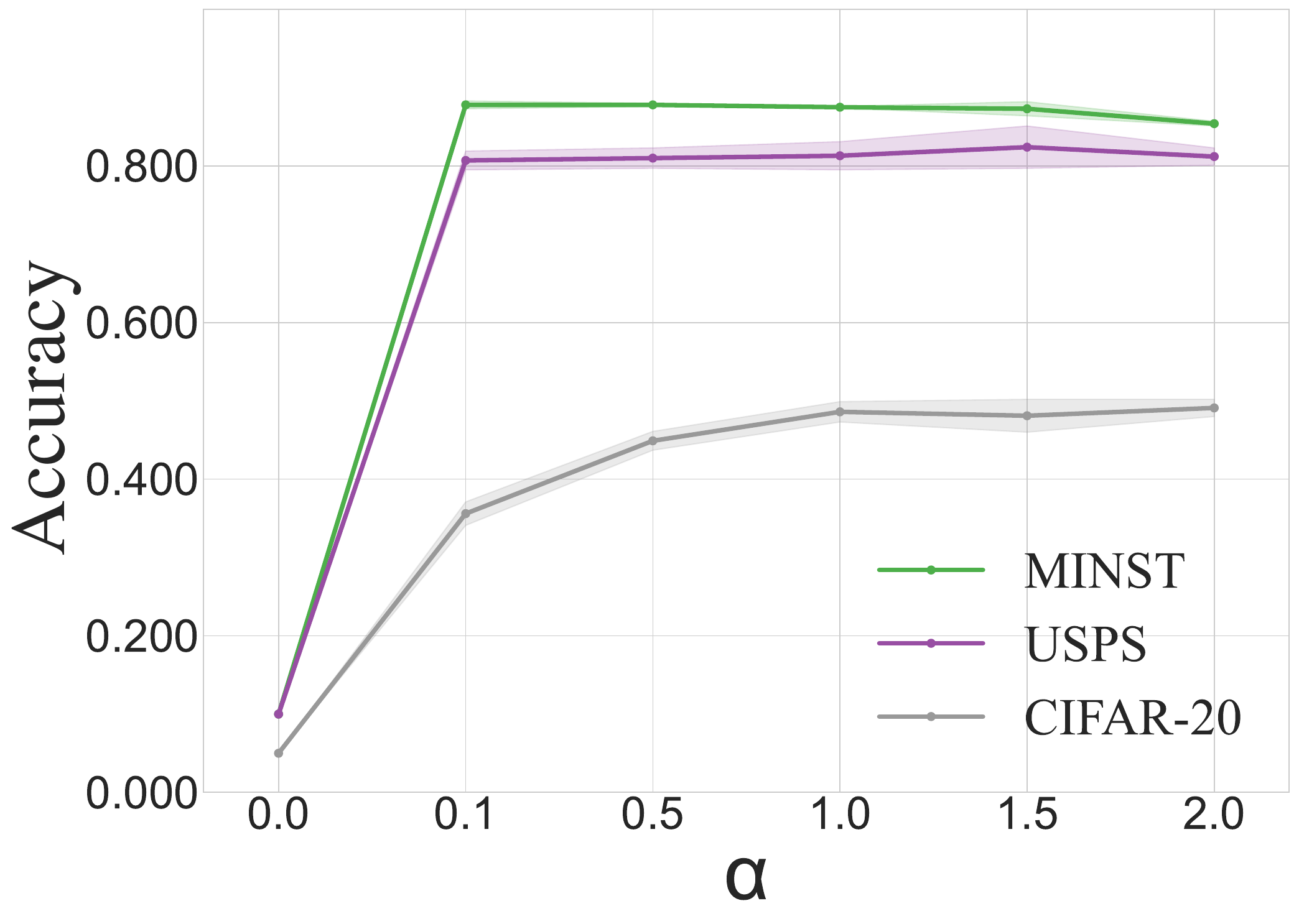}
        {(a) $\alpha$} 
    \end{minipage} 
    \begin{minipage}[b]{0.45\linewidth}
        \centering
        \includegraphics[width=\linewidth]{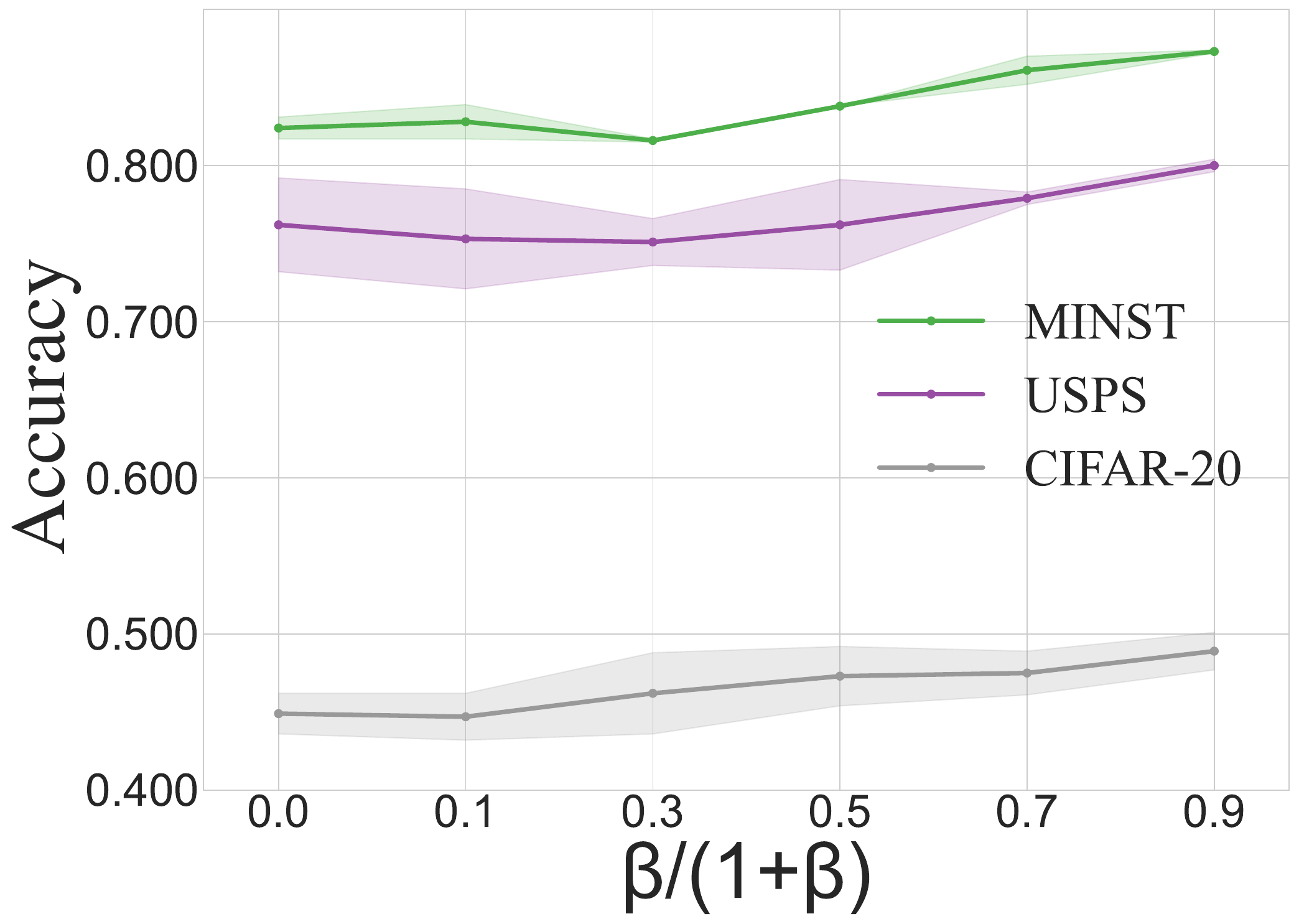}
        {(b) $\beta$} 
    \end{minipage} 
    \caption{Performance of GPAC with different $\alpha$ and $\beta$.}
    \label{fig_4}
\end{figure}

\subsubsection{ImageNet Evaluation}
To evaluate the scalability of GPAC on large-scale datasets with more instances and clusters, we perform experiments based on ImageNet. We implement CLIP with ViT-B/32 as a new deep feature extractor to further assess the universality of our method across different types of deep feature. Following the setup in ProPos \cite{huang2022learning}, we select subsets of ImageNet with 10, 50, 100, 200, and 1000 classes and set $m=1.02$ in GPAC. We compare GPAC with two deep clustering methods, SCAN \cite{van2020scan} and ProPos. As shown in Table \ref{tab6}, GPAC demonstrates comparable performance to deep methods, particularly on smaller subsets with fewer classes. Compared to CLIP+KM, GPAC consistently shows performance improvement, highlighting its potential for effective scaling in deep learning applications.

\begin{table}[h]
	\caption{Scalability evaluation of GPAC on large-scale dataset. Metric: NMI.}
	\label{tab6}
	\centering
	   \begin{tabular}{l |ccc cc}
			\hline
            Dataset &\multicolumn{5}{c}{IMAGENET} \\ 
            \hline
			Sub-classes 
            &10
			&50 
			&100 
			&200 
                &1000 \\
          \hline
            CLIP+KM &0.878 &0.809 &0.764 &0.724 &0.662 \\
            SCAN &0.898 &0.805 &0.787  &0.757 &0.707\\
            ProPos &0.959 &0.828 &\textbf{0.835}  &\textbf{0.806} &-\\
            CLIP+GPAC &\textbf{0.978}  &\textbf{0.860} & 0.822  &0.781 &\textbf{0.715} \\
           \hline
		\end{tabular}
\end{table}

\subsubsection{Parameter Sensibility Analysis}
We evaluate the parameter sensitivity of the proposed algorithm by analyzing the effects of four key parameters: $m$, $k$-NN, $\alpha$, and $\beta$. The parameter $m$ governs the degree of fuzziness in the clustering, directly influencing the confidence of the predictions by controlling the level of overlap between clusters. The $k$-nearest neighbors ($k$-NN) parameter defines the scope of local neighborhood influence, which in turn shapes the local structure and proximity relationships within the data. The hyperparameter $\alpha$ regulates the strength of the self-constrained term, affecting the size of the clusters. $\beta$ controls the trade-off between global and local clustering, balancing the broader dataset structure with the finer local details. For each parameter, fix the other parameters to the default settings and change one parameter to perform sensitivity experiments.

The fuzzy weight exponent $m$ is a fundamental parameter in fuzzy clustering algorithms. To investigate its impact, we compare the influence of $m$ on GPAC, PAC, and FCM within the range $(1, 1.5]$ with a step size of 0.05. The experimental results are presented in Fig. \ref{fig_2}. Based on the results, we observe that GPAC exhibits minimal sensitivity to $m$, maintaining stable performance across various values of $m$. However, for the COIL-100 dataset, it is evident that when the number of clusters is too large, larger values of $m$ can lead to overly fuzzy outputs. Therefore, we recommend selecting $m$ based on the number of clusters to be formed: as the number of clusters increases, smaller values of $m$ should be chosen.

We test the influence of the $k$-NN parameter within the values $\{5, 10, 20, 40, 80, 160, 320\}$. The results shown in Fig. 7. As illustrated in Fig. \ref{fig_3}, we find that the performance of GPAC tends to decrease as $k$ increases. This is because the effectiveness of local clustering diminishes with a larger number of neighbors, as the local structure becomes more global. Additionally, larger values of $k$ introduce the extra computational burden. Therefore, from the average of all results, we find that $k = 10$ is an optimal choice for most datasets, offering a good balance between performance and computational efficiency.

Finally, we test the influence of $\alpha$ and $\beta$ and the results shown in Fig. \ref{fig_4}, respectively. From experiments we observe that (i) $\alpha = 1$ provides a good balance between the clustering and self-constrained terms; (ii) $\beta$ controls the strength of local clustering, and its effect improves as $\frac{\beta}{1+\beta}$ approaches 1.

\subsubsection{Effect of Auxiliary Hard Assignment Matrix $\boldsymbol{V}$}
\begin{figure}[h]
    \centering
    \scriptsize
    \begin{minipage}[b]{0.45\linewidth}
        \centering
        \includegraphics[width=\linewidth]{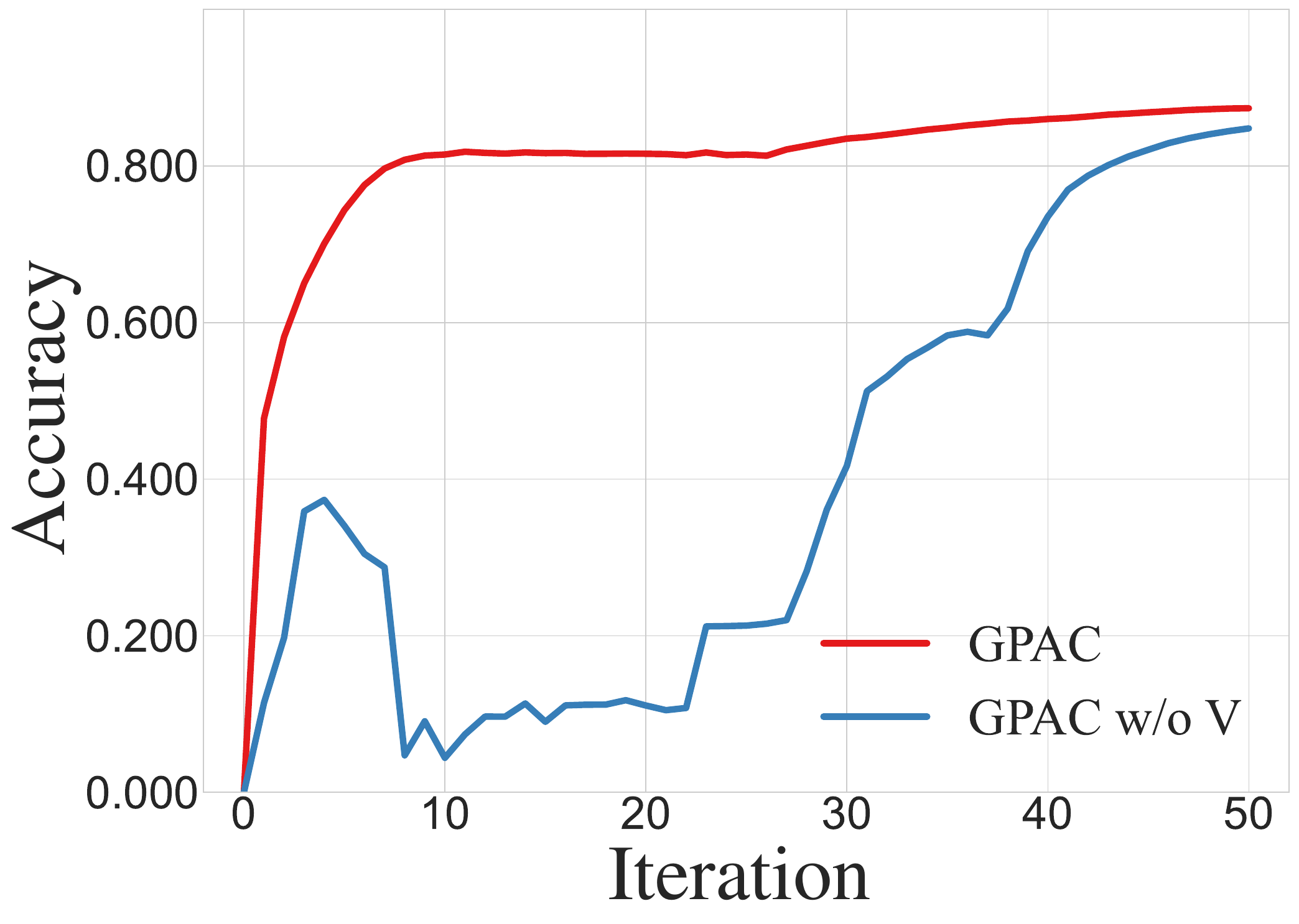}
        {(a) MNIST} 
    \end{minipage} 
    \begin{minipage}[b]{0.45\linewidth}
        \centering
        \includegraphics[width=\linewidth]{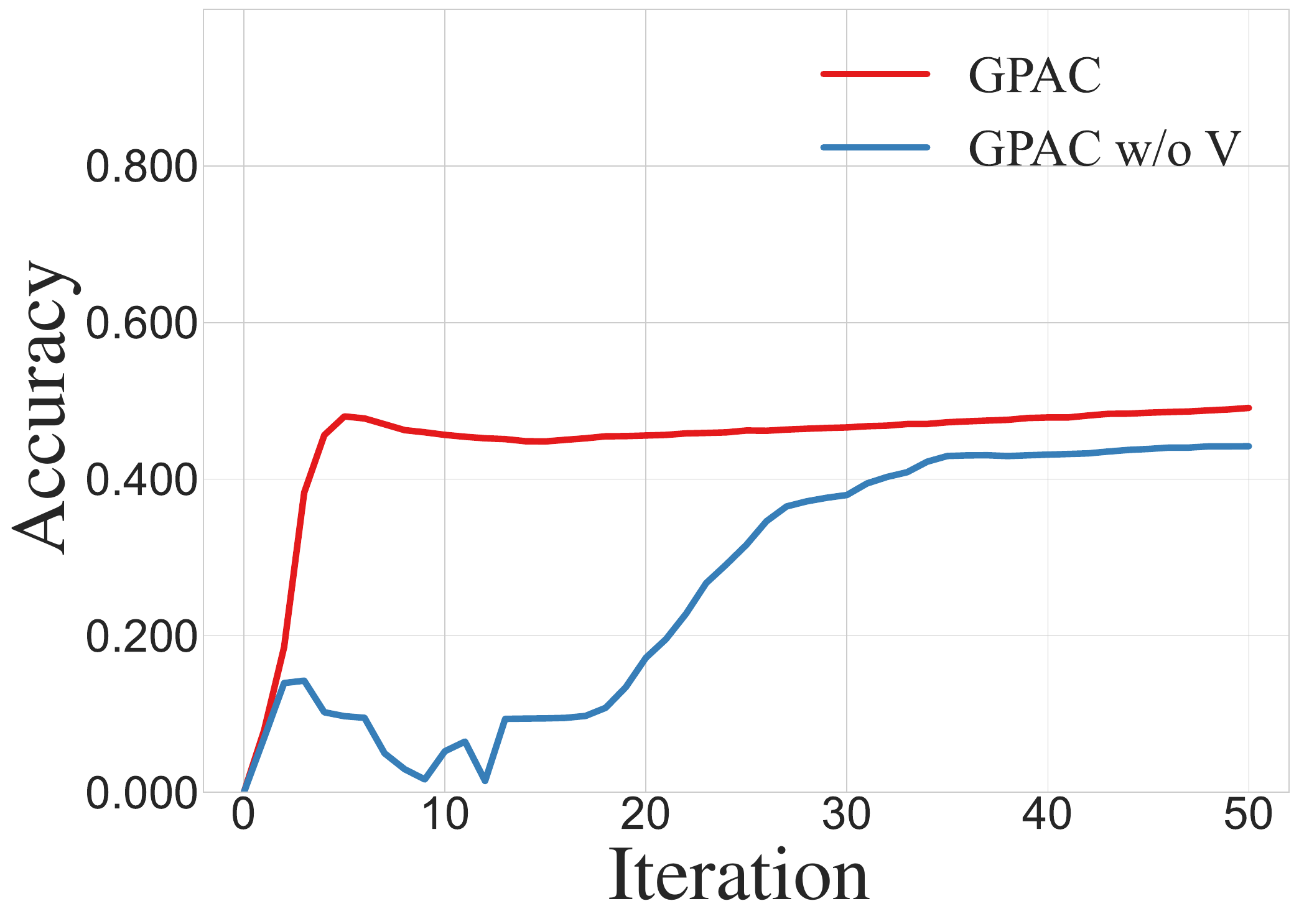}
        {(b) USPS} 
    \end{minipage} 
    
      \vspace{0.3cm} 
      
      \begin{minipage}[b]{0.45\linewidth}
        \centering
        \includegraphics[width=\linewidth]{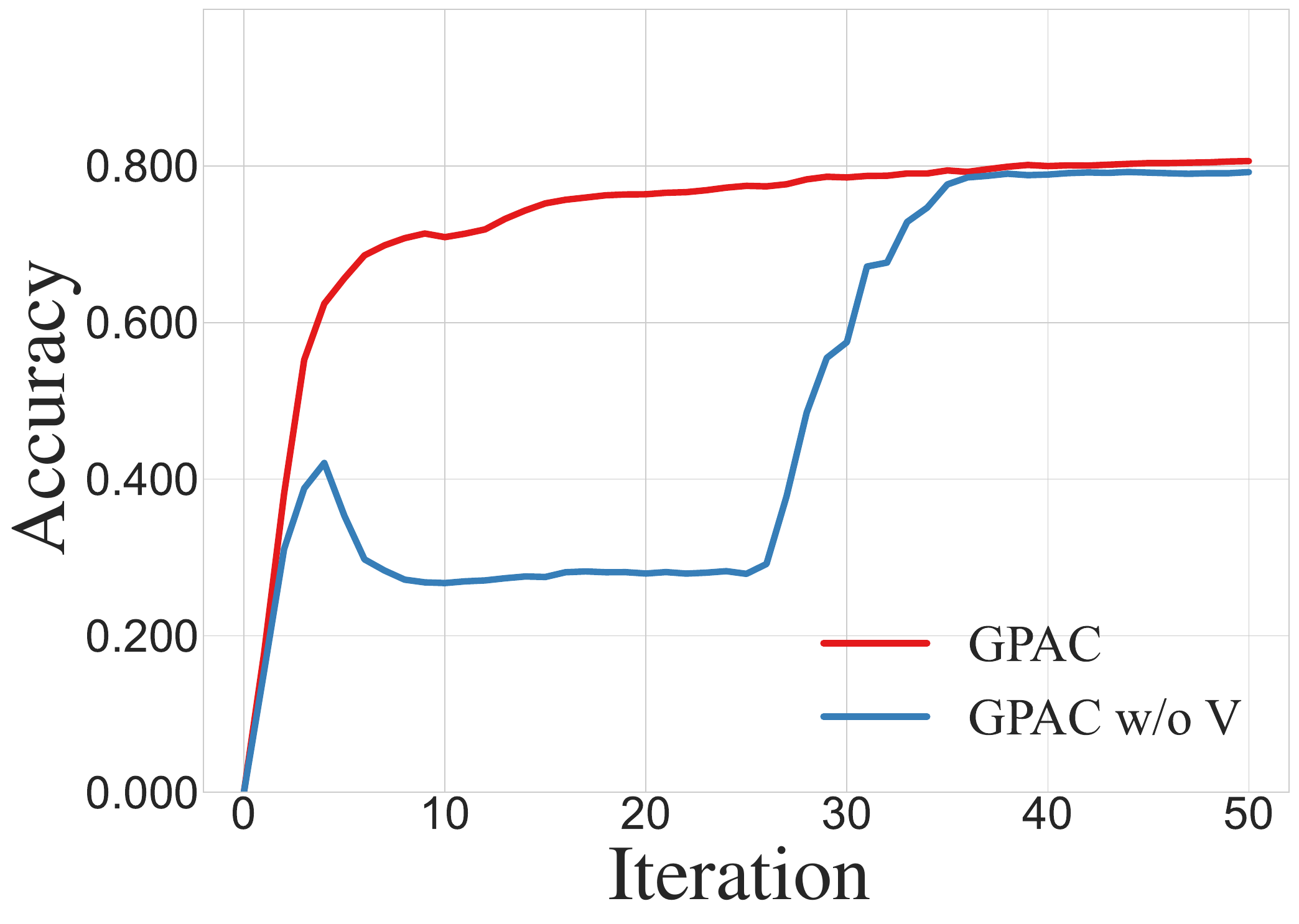}
        {(c) CIFAR-20} 
    \end{minipage} 
    \begin{minipage}[b]{0.45\linewidth}
        \centering
        \includegraphics[width=\linewidth]{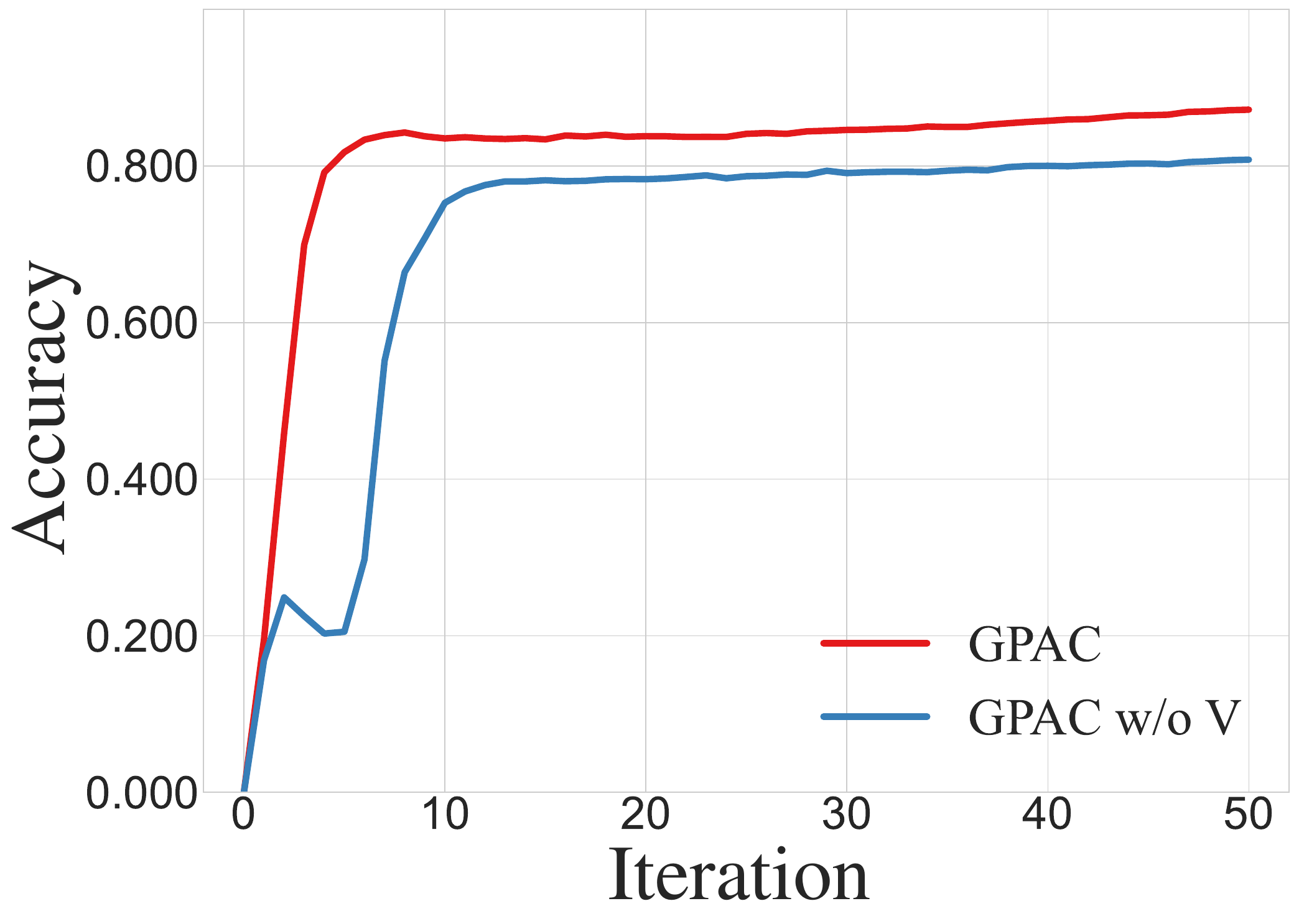}
        {(d) STL-10} 
    \end{minipage} 
    \caption{Effect of Auxiliary Hard Assignment Matrix $\boldsymbol{V}$.}
    \label{fig_5}
\end{figure}

We evaluate the impact of the hard assignment matrix $\boldsymbol{V}$, which facilitates faster convergence during the optimization process. We replace $\boldsymbol{V}$ with $\boldsymbol{P}$ and mark it as GPAC without (w/o) $\boldsymbol{V}$. We remove Eq. (\ref{eqn-2-33}) in experiments to avoid the effects of clustering score regularization. As shown in Fig. \ref{fig_5}, we record the accuracy at each iteration. The experiments indicate that $\boldsymbol{V}$ significantly accelerates convergence in the early stages and helps maintain stability in the predictions during the later stages, ultimately improving overall performance.

\subsubsection{Effect of Local Clustering Constraint}
\begin{figure}[h]
    \centering
    \scriptsize
    \begin{minipage}[b]{0.45\linewidth}
        \centering
        \includegraphics[width=\linewidth]{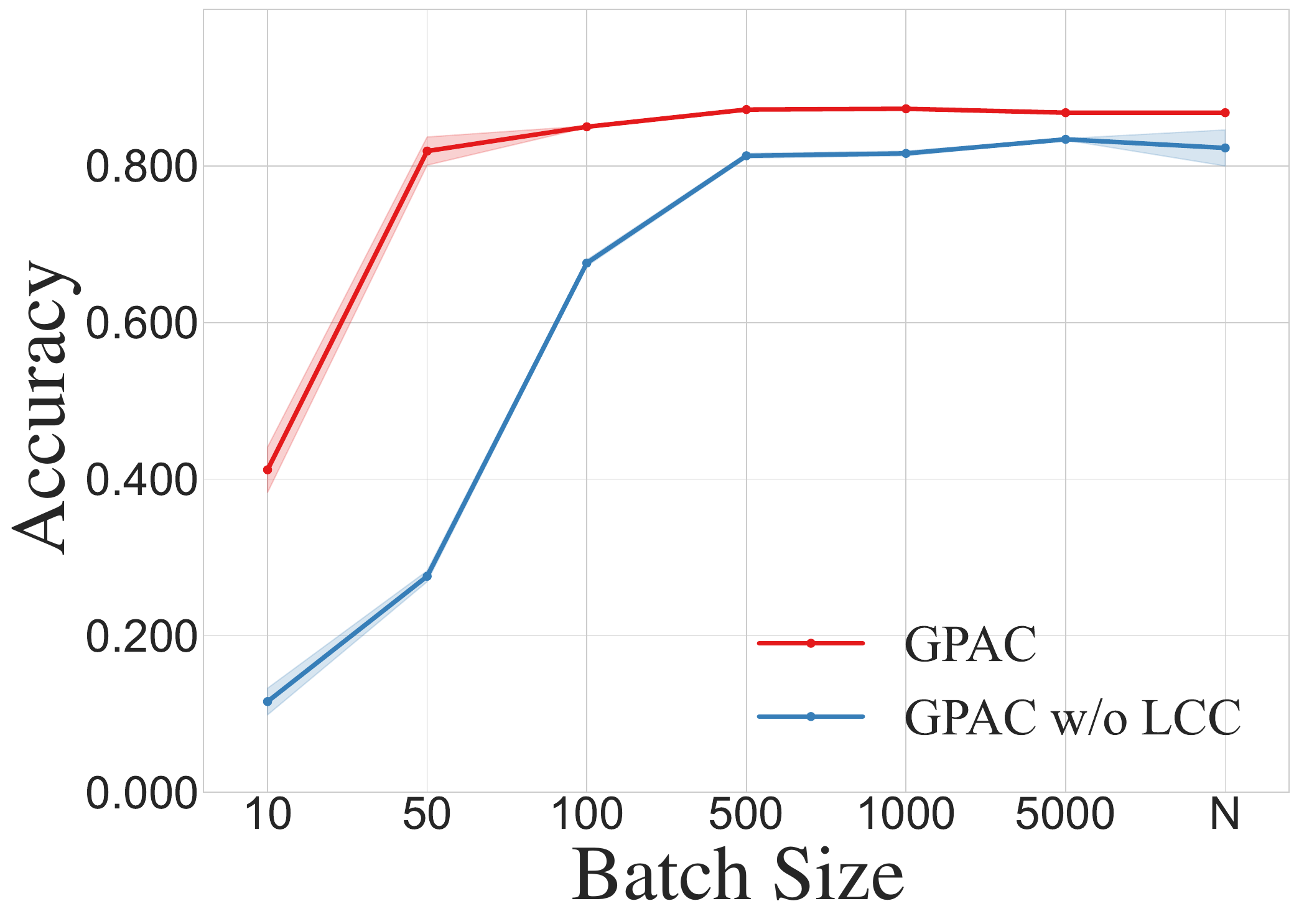}
        {(a) MNIST} 
    \end{minipage} 
    \begin{minipage}[b]{0.45\linewidth}
        \centering
        \includegraphics[width=\linewidth]{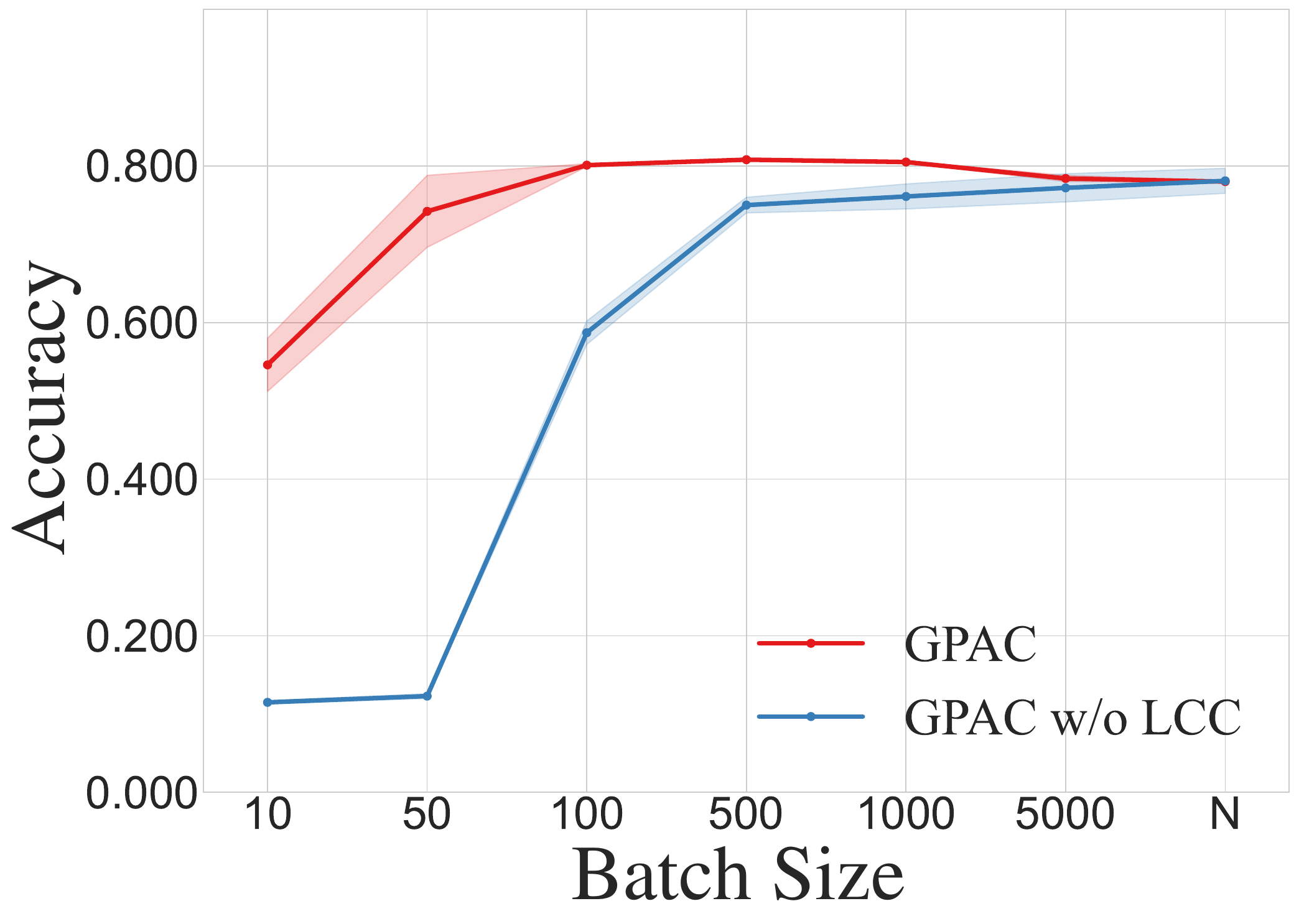}
        {(b) USPS} 
    \end{minipage} 
    
      \vspace{0.3cm} 
      
      \begin{minipage}[b]{0.45\linewidth}
        \centering
        \includegraphics[width=\linewidth]{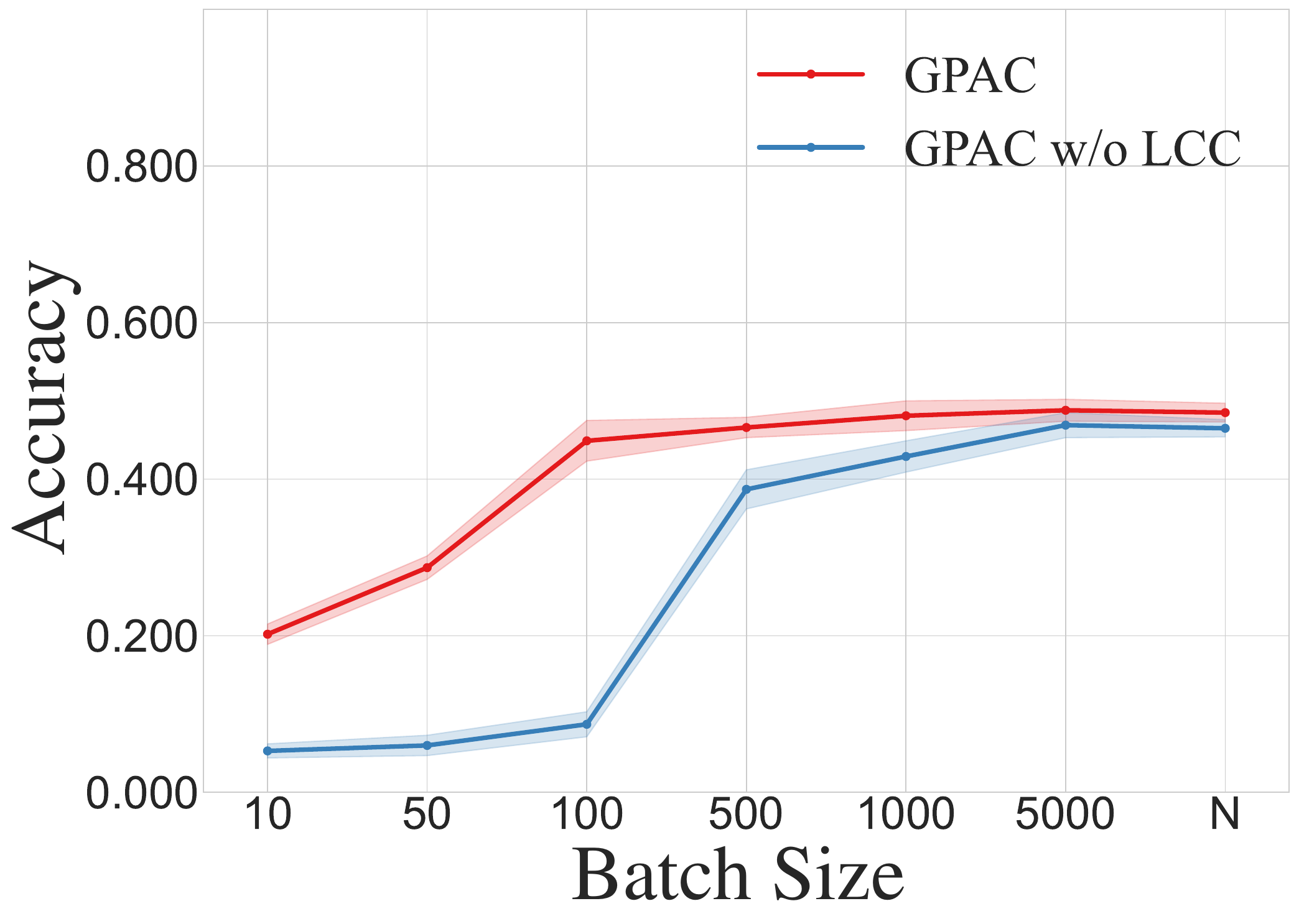}
        {(c) CIFAR-20} 
    \end{minipage} 
    \begin{minipage}[b]{0.45\linewidth}
        \centering
        \includegraphics[width=\linewidth]{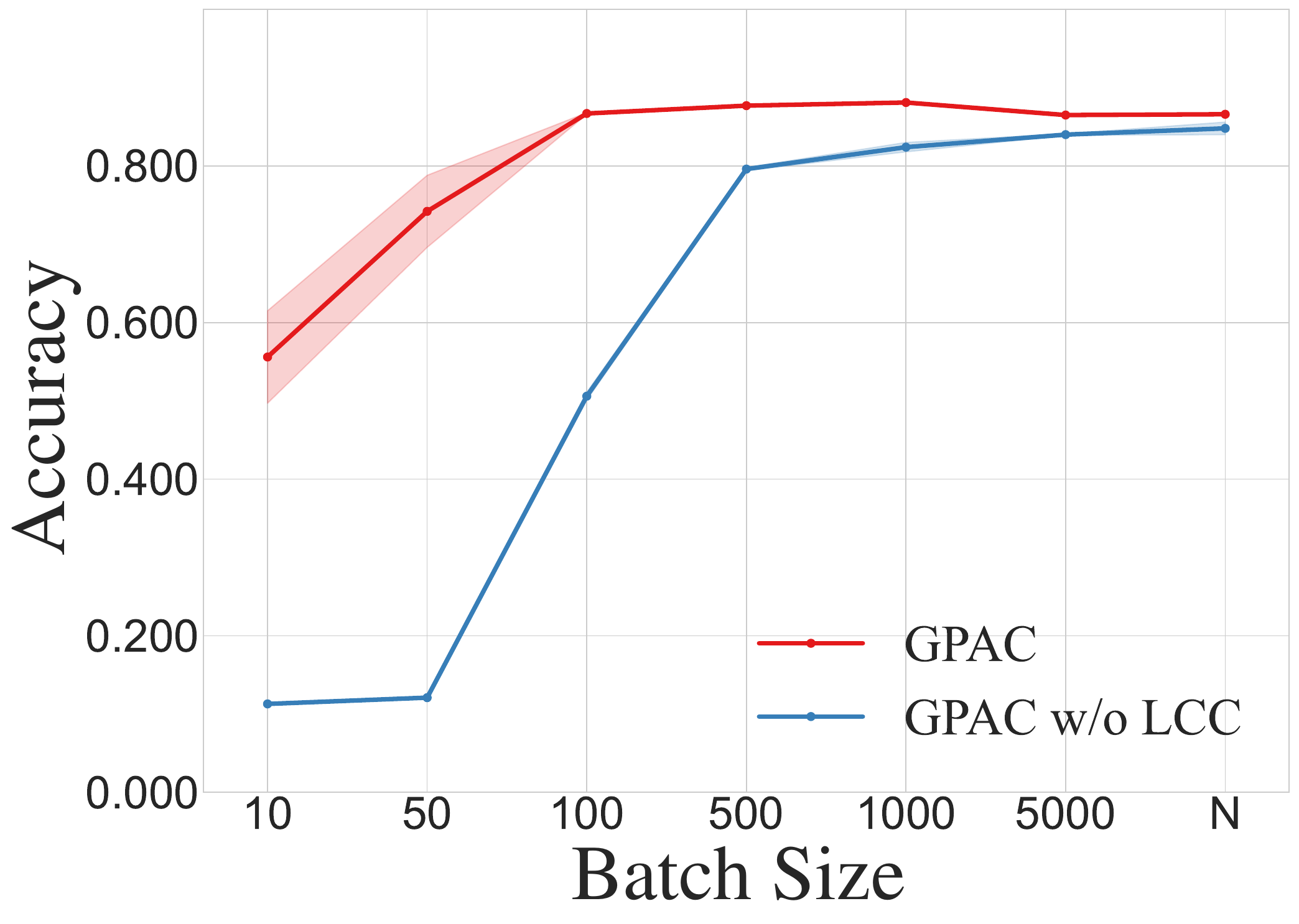}
        {(d) STL-10} 
    \end{minipage} 
    \caption{Performance of GPAC and GPAC w/o LCC with different batch size.}
    \label{fig_6}
\end{figure}

We propose global and local clustering to jointly explore unlabeled data in GPAC. To comprehensively assess the impact of the local clustering constraint (LCC), we perform a series of experiments. In particular, we integrate the LCC into the FCM problem, resulting in a final iteration solution that is similar to that presented in Eq. (\ref{eqn-2-21}). For comparison, we also evaluate the performance of GPAC without LCC by setting the regularization parameter $\beta = 0$, effectively removing the local clustering influence.
The experimental results, summarized in Table \ref{tab5}, demonstrate a marked improvement in clustering accuracy when the LCC is incorporated. Specifically, the inclusion of local neighborhood information significantly enhances the ability of fuzzy clustering methods to capture the underlying data structure, particularly in cases with complex or highly localized patterns. Additionally, as shown in Fig. \ref{fig_6}, the local clustering mitigates the typical performance degradation caused by mini-batch global clustering, where the lack of broader context leads to a suboptimal solution. 

\begin{table}[h]
	\caption{Clustering results on the ablation study of Local Clustering Constraint (LCC).}
	\label{tab5}
	\centering
	   \begin{tabular}{c c|cccc}
			\hline
            Method &+LCC &Metric &MNIST &USPS  &CIFAR-20 \\
          \hline
            \multirow{3}{*}{FCM} &\multirow{3}{*}{$\times$}   
                &NMI  &0.489 &0.639 &0.471\\
                & &ACC  &0.546 &0.623 &0.427\\
                & &ARI  &0.370 &0.532 &0.259\\
            \hline 
            \multirow{3}{*}{FCM} &\multirow{3}{*}{$\checkmark$}      
                &NMI  &0.705 &0.772 &0.508\\
                & &ACC  &0.699 &0.787&0.465\\
                & &ARI  &0.603 &0.699 &0.316\\
            \hline
            Boost  & &Avg. &+0.207 &+0.155 &+0.044\\
            \hline
            \multirow{3}{*}{GPAC} &\multirow{3}{*}{$\times$}      
                &NMI  &0.750 &0.731 &0.476\\
               & &ACC  &0.824 &0.774 &0.469\\
               & &ARI  &0.718 &0.663 &0.317\\
            \hline
            \multirow{3}{*}{GPAC} &\multirow{3}{*}{$\checkmark$}      
                &NMI  &0.852 &0.836 &0.514\\
               & &ACC  &0.875 &0.808 &0.486\\
               & &ARI  &0.821 &0.770 &0.335\\
            \hline
          Boost  & &Avg. &+0.085 &+0.082 &+0.024\\
           \hline
		\end{tabular}
\end{table}

\begin{figure}[h]
    \centering
    \scriptsize
    \includegraphics[width=0.8\linewidth]{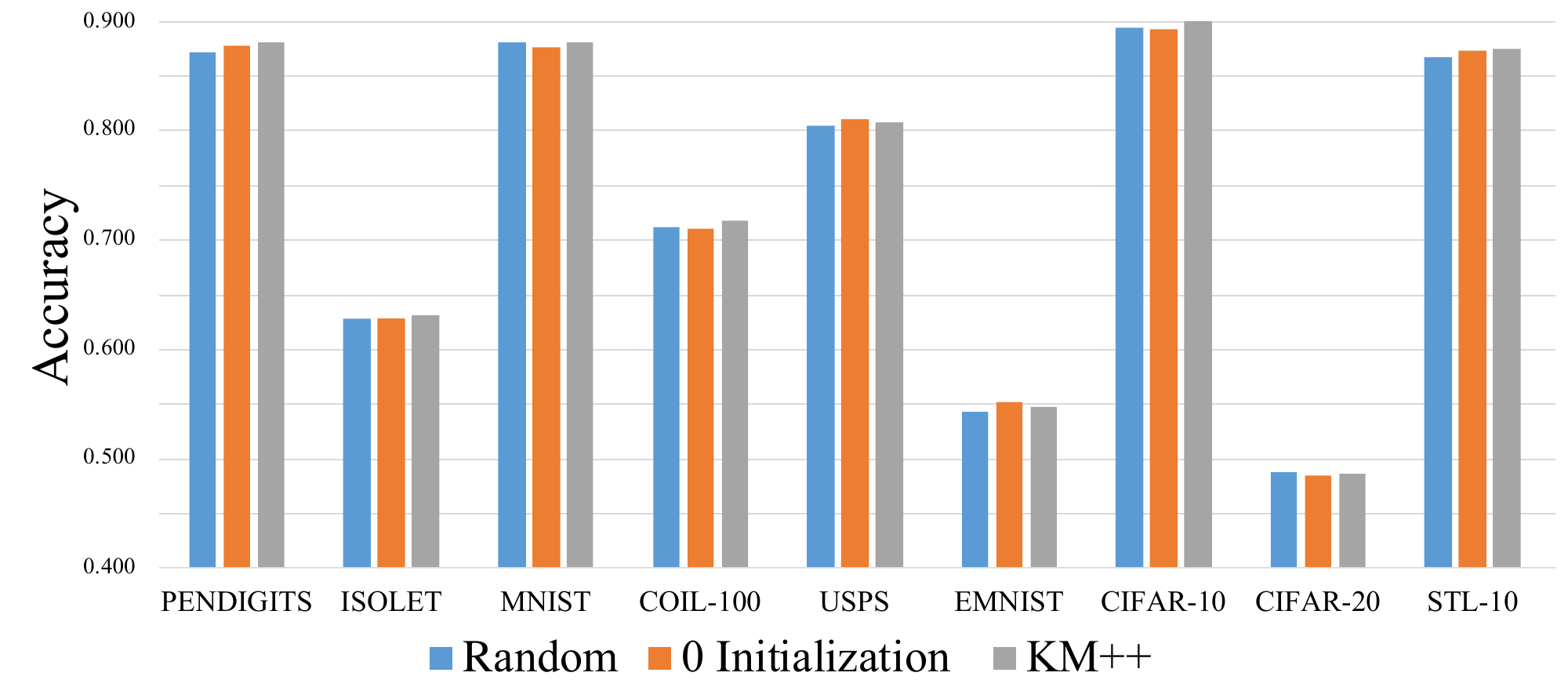}
    \caption{Impact of initialization methods for $\boldsymbol{V}$.}
    \label{fig_7}
\end{figure}

\subsubsection{Effect of Initialization}
We compare different initialization approaches to test the effect of initialization in GPAC, using random, zero initialization, KM++ initialization for $\boldsymbol{V}$. Random initialization assigns a random cluster to each sample, zero initialization fills $\boldsymbol{V}$ with 0, and KM++ initialization uses KM++ algorithm to assign samples. The experimental results, shown in Fig. \ref{fig_7}, demonstrate that the best performance of the proposed algorithm is achieved with KM++ initialization. Additionally, we can conclude that GPAC is not sensitive to initialization and random initialization is also a reasonable choice.

\subsubsection{Execution Time}
\begin{figure}[h]
    \centering
    \scriptsize
    \includegraphics[width=0.9\linewidth]{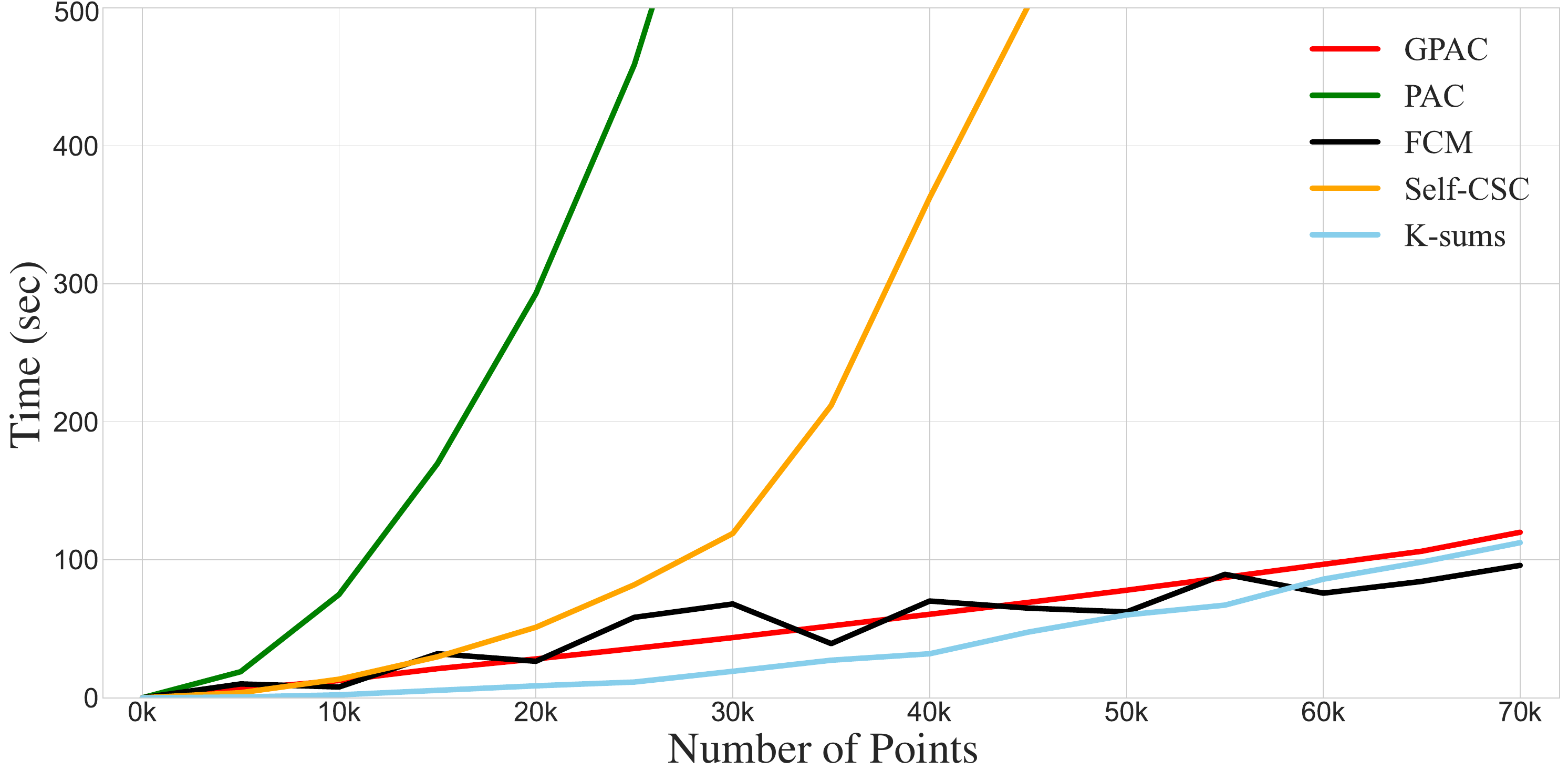}
    \caption{Comparison of running times for GPAC and other algorithms on MNIST subsets.}
    \label{fig_8}
\end{figure}

The theoretical time complexity of the GPAC algorithm is linear. To evaluate its practical execution time, we sample subsets of varying sizes from the MNIST dataset. We record the actual running time of the GPAC optimization process, scaling with increasing sub-dataset size. For comparison, we also report the running times of several other algorithms, including PAC, Self-CSC, K-sums, and FCM. The algorithms PAC, GPAC, and FCM are implemented in Python, Self-CSC in Matlab, and K-sums in C++. The reported running time reflects the entire execution time, including the computation of pairwise distances, the construction of the $k$-NN graph, and the optimization iterations. All experiments are conducted on a personal computer equipped with an AMD Ryzen 9 7900X 12-Core Processor (4.70 GHz) and 64 GB of RAM, using the official codes for each algorithm. The results are shown in Fig. \ref{fig_8}. GPAC demonstrates exceptional time efficiency, closely matching the performance of FCM and K-sums, despite Python being slower in handling loops compared to C++.

\section{Conclusion}
In summary, a novel fuzzy clustering method GPAC without cluster center was proposed from a very new perspective. We use the inner product of fuzzy probability vectors to explore both adjacent similarities and non-adjacent dissimilarities of samples. Furthermore, we propose local consistency constraint to learn neighborhood relationships to further improve performance. A theoretical model and an elegant iterative optimization solution for GPAC have been developed. Besides, a fast optimization algorithm has been constructed to reduce the complexity of the algorithm to linear to extend model practicability.  Finally, experiments on several benchmarks verified the effectiveness of our proposal.


%





\ifCLASSOPTIONcaptionsoff
  \newpage
\fi





\bibliographystyle{IEEEtran}
\bibliography{IEEEabrv,Bibliography}

\vfill


\end{document}